\documentclass{ieeeaccess}
\usepackage{cite}
\usepackage{amsmath,amssymb,amsfonts}

\usepackage{amsthm,amsmath,amssymb}
\usepackage{mathrsfs}

\usepackage{calligra}
\usepackage{algpseudocode} 
\usepackage{amsmath}

\usepackage[ruled,vlined]{algorithm2e} 
\SetKwRepeat{Loop}{loop}{end}

\usepackage{graphicx}
\ifCLASSOPTIONcompsoc
\usepackage[caption=false, font=normalsize, labelfont=sf, textfont=sf]{subfig}
\else
\usepackage[caption=false, font=footnotesize]{subfig}

\usepackage{textcomp}
\def\BibTeX{{\rm B\kern-.05em{\sc i\kern-.025em b}\kern-.08em
    T\kern-.1667em\lower.7ex\hbox{E}\kern-.125emX}}
\usepackage{multicol}
\usepackage{multirow}

\usepackage{booktabs}

\begin{document}
\history{Date of publication xxxx 00, 0000, date of current version xxxx 00, 0000.}
\doi{10.1109/ACCESS.2022.3190958}

\title{A Density Peaks Clustering Algorithm with Sparse Search and K-d Tree}
\author{\uppercase{YUNXIAO SHAN}\authorrefmark{1}, 
\uppercase{SHU LI\authorrefmark{1,2}, FUXIANG LI\authorrefmark{1}, YUXIN CUI\authorrefmark{1}, SHUAI LI}\authorrefmark{3}, MING ZHOU\authorrefmark{1}, AND XIANG LI\authorrefmark{1}}
\address[1]{School of Science, Harbin University of Science and Technology, Harbin 150080, China}
\address[2]{Key Laboratory of Engineering Dielectric and Applications (Ministry of Education), School of Electrical and Electronic Engineering, Harbin University of Science and Technology, Harbin 150080, China}
\address[3]{School of Materials Science and Engineering, Harbin University of Science and Technology, Harbin 150080, China}
\tfootnote{This work was supported in part by the National Natural Science Foundation of China under Grant 51671075 and Grant 51971086, in part by the State Key Laboratory of Solidification Processing in NWPU under Grant SKLSP201606, in part by the Heilongjiang Postdoctoral Fund for scientific research initiation under Grant LBHQ16118, and in part by the Fundamental Research Foundation for Universities of Heilongjiang Province under Grant LGYC2018JC004.}

\markboth
{Author \headeretal: Preparation of Papers for IEEE TRANSACTIONS and JOURNALS}
{Author \headeretal: Preparation of Papers for IEEE TRANSACTIONS and JOURNALS}

\corresp{Corresponding author: Shu Li (lishu@hrbust.edu.cn); Second Correspondence: Fuxiang Li(lifx2013@163.)}

\begin{abstract}
Density peaks clustering has become a nova of clustering algorithm because of its simplicity and practicality. However, there is one main drawback: it is time-consuming due to its high computational complexity. Herein, a density peaks clustering algorithm with sparse search and K-d tree is developed to solve this problem. Firstly, a sparse distance matrix is calculated by using K-d tree to replace the original full rank distance matrix, so as to accelerate the calculation of local density. Secondly, a sparse search strategy is proposed to accelerate the computation of relative-separation with the intersection between the set of $k$ nearest neighbors and the set consisting of the data points with larger local density for any data point. Furthermore, a second-order difference method for decision values is adopted to determine the cluster centers adaptively. Finally, experiments are carried out on datasets with different distribution characteristics, by comparing with other six state-of-the-art clustering algorithms. It is proved that the algorithm can effectively reduce the computational complexity of the original DPC from $O(n^2K)$ to $O(n(n^{1-1/K}+k))$. Especially for larger datasets, the efficiency is elevated more remarkably. Moreover, the clustering accuracy is also improved to a certain extent. Therefore, it can be concluded that the overall performance of the newly proposed algorithm is excellent.
\end{abstract}


\titlepgskip=-15pt

\maketitle

\section{Introduction}
\label{sec:introduction}
\PARstart{C}{luster} analysis as an important exploration technology of data mining, is committed to reveal the inherent attributes and laws hidden behind the seemingly disorganized unknown data \cite{b1,b2,b3}. It provides support for decision-making and has been successfully applied in many fields such as image pattern recognition, social network mining, market statistical analysis, medical research and engineering systems \cite{b4,b5,b6,b7,b8,b9,b10,b11,b12,b13}. With the extremely strong penetration and rapid development of the Internet, many professional fields are faced with explosive growth of data storage. This leads to high computational complexity and difficulty in mining valuable information. In 2014, \emph{Science} published clustering by fast search and find of density peaks (DPC) \cite{b14}. Due to its novel design idea and robust performance, DPC instantly became the topic center of scholars in related fields. Compared with classical clustering algorithms \cite{b15},\cite{b16}, DPC possesses several advantages. Firstly, the cluster centers can be identified directly through the decision graph, which consists of local density and relative-separation for all data. Secondly, it can handle non-convex datasets well and no iterative process is required. Furthermore, it is insensitive to outliers. However, there are still some shortages for DPC to be further improved, including the sensitivity to cut-off distance, the high computational complexity as well as the determination of cluster centers manually.

For the original DPC algorithm published by \emph{Science} [14], the local structure of data \cite{b17} is not considered and it is sensitive to the cut-off distance. Some scholars try to solve this problem by integrating the $k$ nearest neighbors with DPC in various ways. Du et al. \cite{b18} introduced the idea of $k$ nearest neighbors into DPC and redefined the calculation method of local density. Thus, the so-called DPC-KNN algorithm was successfully established. Xie et al. \cite{b19} also gave another measure of local density using $k$ nearest neighbors and designed a new allocation strategy for non-center points. In order to elevate the possibility of selecting the correct cluster centers, Liu et al. \cite{b20} modified the distance calculation method by considering the distance factors and neighbor information simultaneously. In addition, a two-step allocation strategy was also adopted to allocate non-center points. Recently, Liu et al. \cite{b21} have suggested a mixed density clustering method by defining two types of local density. One is based on $k$ nearest neighbors, while the other is determined by local spatial position deviation. Local density description derived from $k$ nearest neighbors supplies a novel way for DPC algorithm. Under this condition, the sensitivity to the hyper-parameter $k$, i.e. the number of nearest neighbors, is obviously depressed, compared with the sensitivity to cut-off distance used in the original DPC algorithm \cite{b14}. However, for these DPC algorithms based on $k$ nearest neighbors, mentioned-above, searching neighbors mainly resorts to violent means. Therefore, these algorithms still have the same high computational complexity as original DPC.

For DPC algorithms, the high computational complexity commonly comes from the distance calculation, which is related with any two points among all data points. With the increase of dataset size, as groundwork, calculating distance between any two points is time-consuming and even impossible for DPC to be implemented. In order to reduce the computational complexity and lift the clustering efficiency, some improvement strategies have been proposed. Gong et al. developed the efficient distributed density peaks clustering algorithm (EDDPC) \cite{b22}, which eliminates unnecessary distance calculation and data shuffling by Voronoi diagram, data replication and data filtering. Bai et al. \cite{b23} tried to save a part of the computational effort required for distance by combining DPC and K-means. Xu et al. introduced the idea of replacing all data points with non-empty grid into DPC \cite{b24}, \cite{b25}. Two prescreening strategies were proposed to determine cluster centers by screening points with the feature of higher local density. Xu et al. \cite{b26} also proposed a fast density peaks clustering algorithm with sparse search (FSDPC), which introduced the idea of third-party random points to measure the distance. These optimization algorithms have shown good ability in reducing complexity and improve clustering efficiency. However, they either sacrifice accuracy, resulting in poor clustering effect, or decrease the clustering stability, making the clustering result for each run vary obviously. Therefore, it is not satisfying for these algorithms to get reliable and reasonable clustering results.

As for determining cluster centers, it is difficult for the original DPC algorithm to identify the correct cluster centers manually, if the dividing line between the center points and the non-center points on the decision graph is not very clear. Tong et al. \cite{b27} obtained the initial clusters by a designed pre-clustering method, and then gave the final clusters according to the Scale Space Theory. Lv et al. \cite{b28} proposed a method to determine the cluster centers automatically according to the decision value defined by the product of local density and relative-separation. Flores et al. \cite{b29} proposed a strategy to find cluster centers adaptively by searching the gaps among data points on the one-dimensional decision graph mapped. Lin et al. \cite{b30} introduced a hyper-parameter, neighbor radius, to select a group of possible density peaks, as preliminary clustering results. Then the final clustering results were formed by merging the clusters with single-bond clustering method. However, some of these algorithms reached the goal of adaptive determination of cluster centers, at the cost of introducing an additional hyper-parameter. Meanwhile, some algorithms became more time-consuming and inefficient, especially for large or complex dataset.

To sum up, various DPC algorithms have been proposed, but very few of these algorithms could achieve satisfactory clustering accuracy, computational complexity as well as the adaptive determination of cluster centers simultaneously. Therefore, the main purpose of the present work is to develop an improved DPC algorithm, in which the computational complexity is reduced significantly and determining cluster centers could be carried out automatically, under the condition that the clustering accuracy is guaranteed without any additional hyper-parameter introduced. As a result, a density peaks clustering algorithm with sparse search and K-d tree (SKTDPC) was proposed. As a type of data structure, K-d tree \cite{b31} was adopted to find $k$ nearest neighbors and thus the computational complexity of local density could be firstly reduced remarkably. Secondly, a strategy of sparse search was proposed to accelerate the calculation of relative-separation significantly. Furthermore, the method for automatic determination of cluster centers was described based on second-order difference of the decision value. Besides the main work, the applicability of SKTDPC algorithm is analyzed briefly on higher dimensional datasets.

The rest of this paper is organized as follows. Section \uppercase\expandafter{\romannumeral2} reviews the works related with DPC algorithms. Section \uppercase\expandafter{\romannumeral3} depicts the SKTDPC algorithm newly developed, including fast search for $k$ nearest neighbors, dual acceleration for local density and relative-separation, and determining cluster centers adaptively etc. Section \uppercase\expandafter{\romannumeral4} verifies the validity of the SKTDPC algorithm and makes a comparison with other state-of-the-art clustering algorithms thoroughly on various datasets. Finally, Section \uppercase\expandafter{\romannumeral5} is the concluding remarks of the paper.

\section{RELATED WORKS}
Due to the simple and clear principle of DPC algorithm, it shows great performance on dataset with any shape or dimension. DPC has become a new favorite in the field of clustering algorithm research. The currently proposed SKTDPC algorithm is designed on the basis of the original DPC, the idea of $k$ nearest neighbors and sparse search. Therefore, this section briefly reviews the original DPC algorithm. The role of the idea for $k$ nearest neighbors and sparse search in DPC are also analyzed.

\subsection{THE ORIGINAL DPC ALGORITHM}
In original DPC algorithm \cite{b14}, there are two important variables, the local density $\rho_i$ and the relative-separation $\delta_i$ for any data point $x_i$ in dataset labeled by $S=\{x_1,x_2,\ldots,x_n\}$ with $n$ data.

One of the important variables, the local density 
$\rho_i$, is defined as follows:
\begin{equation}
\rho_i=\sum_{j\neq i} \mathcal{X}(d_{i,j}-d_c),  ~\mathcal{X}(a)=\begin{cases}
1,& \text{$a<0$}\\
0,& \text{$a\geq 0$}
\end{cases} \label{eq1}
\end{equation}
where $d_c$ is the cut-off distance, which is regarded as a hyper-parameter for the algorithm, and $d_{i,j}$ is the Euclidean distance between data points $x_i$ and $x_j$. 

Another important variable, the relative-separation $\delta_i$, could be determined by searching for the nearest data point $x_j$, which has a relatively larger local density compared with $x_i$:
\begin{equation}
\delta_i=\min_{j:\rho_j > \rho_i}(d_{i,j})\label{eq2}
\end{equation}
This is the condition that the point $x_i$ is a non-maximum local density point. Especially, for the data point with the maximum local density, the relative-separation $\delta_i$ is marked specifically by $\delta_{\text{max}}$ and is given as:
\begin{equation}
\delta_{\text{max}}=\max_{j}(d_{i,j})\label{eq3}
\end{equation}

DPC selects cluster centers based on the core idea, that the cluster centers are surrounded by these data points with relatively lower $\rho_i$ as well as the cluster centers possess relatively larger distance from points with local density higher than the cluster centers. That is, the data points $x_i$ with relatively large $\rho_i$ and $\delta_i$ have a higher probability of being identified as cluster centers. This is the main characteristic of the original DPC algorithm different from other density clustering algorithms.  

Based on the density $\rho_i$ and distance $\delta_i$, two critical tasks are performed by the DPC. Firstly, cluster centers are determined. By drawing decision graph in the two-dimensional coordinate system with $\rho_i$ and $\delta_i$ as the abscissa and ordinate, respectively, each data point corresponds to a position in the graph and these points in the upper right corner of the decision graph are selected as cluster centers \cite{b32,b33,b34}. Secondly, non-cluster center points are assigned \cite{b35}, \cite{b36}. The allocation principle is that each non-center point and its nearest point with a higher local density have the same cluster. For the determination of cluster centers, if the boundary between the center points and the non-center points on the decision graph is obvious, the cluster centers could be identified quickly. Otherwise, it is not easy to distinguish the ideal cluster centers manually. This is the problem for determining cluster centers adaptively, which will be solved in the Section \uppercase\expandafter{\romannumeral3}.D.

For the computational complexity of the original DPC, it depends on the Euclidean distance calculation between any two data points in dataset $S$  \cite{b37}. These distances construct the symmetric full-rank matrix $D$:
\begin{equation}\label{eq5}
D=\bigg [d_{i,j}\bigg ]_{n\times n}
\end{equation}
This is groundwork for further determine both the local density $\rho_i$ and the relative-separation $\delta_i$ of each data point $x_i$. For the dataset $S=\{x_1,x_2,\ldots,x_n\}$ with $n$ data, the computational complexity of DPC is mainly determined by two parts together. One is the computational complexity required to compute the distance matrix, which is $O(n^2)$, while another is the complexity required to allocate the remaining points, which approximately is $O(n)$. Then, the overall computational complexity of DPC is $O(n^2)$.

\subsection{THE IDEA OF K NEAREST NEIGHBORS AND SPARSE SEARCH}
From Equation \eqref{eq1}, we can see that for the data point $x_i$ the local density $\rho_i$ is defined by the number of the nearest neighbors which are located in the circular area with the cut-off distance $d_c$ as the radius. Thus, Equation \eqref{eq1} can also be reformulated by searching for the set $\text{NN}(x_i)$ which consists of the nearest neighbors of the data point $x_i$:
\begin{equation}\label{eq4}
\begin{aligned}
  &\text{NN}(x_i)=\{x_j|d_{i,j}<d_c, j\neq i\}    \\\hfill
  &\rho_i=|\text{NN}(x_i)|\hfill
\end{aligned}
\end{equation}
where $|\text{NN}(x_i)|$ indicates the number of elements in the set $\text{NN}(x_i)$.

In order to achieve better clustering results, for the original DPC algorithm to be implemented, the cut-off distance $d_c$ should be adjusted as a hyper-parameter. This method defining the local density $\rho_i$ ignores the local distance information of data points and makes the algorithm more sensitive to hyper-parameter. For solving this problem, inspired by the idea of $k$ nearest neighbors, the hyper-parameter $d_c$ could be replaced by the hyper-parameter $k$, the number of the nearest neighbors. Some scholars have proposed various forms to define the local density $\rho_i$ with $k$ nearest neighbors. In this way, not only the algorithmic sensitivity to hyper-parameter can be depressed, but also the local distance information can be integrated into the local density definition to describe the data density more reasonably. More specifically, $d_c$ has a wide range of adjustment and has a great impact on algorithm performance. On the contrary, the value of $k$ increases with a step size of 1, and the algorithm can achieve a satisfactory state relatively when $k$ does not exceed 20 generally. Therefore, this strategy to define the local density is still maintained by the present work, but a type of data structure, the K-d tree is adopted to accelerate search for the $k$ nearest neighbors. This will be depicted in Sections III.A and III.B.

It can be found that the calculation of the distance between any two points among all data points is time-consuming at the beginning for DPC algorithms to be implemented. Thus, DPC algorithms have high computational complexity, no matter how to define $\rho_i$, with the hyper-parameter $d_c$ or the hyper-parameter $k$, discussed in previous section. In view of this, some scholars have introduced the idea of sparse search into DPC algorithm to reduce the complexity. As the name implies, sparse search refers to a class of methods that can reduce internal friction effectively by replacing a large number of calculations with as little computation as possible. \cite{b22,b23,b24,b25,b26} proposed different sparse methods to speed up the calculation process of DPC. Inspired by this idea, we design a new and effective sparse search strategy to reduce the complexity of DPC.

It should be stressed that the present analysis on computational complexity does not take into account the influence of feature dimension of data, which will be discussed in the Section III.F. In the next section, an algorithm called SKTDPC will be developed, which can significantly reduce the computational complexity by the important strategies of sparse search and K-d tree.

\section{THE DENSITY PEAKS CLUSTERING ALGORITHM WITH SPARSE SEARCH AND K-D TREE}
In this section, the density peaks clustering algorithm with sparse search and K-d tree (SKTDPC) is proposed. Aiming to reduce the computational complexity, the present algorithm adopts K-d tree to find $k$ nearest neighbors quickly and uses a strategy of sparse search to accelerate the calculation of relative-separation significantly. Moreover, a method for automatic determination of cluster centers is developed. Finally, the process of the algorithm and the complexity are depicted and discussed, respectively.

\subsection{FAST SEARCH OF K NEAREST NEIGHBORS BASED ON K-D TREE}
As discussed above, introducing the hyper-parameter $k$, the number of the nearest neighbors, to replace the hyper-parameter $d_c$, the cut-off distance, could reduce the algorithmic sensitivity to hyper-parameter obviously. However, for the $k$ nearest neighbors of each data point to be found out, the distance between any two points in dataset $S$ should also be calculated. Therefore, there is still the problem of high computational complexity. To accelerate the search for $k$ nearest neighbors and save a large part of the computational effort required for distance, as a type of data structure, the classical K-d tree is adopted in the present work.

K-d tree is a typical binary tree, which stores data points in $K$ dimensional space for quick retrieval. It represents the division of $K$ dimensional space and constitutes a series of $K$ dimensional hyperrectangular regions. The fast determination of $k$ nearest neighbors based on K-d tree is divided into two steps. The first part is the K-d tree construction process, the second part is the process of search for $k$ nearest neighbors with K-d tree, the specific process described by Algorithm 1 is as follows.

 \begin{algorithm*}\label{alg1}
            \caption{The K-d tree determines for $k$ nearest neighbors.}
            \KwIn{The dataset $S=\{x_1,x_2,\ldots,x_n\}$, where $x_i=(x_i^{(1)},x_i^{(2)},\ldots,x_i^{(K)})^T, i=1,2,\ldots,n$; The number of the nearest neighbors $k$}
            \KwOut{K-d tree; The set $Q$ of $k$ nearest neighbors of $x_{\text{target}}$(here $x_{\text{target}}$ is any point $x_i$ in dataset $S$) }
            
            \textbf{build the K-d tree}\;
            \While{any leaf node has non-unique data}{
for each dimension $h(h=1,2,\ldots,K)$ get its variance\; 
choose $h$ according to the maximum variance\;
sort the data values in dimension $h$ and get its median $m$ as the segmentation point\;
store the data point of segmentation on the corresponding node\;
other data are divided by $m$ into two subsets (The left sub-node corresponds to the subset of the coordinates $x^{(h)}$ which is equal to or less than the segmentation point. The right sub-node corresponds to the  subset of coordinates $x^{(h)}$ greater than the segmentation point)\;            
        }
\textbf{Search $k$ nearest neighbors based on constructed K-d tree}\;
Build an empty set $Q$\;
Recursive visit down K-d tree until leaf node is reached. At this point, the leaf node is the current node $(x_{\text{current}})$ and it is marked as visited\;				
\Loop{}{
                calculate the distance $d(x_{\text{target}},x_{\text{current}})$\;
add the $x_{\text{current}}$ to $Q$\;
                \eIf{$|Q|<k$, $|Q|$ indicates the number of elements in the set $Q$}
                {
                        \eIf{there is another child node in the current node}
{recursive visit down K-d tree until leaf node is reached and marked it as visited\;
update the leaf node to the current node\;
}
{find a parent node which has not been visited and mark it as visited\;
update the parent node to the current node\;

}
                }
                {
                    break~ \textbf{loop}\;
                }
        }
\eIf{the current node is leaf node }{

\While{the current node has parent node }{
\textbf{do Block A:}\;
find a parent node which has not been visited and mark it as visited\;
update the parent node to the current node\;
calculate the distance $d(x_{\text{target}},x_{\text{current}})$\;
\If{$d(x_{\rm{target}},x_{\rm{current}})<Q_{\rm max}$, $Q_{\text max}$ is the farthest distance from $x_{\text{target}}$ in $Q$}{replace the point farthest from $x_{\rm{target}}$ in $Q$ with the current node\;}
\textbf{do Block B:}\;
\If{there is another child node in the current node}{
calculate the distance $d(x_{\text{target}},l_{\text{cur-tan}})$, $l_{\text{cur-tan}}$ is the tangent line of the current node\;
\If{$d(x_{\rm{target}},x_{\rm{current}})<Q_{\rm max}$}{
recursive visit down K-d tree until leaf node is reached and marked it as visited\;
update the leaf node to the current node\;
calculate the distance $d(x_{\text{target}},x_{\text{current}})$\;
\If{$d(x_{\rm{target}},x_{\rm{current}})<Q_{\rm max}$}{

replace the point farthest from $x_{\text{target}}$ in $Q$ with the current node\;
}

}

}

}

}
{\While{any leaf node has non-unique data}{
\textbf{do block B}\;
\textbf{do block A}\;

}

}
\end{algorithm*}

Note that in order to reduce the cost of backtracking and achieve the best balance of data segmentation, this algorithm selects the dimension with the largest variance and the median in this dimension as dividing criterion each time. The larger the variance is, the more scattered the data points are in this dimension. In the whole paper, the dataset $S$ takes the form of $S=\{x_1,x_2,\ldots,x_n\}\in\mathcal{R}^{n\times K}$, where $n$ and $K$ are the number and spatial dimension of data points respectively. In addition, $k$ represents the number of the nearest neighbors for any data point $x_i$. 

The computational complexity of K-d tree construction is $O(nK\text{log}n)$, and the computational complexity of searching $k$ nearest neighbors of  $n$ data points is $O(n(n^{1-1/K}+k))$ in Algorithm 1. Therefore, the overall computational complexity of searching $k$ nearest neighbors by K-d tree is the larger one from $O(nK\text{log}n)$ and $O(n(n^{1-1/K}+k))$, that is, $O(n(n^{1-1/K}+k))$, $K<<n$. Based on K-d tree, the original computational complexity of searching k nearest neighbors is significantly reduced from $O(n^2)$ to $O(n(n^{1-1/K}+k))$. Thus, a lot of unnecessary calculations for distance are saved. This will further result in the acceleration for calculation of the local density $\rho_i$ (Section III.B).

\subsection{THE ACCELERATION FOR CALCULATION OF THE LOCAL DENSITY BY K-D TREE}
Searching for $k$ nearest neighbors is the groundwork for DPC algorithms to be implemented. By using the K-d tree method proposed in Section III.A, the computational complexity could be reduced essentially. As a result, the distance matrix $D$ used in the original DPC (Section II.A), a symmetric full-rank matrix, is changed to the symmetric sparse matrix $\widetilde{D}$. For example, assuming there is a dataset $S=\{x_1,x_2,\ldots,x_{12}\}$ containing 12 elements, the form of the sparse matrix $\widetilde{D}$ may be as follows:
\begin{equation*}
\setlength{\arraycolsep}{0.05pt}
\widetilde{D}\hspace{-1mm}=\hspace{-2mm} \fontsize{6}{4} \selectfont\left[\begin{array}{cccccccccccc}
0&\,d_{1,2}&\,	\text{[]}&\,\text{[]}&\,d_{1,5}&\,d_{1,6}&\,\text{[]}&\,\text{[]}&\, \text{[]}&\,d_{1,10}&\,\text{[]}&\,\text{[]}\\
d_{1,2}&\,0&\,\text{[]}&\,\text{[]}&\,\text{[]}&\,d_{2,6}&\,\text{[]}&\,d_{2,8},\text{[]}&\,\text{[]}&\,\text{[]}&\,\text{[]}\\
\text{[]}&\,\text{[]}&\,0&\,d_{3,4}&\,\text{[]}&\,\text{[]}&\,d_{3,7}&\,\text{[]}&\,\text{[]}&\,\text{[]}&\,d_{3,11}&\,\text{[]}\\
\text{[]}&\,\text{[]}&\,d_{3,4}&\,0&\,d_{4,5}&\,\text{[]}&\,\text{[]}&\,\text{[]}&\,d_{4,9}&\,\text{[]}&\,\text{[]}&\, d_{4,12}\\
d_{1,5}&\,\text{[]}&\,\text{[]}&\,d_{4,5}&\,0&\,\text{[]}&\,\text{[]}&\,\text{[]}&\,d_{5,9}&\,\text{[]}&\,\text{[]}&\,\text{[]}\\
d_{1,6}&\,d_{2,6}&\,\text{[]}&\,\text{[]}&\,\text{[]}&\,0&\,\text{[]}&\,d_{6,8}&\,\text{[]}&\,\text{[]}&\,d_{6,11}&\,\text{[]}\\
\text{[]}&\,\text{[]}&\,d_{3,7}&\,\text{[]}&\,\text{[]}&\,\text{[]}&\,0&\,\text{[]}&\,\text{[]}&\,d_{7,10}&\,\text{[]}&\,d_{7,12}\\
\text{[]}&\,d_{2,8}&\,\text{[]}&\,\text{[]}&\,\text{[]}&\,d_{6,8}&\,\text{[]}&\,0&\,\text{[]}&\,d_{8,10}&\,\text{[]}&\,\text{[]}\\
\text{[]}&\,\text{[]}&\,\text{[]}&\,d_{4,9}&\,d_{5,9}&\,\text{[]}&\,\text{[]}&\,\text{[]}&\,0&\,\text{[]}&\,d_{9,11}&\,\text{[]}\\
d_{1,10}&\,\text{[]}&\,\text{[]}&\,\text{[]}&\,\text{[]}&\,\text{[]}&\,d_{7,10}&\,d_{8,10}&\,\text{[]}&\,0&\,\text{[]}&\,d_{10,12}\\
\text{[]}&\,\text{[]}&\,d_{3,11}&\,\text{[]}&\,\text{[]}&\,d_{6,11}&\,\text{[]}&\,\text{[]}&\,d_{9,11}&\,\text{[]}&\,0&\,\text{[]}\\
\text{[]}&\,\text{[]}&\,\text{[]}&\,d_{4,12}&\,\text{[]}&\,\text{[]}&\,d_{7,12}&\,\text{[]}&\,\text{[]}&\,d_{10,12}&\,\text{[]}&\,0\\
\end{array}\right]
\end{equation*}
where $d_{i,j}$ represents the Euclidean distance between the $i$-th data point and the $j$-th data point calculated by K-d tree in the process of searching the $k$ nearest neighbors; ``[]'' denotes the distance that does not need to be calculated in this process (also known as invalid distance). In this matrix, the first row implies that the algorithm can find out the $k$ nearest neighbors of the data point $x_1$ only by calculating the distance between $x_1$ and $x_2$, $x_5$,  $x_6$, $x_{10}$, instead of calculating all the distances.

Taking the number $k$ of the nearest neighbors for any data point $x_i$ as a hyper-parameter, the set of the $k$ nearest neighbors for $x_i$, calculated by K-d tree, is marked by $\text{NN}_{k}(x_i)$, which could be described as:
\begin{equation}\label{eq6}
\resizebox{.9\hsize}{!}{$\text{NN}_{k}(x_i)=\{x_j\in S|d(x_i,x_j)\leq  d(x_i,N_{k}(x_i)), j\neq i\}$}
\end{equation}

where $d(x_i,x_j)$ is the distance between $x_i$ and $x_j$, which is equivalent to $d_{ij}$, and $N_{k}(x_i)$ is the $k$-th nearest neighbor for point $x_i$.

Thus, the following formula is proposed to describe the local density $\rho_i$ more reasonably:
\begin{equation}\label{eq7}
\rho_i=\frac{1}{\sum_{x_j\in \text{NN}_{k}(x_i)}d(x_i,x_j)}
\end{equation}
It is indicated that the local density $\rho_i$ is not required to calculate on the basis of the symmetric full-rank matrix $D$, which includes all distances between any two data in dataset $S$. Therefore, by introducing K-d tree method, the calculation of the local density $\rho_i$ is accelerated significantly with the sparse matrix $\widetilde{D}$. This is the first acceleration for SKTDPC algorithm.

\subsection{THE ACCELERATION FOR CALCULATION OF THE RELATIVE-SEPARATION BY SPARSE SEARCH}
In Section III.B, $\rho_i$ is obtained based on the symmetric sparse matrix $\widetilde{D}$. According to the definition of relative-separation, Equation \eqref{eq2}, $\delta_i$ of the non-maximum density point is determined by searching for the nearest data point $x_j$, which possesses relatively larger local density $\rho_j$ compared with $x_i(\rho_j>\rho_i)$. In other words, the distance information in $\widetilde{D}$ is not enough to support the acquisition of all $\delta_i$, and it is necessary to complete the rest of the distance calculation. At this point, the computational complexity of getting $\delta_i$ will reach $O(n^2)$ in the worst case. Thus, it would be insignificant to save a large part of the distance calculation for $\rho_i$ by K-d tree in Section III.B, if all of distance calculations were carried out between any two points in dataset $S$. For dealing with this problem, a sparse search strategy is proposed to accelerate the calculation of the relative-separation $\delta_i$.

For expressing more clearly, the local density $\rho_i$ for all of data points obtained in Section III.B are sorted in descending order to generate a new sequence ${\rho_i}^{*}(i=1,2,\ldots,n)$. For any point $x_i$, set $B(x_i)$ is introduced to represent the set including all points $x_j$ with greater local density $\rho_j$ compared with $x_i(\rho_j>\rho_i)$. That is to say,   consists of these data points corresponding to ${\rho_1}^{*}$ to ${\rho_{i-1}}^{*}$. Thus, the set $B(x_i)$ is explicit. In addition, for any point $x_i$, the set of $k$ nearest neighbors, $\text{NN}_k(x_i)$, has also been determined in Section III.B. Then, for $x_i$ with non-maximum density, based on $\text{NN}_k(x_i)$ and $B(x_i)$, if $\exists j~ s.t.~\rho_j>\rho_i$, the sparse search strategy for $\delta_i$ is defined as:


\begin{small}
\begin{equation}\label{eq8}
\begin{split}
\delta_i\hspace{-1mm}=\hspace{-1mm}\begin{cases}
\min\limits_{x_j\in {\text{NN}}_k(x_i)\cap B(x_i)}d(x_i,x_j),
&\hspace{-3mm}\text{when $\text{NN}_k(x_i)\hspace{-1mm}\cap\hspace{-1mm} B(x_i)\hspace{-1mm}\neq\hspace{-1mm} \phi$}\\
\qquad\quad\min d(x_i,x_j), 
&\hspace{-3mm}\text{otherwise}
\end{cases}
\end{split}
\end{equation}
\end{small}

When $x_i$ is the point with maximum density, $\delta_{\text{max}}$ has the same expression as that defined by the original DPC, Equation \eqref{eq3}.

Equation \eqref{eq8} indicates that both the sets $\text{NN}_k(x_i)$ and $B(x_i)$ are important for determining $\delta_i$ quickly. One is the set of $k$ nearest neighbors, while the other is the set of these points with greater local density than $\rho_i$. When the two sets have an intersection, it means that $\delta_i$ must be the minimum of distances of $x_i$ from the point $x_j$ in the intersection. All these distance values are known information determined by looking for $k$ nearest neighbors with K-d tree. They are stored in the sparse matrix $\widetilde{D}$. In this case, $\delta_i$ can be obtained directly, without any additional calculation for distance. Under the condition that there is no intersection, the unknown distances between $x_i$ and points with local density greater than $\rho_i$ need to be calculated to find the nearest distance. Therefore, whether the intersection is non-empty or not plays an important role in determining the computational complexity. 

Normally, most of $\delta_i$ could be obtained by the non-empty intersection directly. This can be explained as follows. Here all data points in dataset $S$ are divided into four parts: points with large $\rho_i$ and large $\delta_i$ (part 1), points with large $\rho_i$ and small $\delta_i$(part 2), points with small $\rho_i$ and small $\delta_i$ (part 3) as well as points with small $\rho_i$ and large $\delta_i$ (part 4). Firstly, there are very few data points with large $\rho_i$ and large $\delta_i$, compared with total data. These points are potential cluster centers. Secondly, for points with large $\rho_i$, including part 1 and part 2, the total amount of elements in the set $B(x_i)$ is relatively very small, since $B(x_i)$ is defined by these points with local density greater than $\rho_i$ and $\rho_i$ itself is also large. In addition, for the small amount of distance calculation, some of the known distances have already been stored in the symmetric sparse matrix $\widetilde D$, except for the distances between $x_i$ and the $k$ nearest neighbors. Thus, even though the intersection is empty, the distance calculation, which needs to be supplemented, is not time-consuming at all. Thirdly, the points in part 3 take a vast proportion of total data. They are non-center points with low local density $\rho_i$. Small $\rho_i$ implies that the averaged distance between $x_i$ and the $k$ nearest neighbors is obviously large, i.e. the $k$ nearest neighbors are not concentrated around $x_i$, but are decentralized. Further, the $k$-th nearest neighbor of $x_i$ is relatively far away from $x_i$. Meanwhile, small $\delta_i$ indicates that the shortest distance from $x_i$ is small for the points $x_j$ with larger local density $\rho_j$. Thus, for any point $x_i$, there is a high possibility that the point $x_j$, which determines the value of $\delta_i$, (with larger local density $\rho_j$ and minimum of distance from $x_i$) is located in the nearest neighbors set $\text{NN}_k(x_i)$. That is to say the intersection of $\text{NN}_k(x_i)$ and $B(x_i)$ is non-empty very likely. Finally, the points in part 4 are commonly regarded as outliers, which are also very few. Corresponding calculation required for distance are even negligible actually. 

\begin{figure*}[ht]
\centering
	\subfloat{\includegraphics[width = 0.9\textwidth]{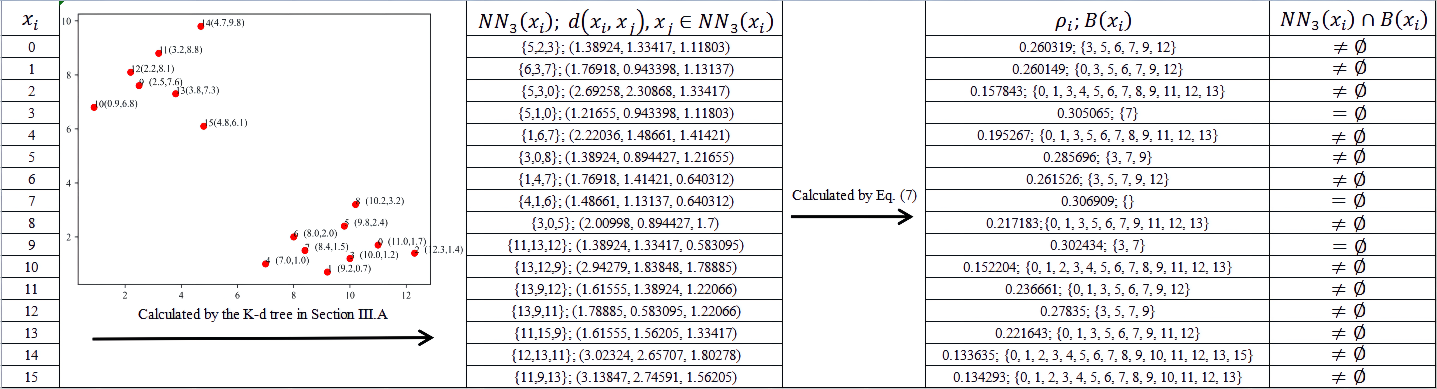}}
\caption{\textbf {Example demonstration of sparse search strategy for obtaining $\delta_i$.} }
\label{fig:labe0}
\end{figure*}

In order to show the process of obtaining  $\delta_i$  more clearly and intuitively by the proposed sparse search strategy, an example is used to confirm the fact that the intersection of  $\text{NN}_k(x_i)$ and $B(x_i)$  is not empty in most cases, as shown in Fig. \ref{fig:labe0}. There is a dataset with 16 data points that can be seen in Fig. \ref{fig:labe0}. Firstly, the 3 nearest neighbors set  $\text{NN}_3(x_i)$ of any sample point  $x_i$ and the distance $d(x_i,x_j), x_j\in {\text{NN}}_3(x_i)$  are obtained by the sparse calculation method of K-d tree in Section III.A. Then, the $\rho_i$  is obtained by using \eqref{eq7}, and the larger density set  $B(x_i)$ of the sample point $x_i$  can be obtained. Finally, the $\delta_i$  is obtained from the $\text{NN}_3(x_i)\cap B(x_i)$  shown in the last column in Fig. \ref{fig:labe0}. From Fig. \ref{fig:labe0}, we can clearly see that the $\text{NN}_3(x_i)\cap B(x_i)$  of sample points 0, 1, 2, 4, 5, 6, 8, 10, 11, 12, 13, 14 and 15 is not empty. This means that the $\delta_i$  of these data points can be obtained directly from the  $\text{NN}_3(x_i)\cap B(x_i)$. That is, the nearest distance from  $x_i$ in the $\text{NN}_3(x_i)\cap B(x_i)$  is $\delta_i$, without any additional calculation. Only the $\text{NN}_3(x_i)\cap B(x_i)$  of 3, 7 and 9 three points is empty, so we need to add a little bit of calculation. It is worth noting that although these three points require supplementary calculation, 3 and 9 only needs to be calculated with a very small number of points in the set \{7\}, \{3, 7\} respectively. Only point 7 with maximum density needs to calculate the distance from the remaining points. It can be seen that the proposed sparse search strategy effectively avoids the distance calculation of low-density points, which account for most of the data volume. For a very small proportion of high-density points, even if a little additional calculation is needed, the amount of calculation is very small. For the whole algorithm, it is almost negligible.

On the whole, a very small amount of extra distance needs to be calculated to obtain $\delta_i$ of all data points. The sparse search strategy captures the crux of these low density points which occupy the main amount of computation, and obtains $\delta_i$ through ingenious intersection strategy directly, so that a lot of distance calculation is saved. At this point, we have succeeded in reducing the computational complexity of this step to much less than $O(n(n^{1-1/K}+k))$. The second acceleration of the SKTDPC algorithm has been achieved. Therefore, the problem of high computational complexity is solved fundamentally, for the original DPC algorithm as well as a series of extended DPC algorithms based on $k$ nearest neighbors, by a simpler and more efficient strategy.

\subsection{ADAPTIVE DETERMINATION OF CLUSTER CENTERS}
For the determination of cluster centers, it is difficult for the original DPC algorithm to distinguish center points and non-center points on the decision graph manually, if the boundary is not very clear. A reasonable adaptive way can deal with this problem. In this section, a simple and efficient method is proposed.

As an important judgment basis, the variable, called decision value $\gamma$, is also introduced, which is defined by the product of $\rho_i$ and $\delta_i$:
\begin{equation}\label{eq9}
\gamma_i=\rho_i \times \delta_i
\end{equation}
By re-arrangement in descending order of $\gamma$ value, the newly generated sequence of decision value is marked by $\gamma^{*}$. As analyzed above, the data points with both large $\rho_i$ and large $\delta_i$ are potential to become cluster centers. Under this condition, the decision value is also large. Thus, it is crucial to determine adaptively the boundary between the center points and the non-center points in $\gamma^{*}$.

\begin{figure*}[ht]
\centering
	\subfloat[Distribution of data points in dataset SS2.]{\includegraphics[width = 0.5\textwidth]{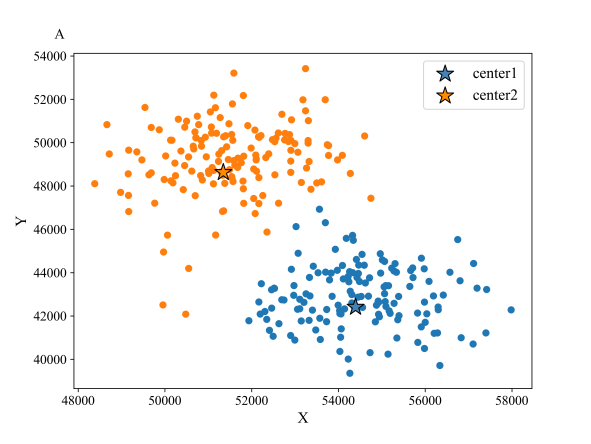}\centering}
	\hfill
	\subfloat[The data points are arranged in order $\gamma^*$ from the largest to the smallest.]{\includegraphics[width = 0.5\textwidth]{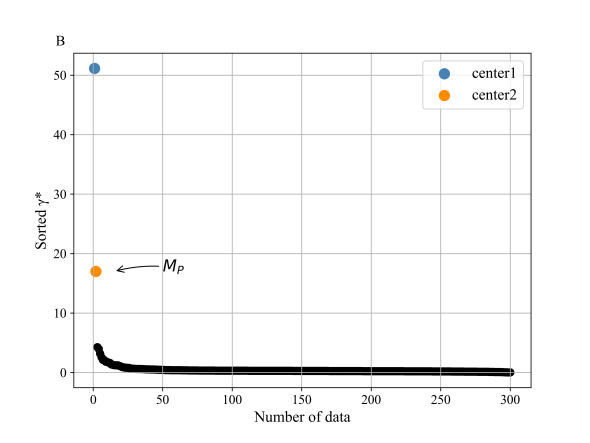}}
	\hfill
\caption{\textbf {A simple two-dimensional synthetic dataset as the example. }}
\label{fig:label}
\end{figure*}

Due to the essential distinction between the center points and the non-center points in the degree of change of decision value, the location of the mutation-point is determined adaptively according to this key feature to lock the cluster centers. For $\gamma^{*}$ sequence, there is the relatively large fluctuation, for the center points with relatively large decision value. In contrast, the fluctuation of $\gamma^{*}$ value is not obvious for the non-center points with relatively small decision value. The second-order difference for $\gamma^{*}$ value is used to describe this fluctuation.

When searching the cluster centers, it is necessary to narrow the search appropriately. That is to say, the points in the front position of $\gamma^{*}$ with relatively large $\gamma$ value that may become the cluster centers needs to be focused on. Because the points with small $\gamma$ value does not have the characteristics of becoming the cluster centers, either $\rho_i$ or $\delta_{i}$ is small, or both are small. Here, the search range is locked in $[1,\lfloor \sqrt{n} \rfloor]$ preliminarily, where $n$ is the number of data. Previous studies have shown that this search range $[1,\lfloor \sqrt{n} \rfloor]$ is appropriate. In this way, not only the search efficiency can be improved, but also the removal of irrelevant data points will play a positive role in reducing disruption for determining the cluster centers.

Through the analysis of $\gamma^{*}$ value of a large number of datasets with different distributions, it is found that there is a general rule that the $\gamma$ value of the point with the most potential to be the cluster centers, $\gamma_1^{*}$, is much larger than $\gamma_2^{*}$ and the value behind it generally. If the $\gamma_1^{*}$ value is placed within the search range to determine the location of the mutation-point, the difference value of $\gamma_1^{*}$ will be very large, which may lead to the wrong mutation-point being found, thereby affecting the determination of the cluster centers. Thus, the search range $[2,\lfloor \sqrt{n} \rfloor]$ is finally adopted in the present work. This treatment does not affect becoming a cluster center for the data point corresponding $\gamma_1^{*}$. Furthermore, it could avoid possible mistake to distinguish the center points and the non-center points.

Based on the second-order difference of $\gamma^{*}$, the mutation-point is described as:
\begin{equation}\label{eq10}
\begin{split}
M_p\hspace{-1mm}=\hspace{-1mm}\max\bigg\{l|l\hspace{-1mm}=\hspace{-1mm}\arg \max\limits_{i}\bigg[&(\frac{i+1}{i})^2\frac{\xi_i}{\gamma_{\text{max}}^{*}-\gamma_{\text{min}}^{*}},\\
&i\hspace{-1mm}=\hspace{-1mm}2,3,\ldots,\lfloor \sqrt{n} \rfloor\hspace{-1mm}-\hspace{-1mm}2\bigg]\bigg\}
\end{split}
\end{equation}
where the function arg max is used to determine the set of variable points $i$ which maximizes $(\frac{i+1}{i})^2\frac{\xi_i}{\gamma_{\text{max}}^{*}-\gamma_{\text{min}}^{*}}$, $M_p$ is the mutation-point that takes the largest $i$ value when arg max returns multiple $i$ value, $\xi_i$ is the second-order difference for $\gamma_i^{*}$, which is defined by Equation \eqref{eq11}, $\gamma_{\text{min}}^{*}$ and $\gamma_{\text{max}}^{*}$ represent the minimum and maximum values of $\gamma_i^{*}$ in $[2,\lfloor \sqrt{n} \rfloor]$,
\begin{equation}\label{eq11}
\xi_i=\mu_i-\mu_{i+1}, i=2,3,\ldots,\lfloor \sqrt{n} \rfloor-2
\end{equation}
\begin{equation}\label{eq12}
\mu_i=\gamma_i^{*}-\gamma_{i+1}^{*}, i=2,3,\ldots,\lfloor \sqrt{n} \rfloor-1
\end{equation}
where $\mu_i$ is the first-order difference between two adjacent terms $\gamma_i^{*}$. The adjustment coefficient $(\frac{i+1}{i})^2$ is to make $(\frac{i+1}{i})^2\frac{\xi_i}{\gamma_{\text{max}}^{*}-\gamma_{\text{min}}^{*}}$ bigger when $\gamma_i^{*}$ is larger, and smaller when $\gamma_i^{*}$ is smaller. The distinction between the center points and the non-center points is more prominent through the adjustment coefficient. $M_p$ is the mutation-point with the largest ordering value when the second-order difference of $\gamma^{*}$ is mutated. Then all data points with $\gamma_i^{*}$ larger than $\gamma_{M_p}^{*}$, including $\gamma_{M_p}^{*}$, are the candidate cluster centers of searching, namely the corresponding points from $\gamma_1^{*}$ to $\gamma_{M_p}^{*}$. Finally, the pseudo-center points with large $\rho_i$ small $\delta_i$ or small $\rho_i$ large $\delta_i$ were removed from these candidate centers as the final cluster centers. Although the $\gamma_i$ values of these pseudo-center points are large, they do not possess the characteristics of the center points, that is, $\rho_i$ and $\delta_i$ are both large. Specifically, the points $x_i(i\in[1,M_p])$ satisfying both $\rho_i>\frac{\sum^{\lfloor \sqrt{n} \rfloor}_{t=1}\rho_t}{\lfloor \sqrt{n} \rfloor}$ and $\delta_i>\frac{\sum^{\lfloor \sqrt{n} \rfloor}_{t=1}\delta_t}{\lfloor \sqrt{n} \rfloor}$ in the candidate center points are retained, and the remaining pseudo-center points are removed to get the final cluster centers.

Next, a two-dimensional synthetic dataset SS2 is adopted as an example to clearly demonstrate the process of determining $M_p$. This dataset has a total of 300 data points ($n=300$). In other words, the subsequent calculation of second-order difference only needs to focus on the 16 largest $\gamma^{*}$ in the range of $[2,17]$, where $\lfloor \sqrt{n} \rfloor=17$. Based on this information, 15 $\mu_i$ from the 16 largest $\gamma^{*}$ are firstly calculated according to Equation \eqref{eq12}; Then, 14 $\xi_i$ are calculated from 15 $\mu_i$ by Equation \eqref{eq11}; Finally, the mutation-point $M_p$ is determined by Equation \eqref{eq10}, which is $M_p=2$. Here, the two candidate center points meet the above conditions. Therefore, the final cluster centers are the two data points corresponding to $\gamma_1^{*}$ and $\gamma_2^{*}$.

The distribution of data points in dataset SS2 is shown in Fig. \ref{fig:label}(a), and the yellow points indicated by arrows in Fig. \ref{fig:label}(b) are mutation-point determined by the proposed second-order difference method. In other words, the mutation-point and the blue point (center 1 and center 2) were identified as the cluster centers of dataset SS2, and the black dots below were the ordinary non-center points.

\subsection{PROCESS OF SKTDPC}
The overall process of SKTDPC algorithm includes three main parts: calculation of dual acceleration for $\rho_i$ and $\delta_i$ based on K-d tree and sparse search strategy, adaptive determination of cluster centers by second-order difference method for $\gamma$, and allocation of non-cluster center points. The implementation of the first part has been described in Sections III.B and III.C. The second part has been described in Section III.D thoroughly. The third part for allocating non-center points follows the original DPC depicted in Section II.A. Therefore, the specific process of SKTDPC algorithm is summarized as following Algorithm 2.

\begin{algorithm}[ht!]  
	\renewcommand{\algorithmicrequire}{\textbf{Input:}}
	\renewcommand{\algorithmicensure}{\textbf{Output:}}
	\caption{SKTDPC algorithm.}  
	\label{alg2}
	\begin{algorithmic}[1] 
		\Require The dataset $S=\{x_1,x_2,\ldots,x_n\}\in R^{n\times K}$; The number of the nearest neighbors $k$		
		\Ensure  The clustering result $Y$ 

      \State{Construct a K-d tree for dataset $S$, by Algorithm 1.} 
      \State{Search the $k$ nearest neighbors of the point $x_i$ based on the constructed K-d tree, by Algorithm 1.                Meanwhile, the symmetric sparse distance matrix $\widetilde D$ is obtained.}
\State{According to Equation (7) by the $k$ nearest neighbors, the first acceleration calculation is carried out to obtain $\rho_i$, and the descending order processing is made for $\rho_i$ to generate $\rho_i^*(i=1,2,\ldots,n)$.}
      \State{According to Equations.(3) and (8) by the sparse search method, the second acceleration calculation is carried out to obtain $\delta_i$.} 
      \State{According to Equation (9), calculate $\gamma_i$ and further obtain $\gamma_i^*$ by descending order processing.} 
      \State{According to Equation (10), calculate the mutation point $M_p$ to determine the candidate cluster centers as the data points with decision value $\gamma_i^*(i=1,2,\ldots,M_p)$. The candidate centers $x_i(i\in[1,M_p])$, which satisfies both $\rho_i>\frac{\sum_{t=1}^{\lfloor \sqrt{n} \rfloor}\rho_t}{\lfloor \sqrt{n} \rfloor}$ and $\delta_i>\frac{\sum_{t=1}^{\lfloor \sqrt{n} \rfloor}\delta_t}{\lfloor \sqrt{n} \rfloor}$ conditions, is the final cluster centers.} 
      \State{Assign the non-center points, according to the allocation principle, that each non-center point and its nearest point with a higher local density have the same cluster.} 
      \State{Return the clustering result $Y$.} 
	\end{algorithmic}
\end{algorithm}

Algorithm 2 summarizes the entire process of SKTDPC. On the one hand, SKTDPC replaces the calculation of distance for symmetric full-rank distance matrix $D$ with the symmetric sparse distance matrix $\widetilde D$. The K-d tree method and the sparse search strategy by the intersection between sets $\text{NN}_k(x_i)$ and $B(x_i)$ are developed to accelerate the calculation of both $\rho_i$ and $\delta_i$. On the other hand, the second-order difference method is used to find the boundary between the center points and the non-center points. Therefore, adaptively determining the cluster centers is achieved quickly and successfully. Finally, the present algorithm can effectively reduce the computational complexity while maintaining or even improving the clustering accuracy.

\subsection{ANALYSIS OF COMPLEXITY}
It should be stressed that the analysis of computational complexity of DPC in some literature only focus on the variable, data number $n$, then the complexity is $O(n^2)$  [18]–[30]. Different from these literature, the present work considers the influence of the two variables, $n$ and $K$, simultaneously. Thus, the computational complexity of the original DPC can be regarded as $O(n^2K)$.

The computational complexity of SKTDPC algorithm is mainly determined by the four parts: (1) Calculation process of local density $\rho_i$ based on K-d tree. The computational complexity of this part mainly depends on the process of searching $k$ nearest neighbors of data points. The process includes construction of the tree and the search of $k$ nearest neighbors. The complexity of these two parts is $O(nK\text{log}n)$ and $O(n(n^{1-1/K}+k))$ respectively. Thus, the complexity of the calculation process of local density $\rho_i$ based on K-d tree is $O(nK\text{log}n)+O(n(n^{1-1/K}+k))$. (2) Acquisition process of relative-separation $\delta_i$ based on the sparse search strategy with intersection between $\text{NN}_k(x_i)$ and $B(x_i)$. The complexity of this part is far less than $O(n(n^{1-1/K}+k))$, as analyzed in Section III.C in detail. (3) Adaptive determination of cluster centers. The complexity of this part is determined by descending order, second-order difference, and determining the center points that satisfy the mean value condition. The complexity of the three steps is $O(n\text{log}n)$, $O(\lfloor \sqrt{n} \rfloor-2)$ and  $O(M_p)$, respectively. Thus, the computational complexity of the whole part is $O(n\text{log}n)+O(\lfloor \sqrt{n} \rfloor-2)+O(M_p)$. (4) Allocation of non-center points. The computational complexity of this part is $O(n)$. The total computational complexity of SKTDPC algorithm is $O(nK\text{log}n)+O(n(n^{1-1/K}+k))+O(n\text{log}n)+O(\lfloor \sqrt{n} \rfloor-2)+O(M_p)+O(n)$. Normally, the dimension $K$ is much smaller than the data number $n$. Therefore, the overall computational complexity of the SKTDPC algorithm is the largest one among the four parts. That is, the complexity is $O(n(n^{1-1/K}+k))$. In order to make it easier to analyze and compare with the complexity $O(n^2K)$ of DPC, $O(n(n^{1-1/K}+k))$ could be transformed into $O(n^2(1/\sqrt[K]{n}+k/n))$, where $1/\sqrt[K]{n}, k/n$ take values in range $(0,1)$. In other words, the computational complexity of SKTDPC algorithm is much lower than the complexity $O(n^2K)$ of DPC.

\section{EXPERIMENTS AND RESULTS}
In this section, the present SKTDPC algorithm is compared with the six state-of-the-art and typical clustering algorithms, including FSDPC \cite{b26}, the original DPC \cite{b14}, DGDPC \cite{b38}, DPC-KNN \cite{b18}, DBSCAN \cite{b16} and the K-means algorithm \cite{b15}. For the seven algorithms, the first five algorithms belong to the DPC series. FSDPC, the original DPC and DGDPC are the algorithms taking cut-off distance $d_c$ as a hyper-parameter, and DGDPC has an extra merging threshold $l$  hyper-parameter. However, SKTDPC and DPC-KNN adopt the hyper-parameter $k$, the number of nearest neighbors. DBSCAN is another kind of density-based clustering algorithm, for which two hyper-parameters, are used. In contrast, the K-means algorithm is a fast clustering algorithm, in which the allocation of non-centers points is determined by the nearest distance from cluster centers.

\subsection{INTRODUCTION TO DATASETS AND METRICS}
To verify the clustering effect and efficiency of the SKTDPC algorithm, 15 commonly used clustering datasets \cite{b38,b39,b40,b41,b42,b43,b44,b45} are adopted in the present work. These include the eight synthetic datasets (http://cs.joensuu.fi/sipu/data-sets/) shown in Table \ref{tab1} and the seven UCI real datasets \cite{b46} (http://archive.ics.uci.edu/ml) which are shown in Table \ref{tab2}. In the two tables, the basic information such as number of data, number of class clusters, and data dimension is listed. Here, in order to visualize the clustering effect clearly and intuitively, in Table I all the eight synthetic datasets are two-dimensional. In contrast, the UCI real datasets are high-dimensional. They are discussed in Sections IV.B, IV.C and IV.D, respectively.

\begin{table}
\centering
\caption{\textbf{CHARACTERISTICS OF SYNTHETIC DATASETS.}}
\setlength{\tabcolsep}{12pt}
\begin{tabular}{lccc}
\hline\hline
Name& Data& Clusters&Dimension \\
\hline
Flame& 240&2&2 \\
Spiral&312&3&2\\ 
Aggregation&788&7&2\\
R15&600&15&2\\
S1&5000&15&2\\
S3&5000&15&2\\
A1&3000&20&2\\
A3&7500&50&2\\
\hline\hline
\end{tabular}
\label{tab1}
\end{table}

\begin{table}
\centering
\caption{ \textbf{CHARACTERISTICS OF UCI REAL DATASETS.}}
\setlength{\tabcolsep}{10pt}
\begin{tabular}{lccc}
\hline\hline
Name& Data& Clusters&Dimension \\
\hline
Seeds& 210&3&7 \\
Iris&150&3&4\\ 
Banknote authentication&1372&2&4\\
Wine&178&3&13\\
Ecoli&336&8&7\\
Parking Birmingham&35501&3&5\\
Pendigits&10992&10&16\\
\hline\hline
\end{tabular}
\label{tab2}
\end{table}

\begin{table*}
\centering
\caption{ \textbf{ACC OF ALGORITHMS ON SYNTHETIC DATASETS.}}
\begin{tabular}{lcccccccccccccc}
\hline\hline
\multirow{2}{*}{Data} &\multicolumn{2}{c}{SKTDPC} &\multicolumn{2}{c}{FSDPC}&\multicolumn{2}{c}{DPC}&\multicolumn{2}{c}{DGDPC}&\multicolumn{2}{c}{DPC-KNN}&\multicolumn{2}{c}{DBSCAN}&\multicolumn{2}{c}{K-means}\\
\cline{2-15} & Acc & Par & Acc & Par & Acc & Par & Acc & Par & Acc & Par & Acc& Par & Acc & Par    \\
\hline
Flame & \bf 1 & 3 &\bf 1 & 6 &\bf 1 & 6 &\bf 1 & 0.5/5 &\bf 1 & 5 & 0.975 & 0.09/8 & 0.838 & 2\\
Spiral & \bf 1 & 4 &\bf 1 & 2 &\bf 1 & 2 &\bf 1 & 0/5 & \bf 1 & 7 &\bf 1 & 0.04/2 & 0.343 & 3\\
Aggregation &\bf 0.997 & 6 & 0.992 & 0.19 & 0.992 & 0.19 & 0.996 & 0.5/5 & 0.996 & 5 & 0.953 & 0.05/8 & 0.860 & 7\\
R15 &\bf 0.997&	5&	0.996	&1	&\bf 0.997&	1	&\bf 0.997&	0.5/5&	0.996&	5&	0.873&	0.02/6&	0.996	&15\\
S1	&\bf 0.997&	7	&0.992&	1	&0.992&	1	&0.992&	1/1&	0.992&	6&	0.975	&0.03/13&	0.992	&15\\
S3&\bf	0.901&	3	&0.858&	2&	0.857&	2&	0.868&	1/1&	0.522&	6&	0.700&	0.04/73&	0.873&	15\\
A1&\bf	0.998&	6&	0.971	&2&	0.973	&2	&0.972&	0.5/1	&0.663&	5	&0.825&	0.03/25&	0.948&	20\\
A3&\bf	0.996&	7&	0.990&	0.2&	0.988&	0.2&	0.991&	1/1&	0.625&	5	&0.921&	0.03/62&	0.989&	50\\
\hline\hline
\end{tabular}
\label{tab3}
\end{table*}

\begin{table*}
\centering
\caption{ \textbf{AMI, ARI OF ALGORITHMS ON SYNTHETIC DATASETS.}}
\begin{tabular}{lcccccccccccccc}
\hline\hline
\multirow{2}{*}{Data} &\multicolumn{2}{c}{SKTDPC} &\multicolumn{2}{c}{FSDPC}&\multicolumn{2}{c}{DPC}&\multicolumn{2}{c}{DGDPC}&\multicolumn{2}{c}{DPC-KNN}&\multicolumn{2}{c}{DBSCAN}&\multicolumn{2}{c}{K-means}\\
\cline{2-15} & AMI & ARI & AMI & ARI & AMI & ARI & AMI & ARI & AMI & ARI & AMI& ARI & AMI & ARI    \\
\hline
Flame &\bf 1&\bf	1	&\bf 1&\bf	1&\bf	1	&\bf 1	&\bf 1	&\bf 1&\bf	1&\bf	1	&0.839&	0.923&	0.386&	0.453\\
Spiral &\bf 1&\bf	1&\bf	1&\bf	1	&\bf 1&\bf	1&\bf	1	&\bf 1	&\bf 1	&\bf 1	&\bf 1	&\bf 1	&0	&0\\
Aggregation&\bf 0.992&\bf	0.996&	0.979&	0.990&	0.978&	0.991&\bf	0.992&\bf	0.996&	0.989&	0.993&	0.949&	0.911&	0.834	&0.730\\
R15 &\bf 0.994&\bf	0.993&	0.993&\bf	0.993&	0.993&	\bf 0.993&	0.993&\bf	0.993&	0.993&\bf	0.993&	0.845&	0.717&\bf	0.994&\bf	0.993\\
S1	&\bf 0.994&\bf 	0.994&	0.991&	0.991&	0.991&	0.991&	0.991&	0.991&	0.991&	0.991&	0.961&	0.962&	0.991&	0.992\\
S3&\bf	0.858&\bf	0.803&	0.796&	0.729	&0.796&	0.728&	0.814&	0.757&	0.576&	0.566&	0.667&	0.392	&0.833&	0.775\\
A1&\bf	0.997&\bf	0.996&	0.958&	0.941&	0.958&	0.942&	0.958	&0.941&	0.782&	0.562&	0.871&	0.733&	0.959&	0.944\\
A3&\bf	0.996&\bf	0.992&0.991&	0.978	&0.990&	0.978&	0.991&	0.979&	0.763	&0.538	&0.932&	0.858&	0.987&	0.980\\
\hline\hline
\end{tabular}
\label{tab4}
\end{table*}

\begin{table*}
\centering
\caption{\textbf{NMI, FMI OF ALGORITHMS ON SYNTHETIC DATASETS.}}
\begin{tabular}{lcccccccccccccc}
\hline\hline
\multirow{2}{*}{Data} &\multicolumn{2}{c}{SKTDPC} &\multicolumn{2}{c}{FSDPC}&\multicolumn{2}{c}{DPC}&\multicolumn{2}{c}{DGDPC}&\multicolumn{2}{c}{DPC-KNN}&\multicolumn{2}{c}{DBSCAN}&\multicolumn{2}{c}{K-means}\\
\cline{2-15} & NMI & FMI & NMI & FMI & NMI & FMI & NMI & FMI & NMI & FMI & NMI& FMI & NMI & FMI    \\
\hline
Flame &\bf 1	&\bf 1	&\bf 1	&\bf 1	&\bf 1	&\bf 1&\bf	1&\bf	1	&\bf 1	&\bf 1	&0.841&	0.964&	0.399&	0.736\\
Spiral &\bf 1&\bf	1&\bf	1	&\bf 1	&\bf 1&\bf	1&\bf	1	&\bf 1&\bf	1&\bf	1	&\bf 1&\bf	1&	0	&0.328\\
Aggregation&\bf 0.992&\bf	0.997&	0.982&	0.987&	0.984&	0.986&\bf	0.992&	0.996&	0.987&	0.990&	0.950&	0.933&	0.837&	0.788\\
R15 &\bf 0.994&\bf	0.993&\bf	0.994&\bf	0.993&\bf	0.994&\bf	0.993	&\bf 0.994&\bf	0.993&\bf	0.994&\bf	0.993&	0.856&	0.736&\bf	0.994&\bf	0.993\\
S1	&\bf 0.994&\bf 	0.994&	0.991&	0.992	&0.991&	0.992&	0.991	&0.992&	0.991&	0.992&	0.962&	0.964&	0.992&	0.992\\
S3&\bf	0.859&\bf	0.817	&0.780&	0.745&	0.799&	0.746	&0.803	&0.766&	0.575&	0.572&	0.670&	0.449&	0.835&	0.790\\
A1&\bf	0.997&\bf	0.997&	0.959&	0.947&	0.959	&0.945	&0.959	&0.947&	0.781&	0.560	&0.873	&0.752&	0.960&	0.947\\
A3&\bf	0.996&\bf	0.992&	0.989&	0.979	&0.990&	0.979&	0.987	&0.978	&0.762&	0.538&	0.935&	0.861&	0.988&	0.980\\
\hline\hline
\end{tabular}
\label{tab5}
\end{table*}

The algorithmic clustering performance could be effectively evaluated by using running time and five well-known indexes, including accuracy (Acc) \cite{b47}, adjusted mutual information (AMI) \cite{b48}, adjusted rand index (ARI) \cite{b48}, normalized mutual information (NMI) \cite{b49} and Fowlers-Mallows index (FMI) \cite{b50}. The five indexes are powerful tools to measure the clustering results, and the maximum values are 1 for all the indexes. The closer to 1 they are, the better the clustering effect is. In addition, running time is an important performance indicator used to measure the clustering efficiency. The smaller the value is, the higher the clustering efficiency is, vice versa.

\begin{figure*}[ht]
\centering
	\subfloat{\includegraphics[width = 0.2\textwidth]{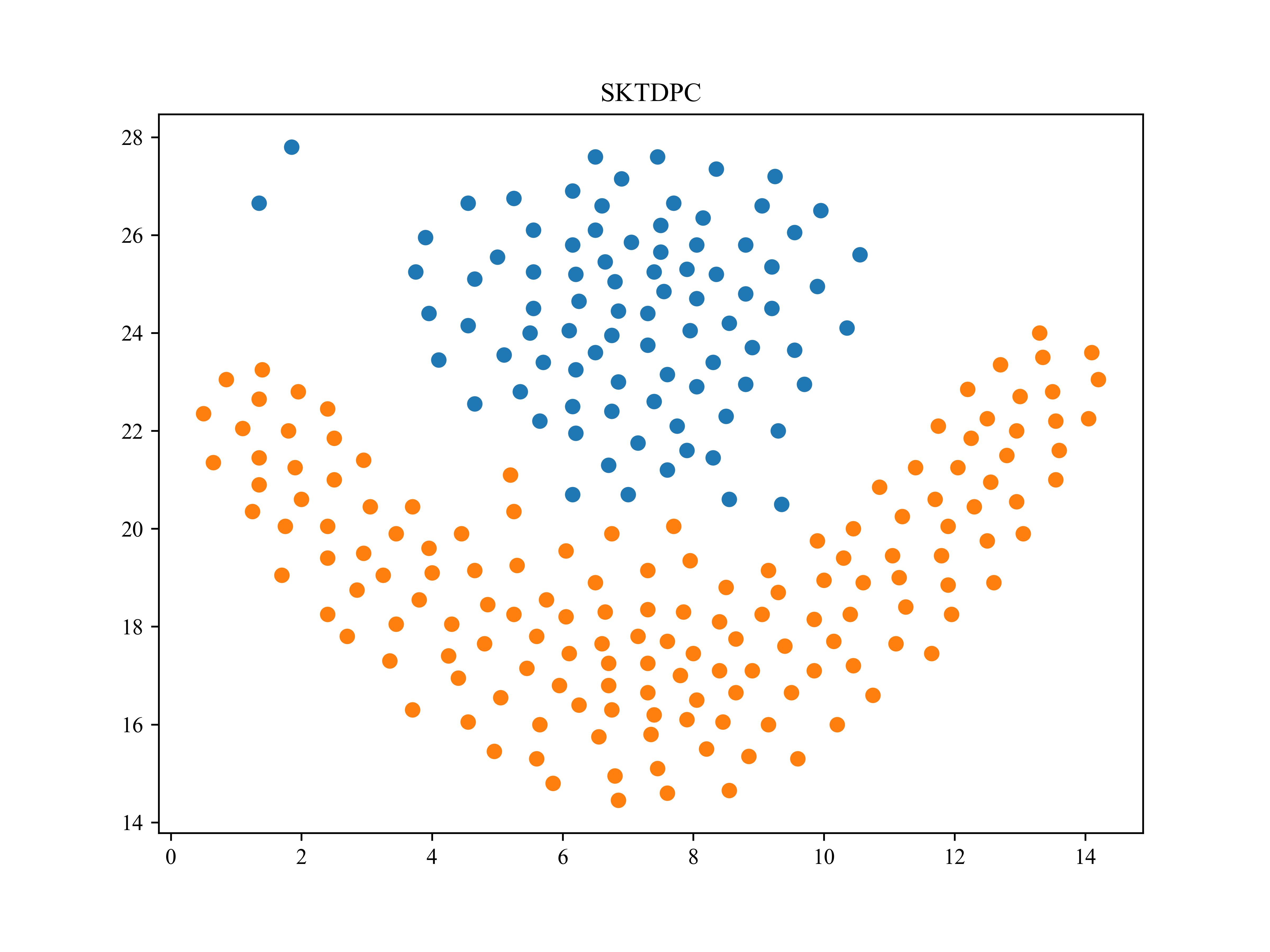}}
	\subfloat{\includegraphics[width = 0.2\textwidth]{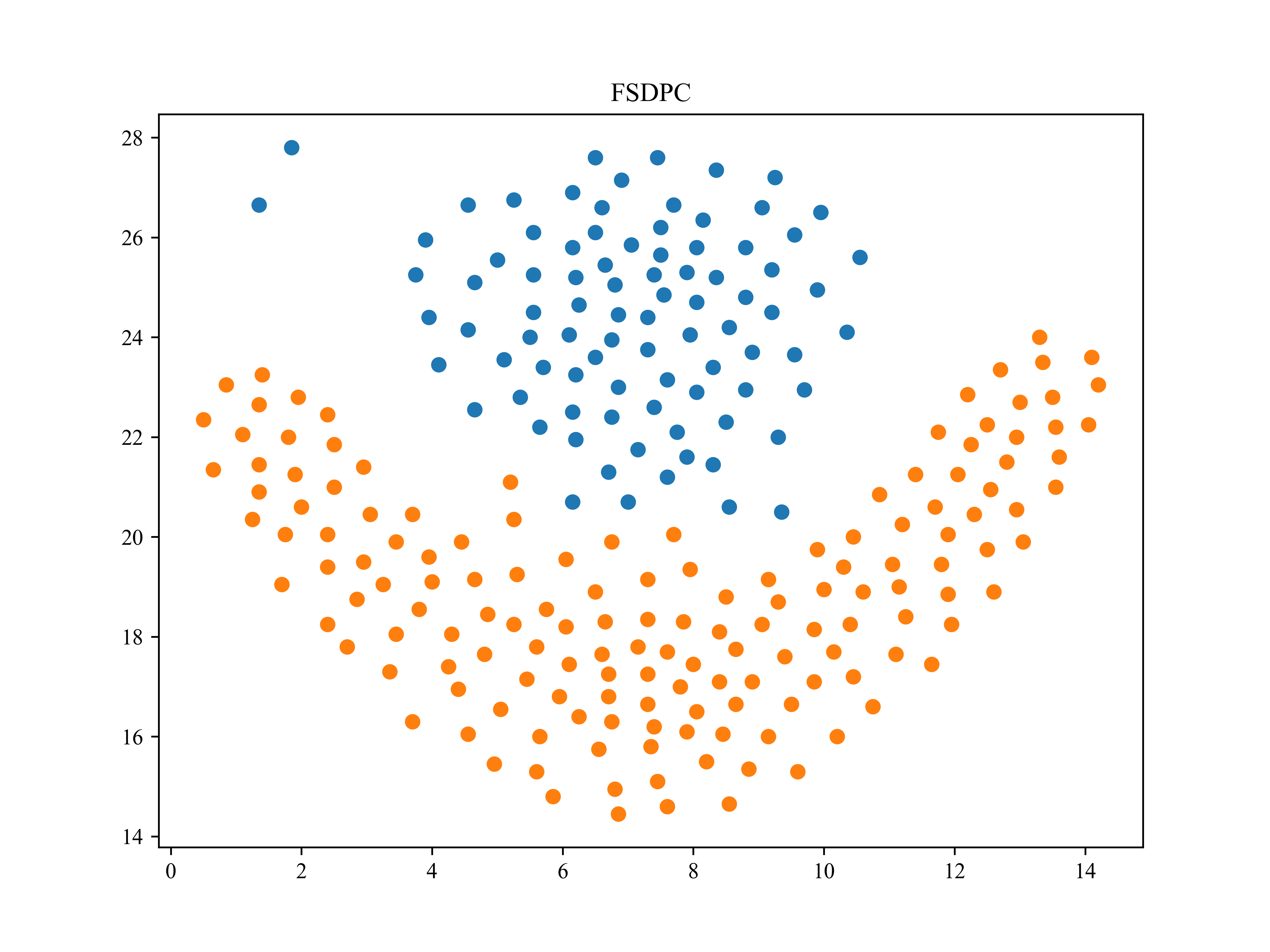}}
\subfloat{\includegraphics[width = 0.2\textwidth]{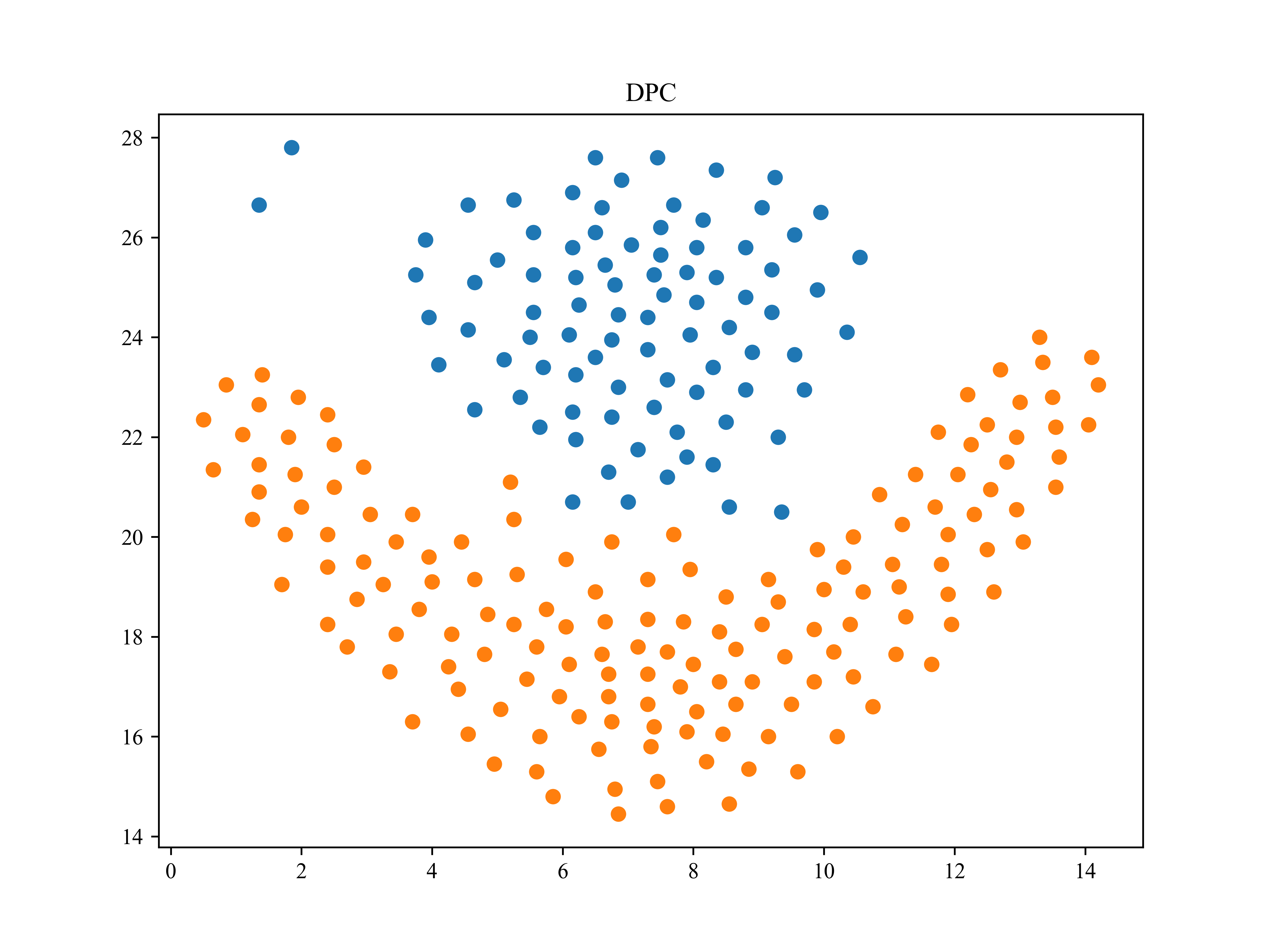}}
\subfloat{\includegraphics[width = 0.2\textwidth]{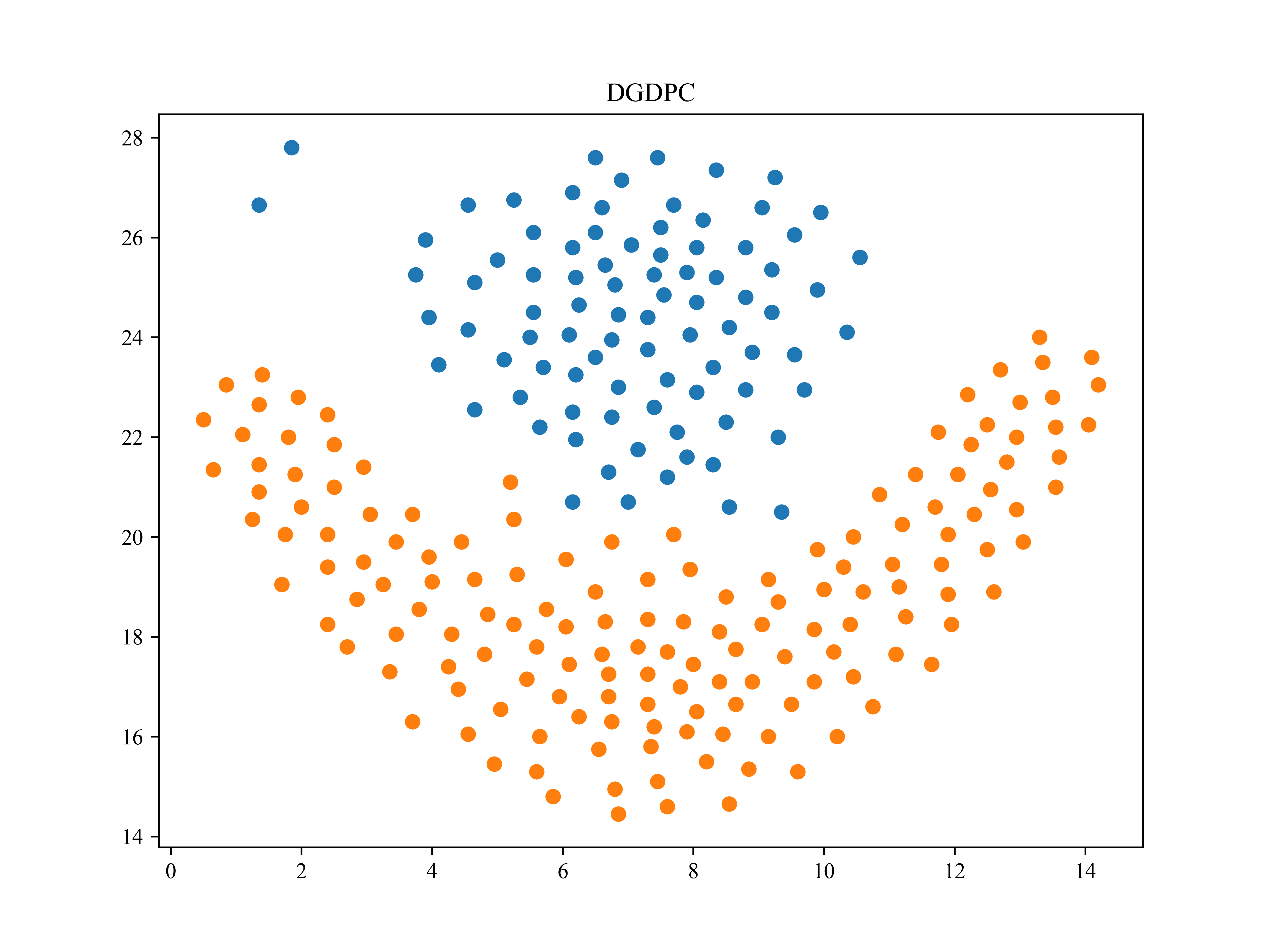}}\\
\subfloat{\includegraphics[width = 0.2\textwidth]{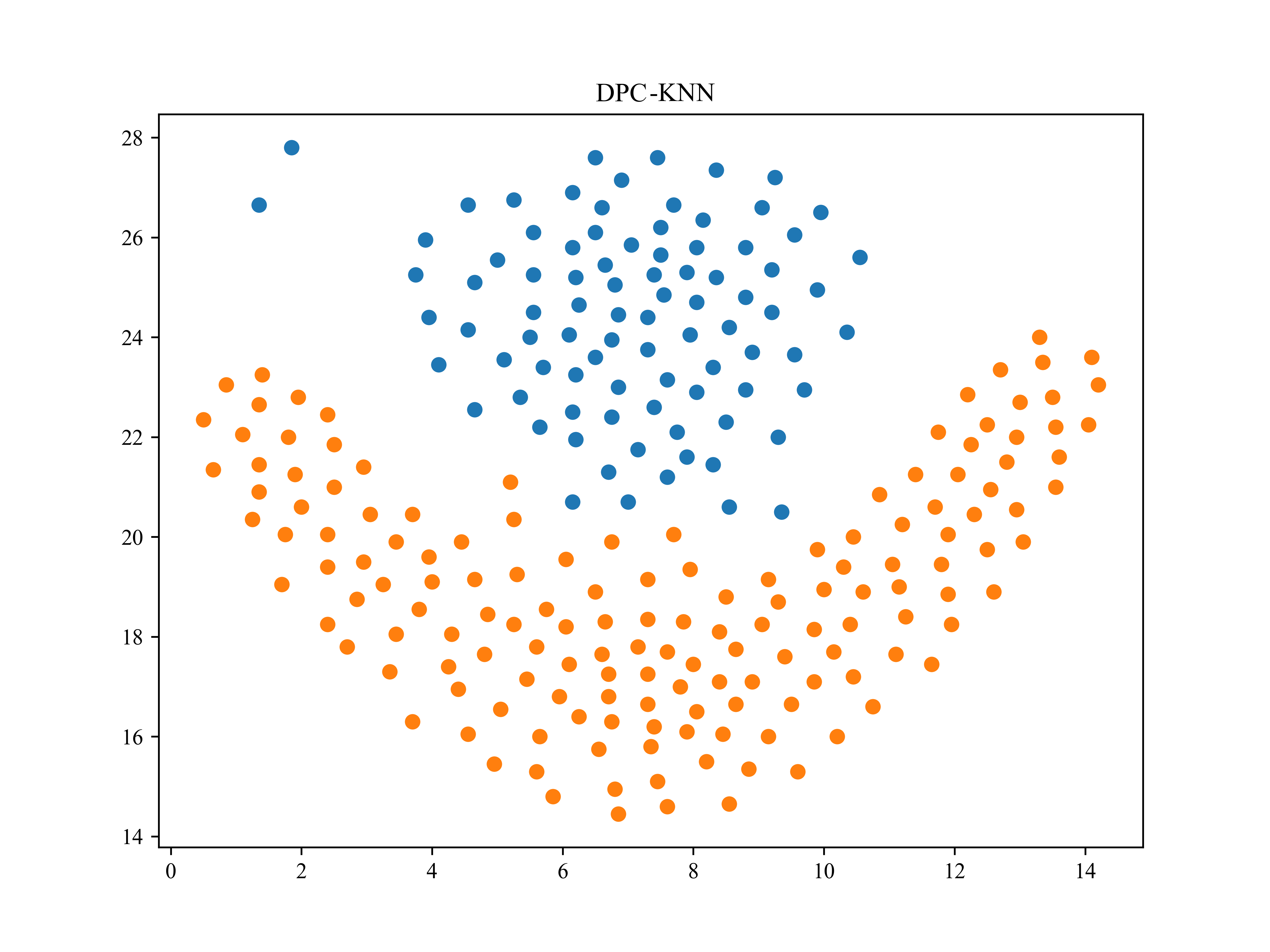}\centering}
\subfloat{\includegraphics[width = 0.2\textwidth]{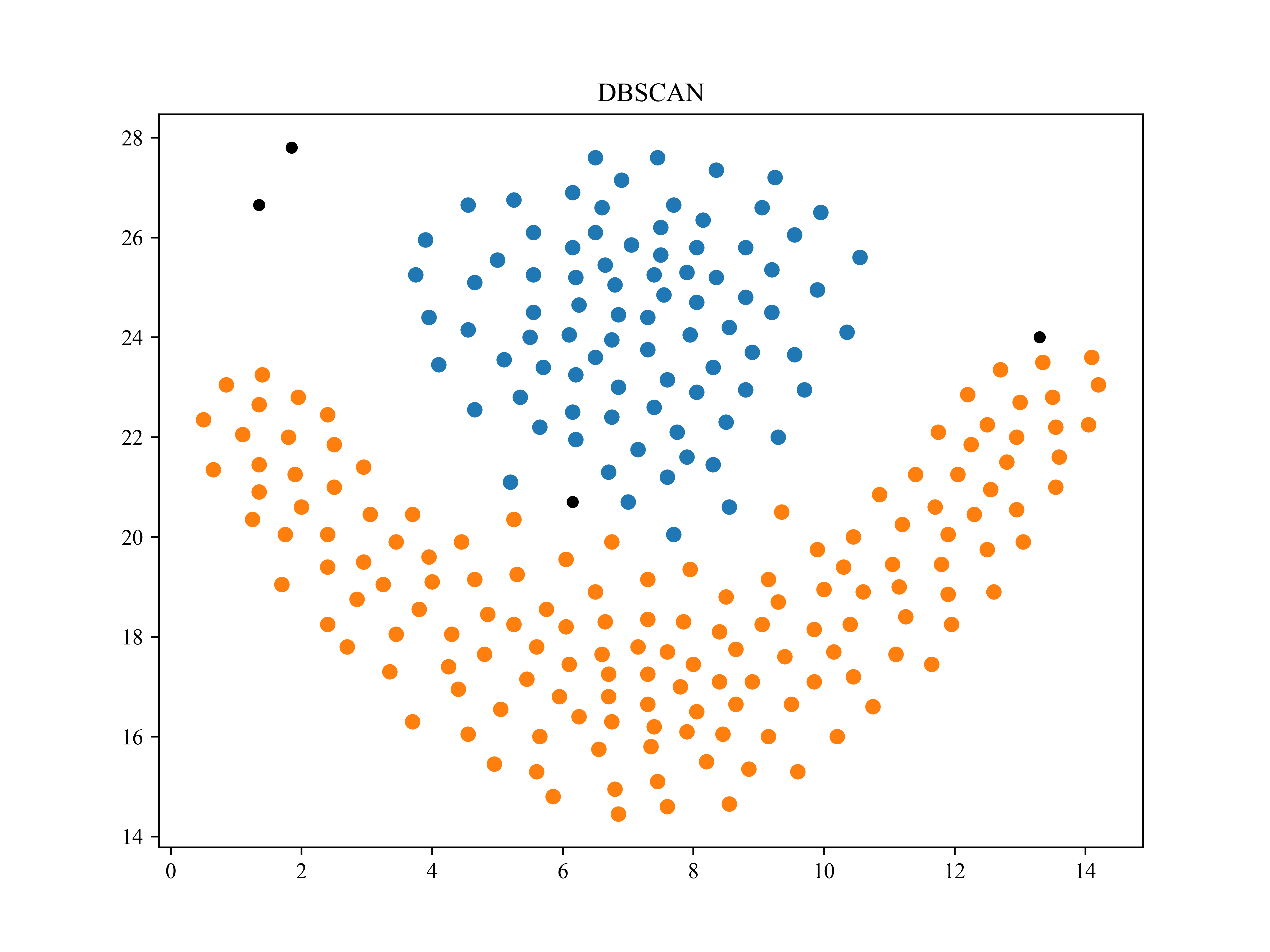}\centering}	
\subfloat{\includegraphics[width = 0.2\textwidth]{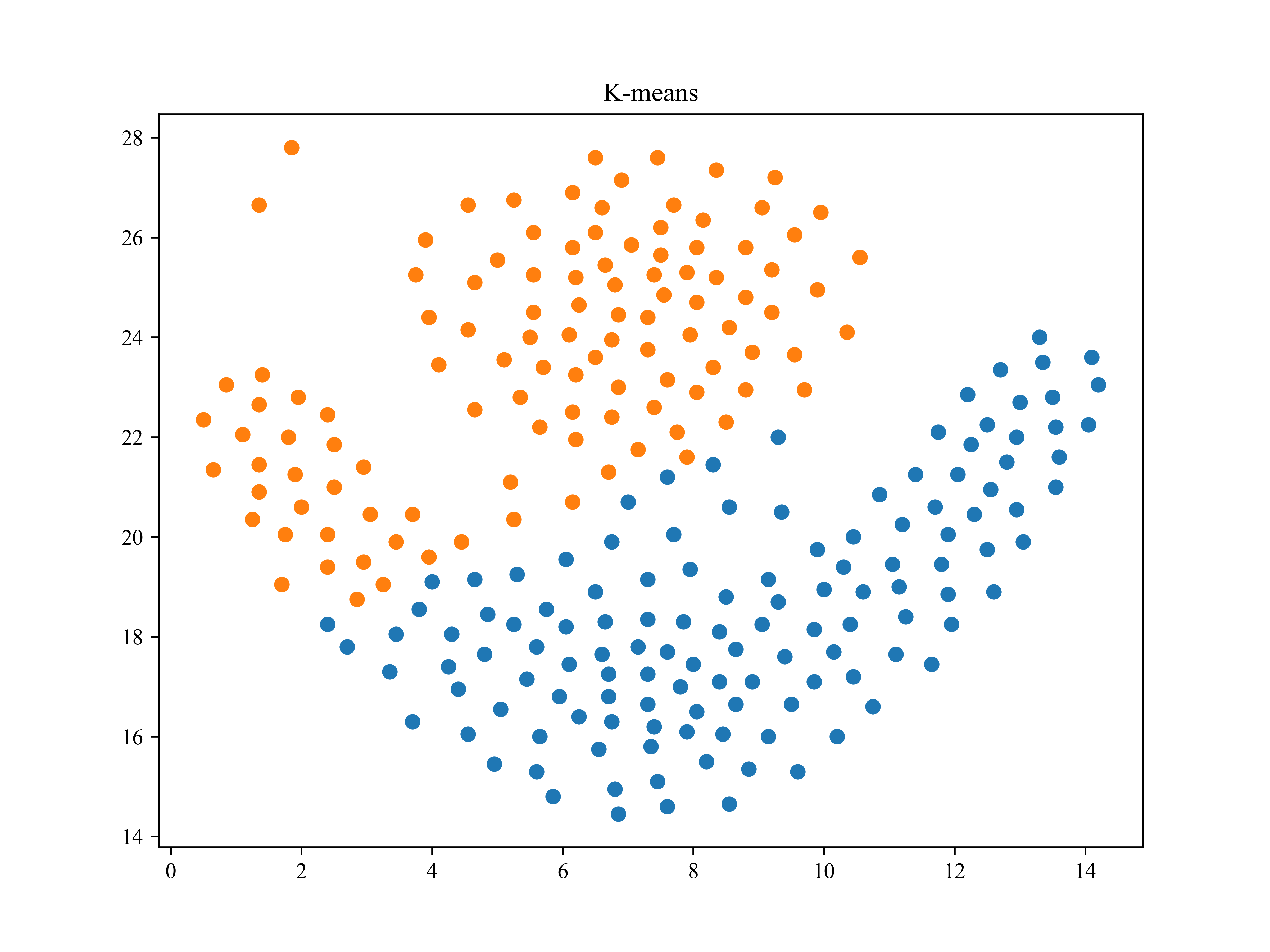}\centering}	
\caption{\textbf {Clustering results of the seven algorithms on Flame dataset.} }
\label{fig:labe2}
\end{figure*}

\begin{figure*}[ht]
\centering
	\subfloat{\includegraphics[width = 0.2\textwidth]{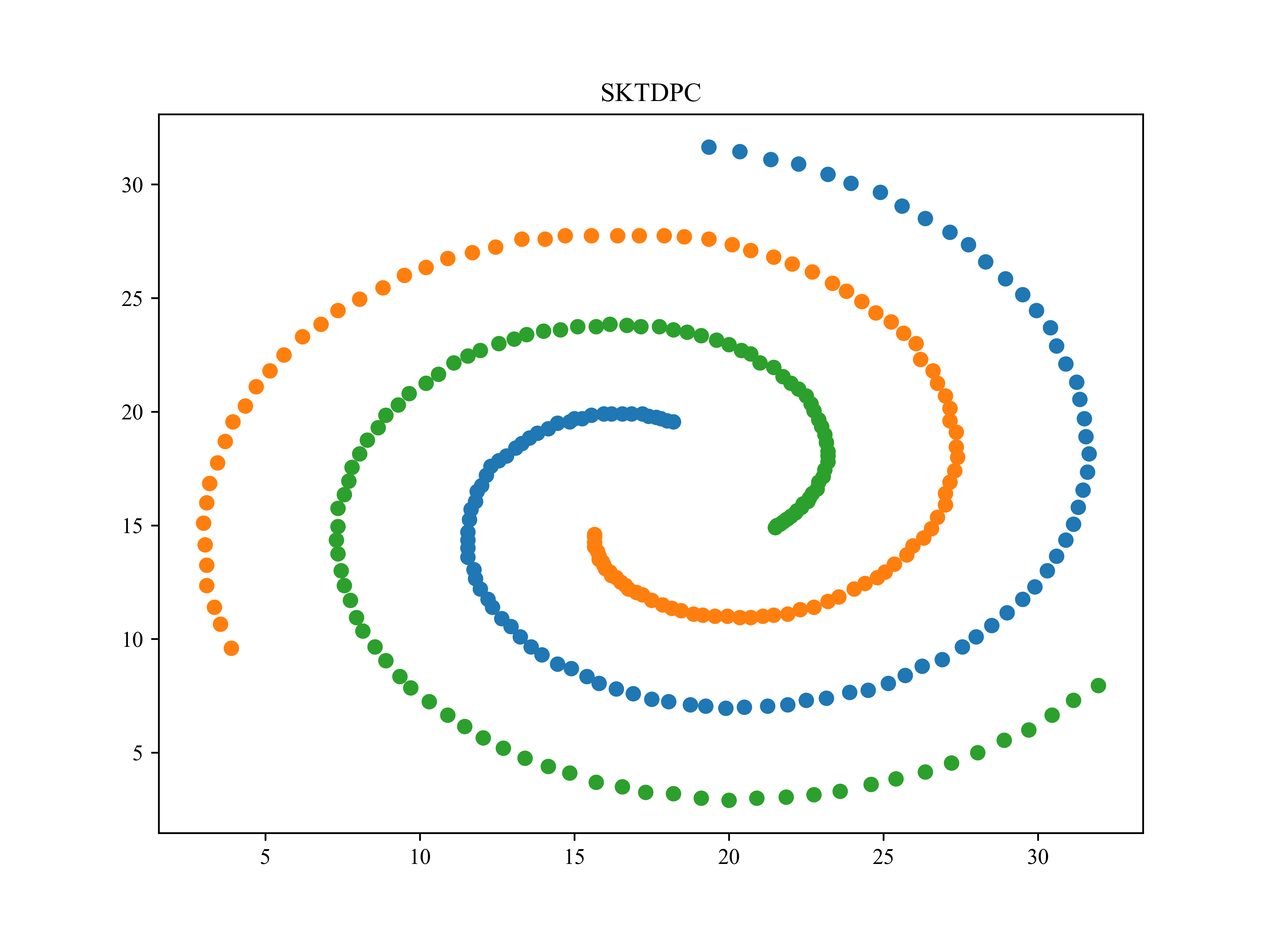}}
	\subfloat{\includegraphics[width = 0.2\textwidth]{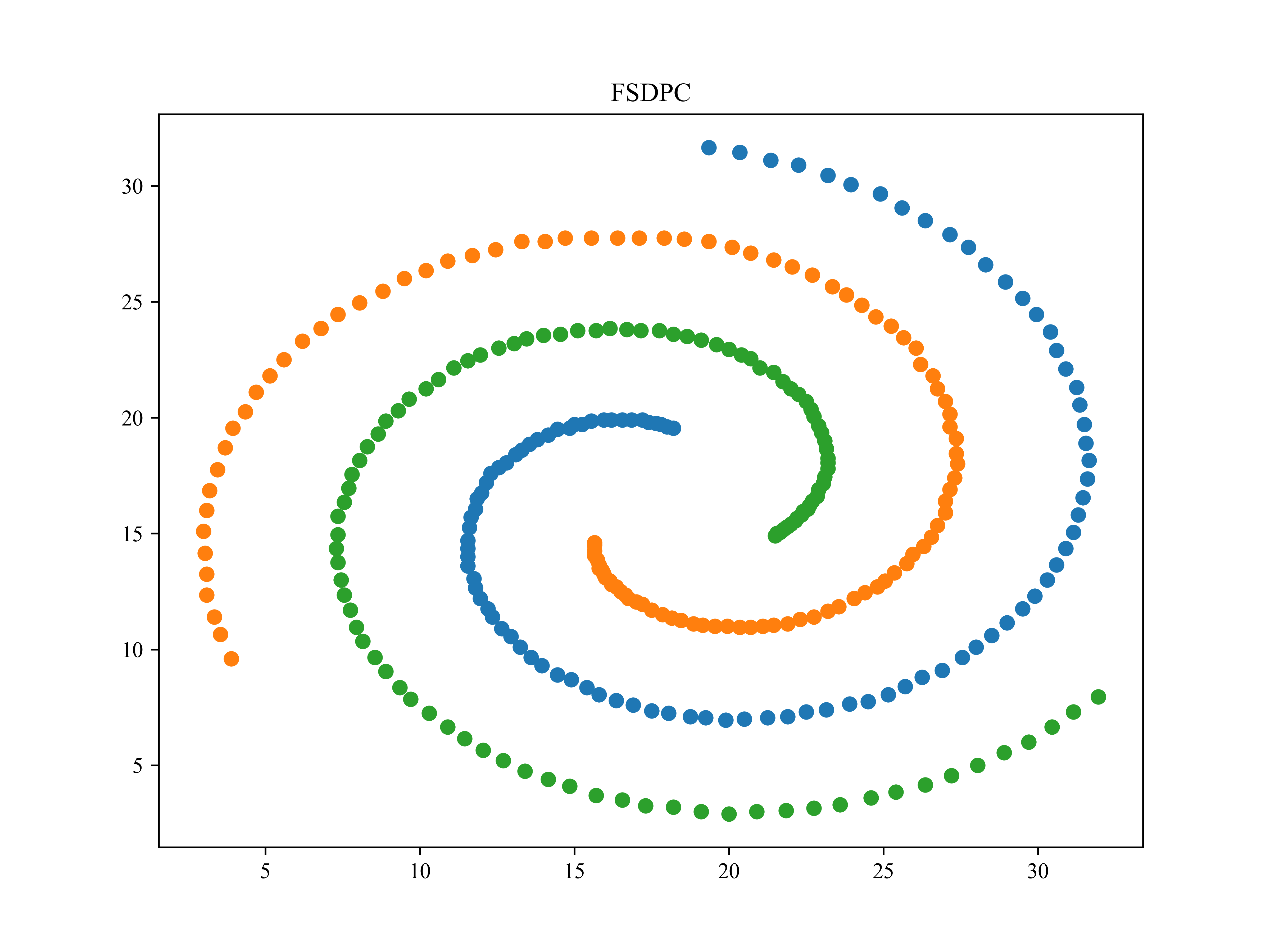}}
\subfloat{\includegraphics[width = 0.2\textwidth]{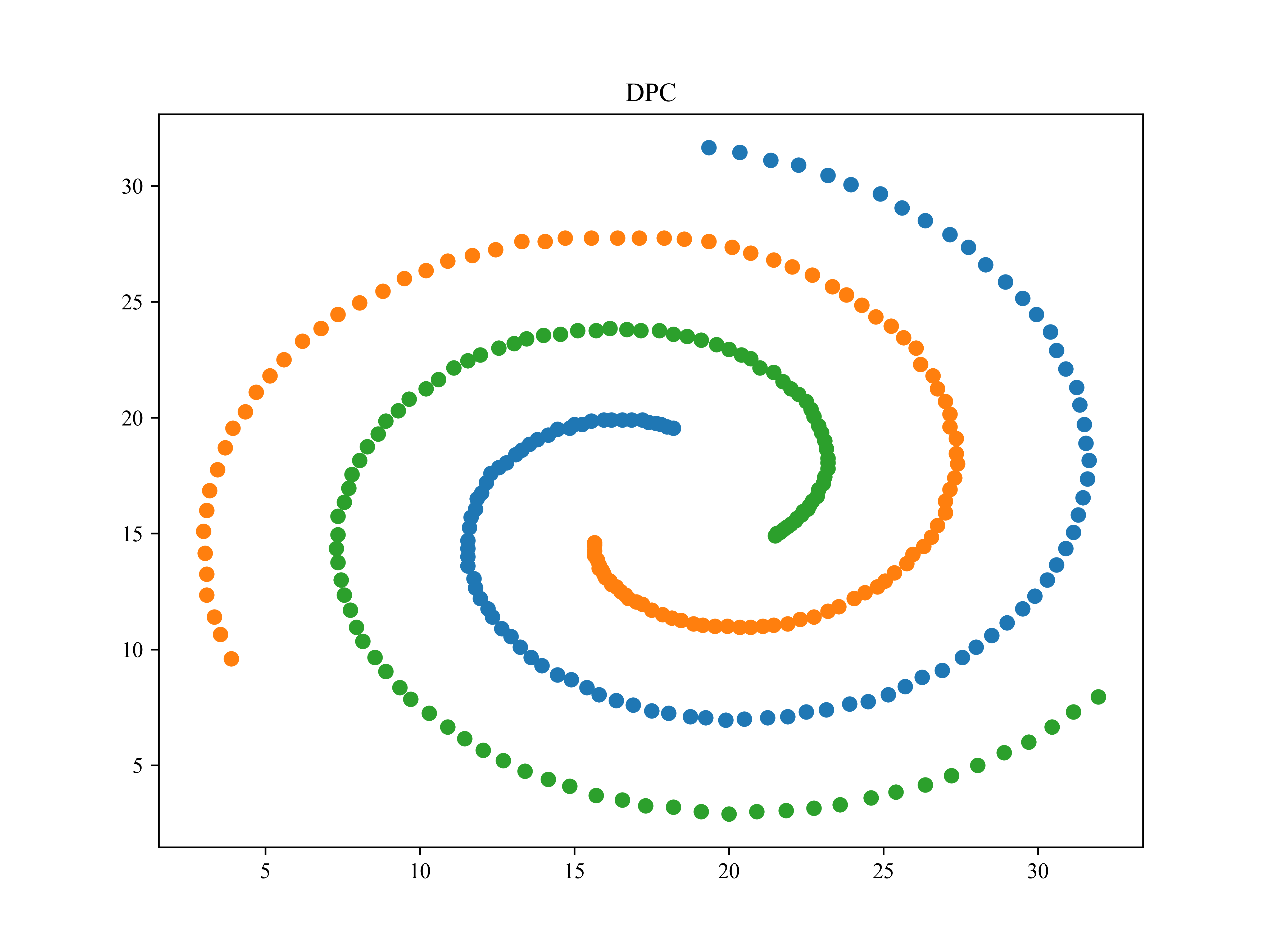}}
\subfloat{\includegraphics[width = 0.2\textwidth]{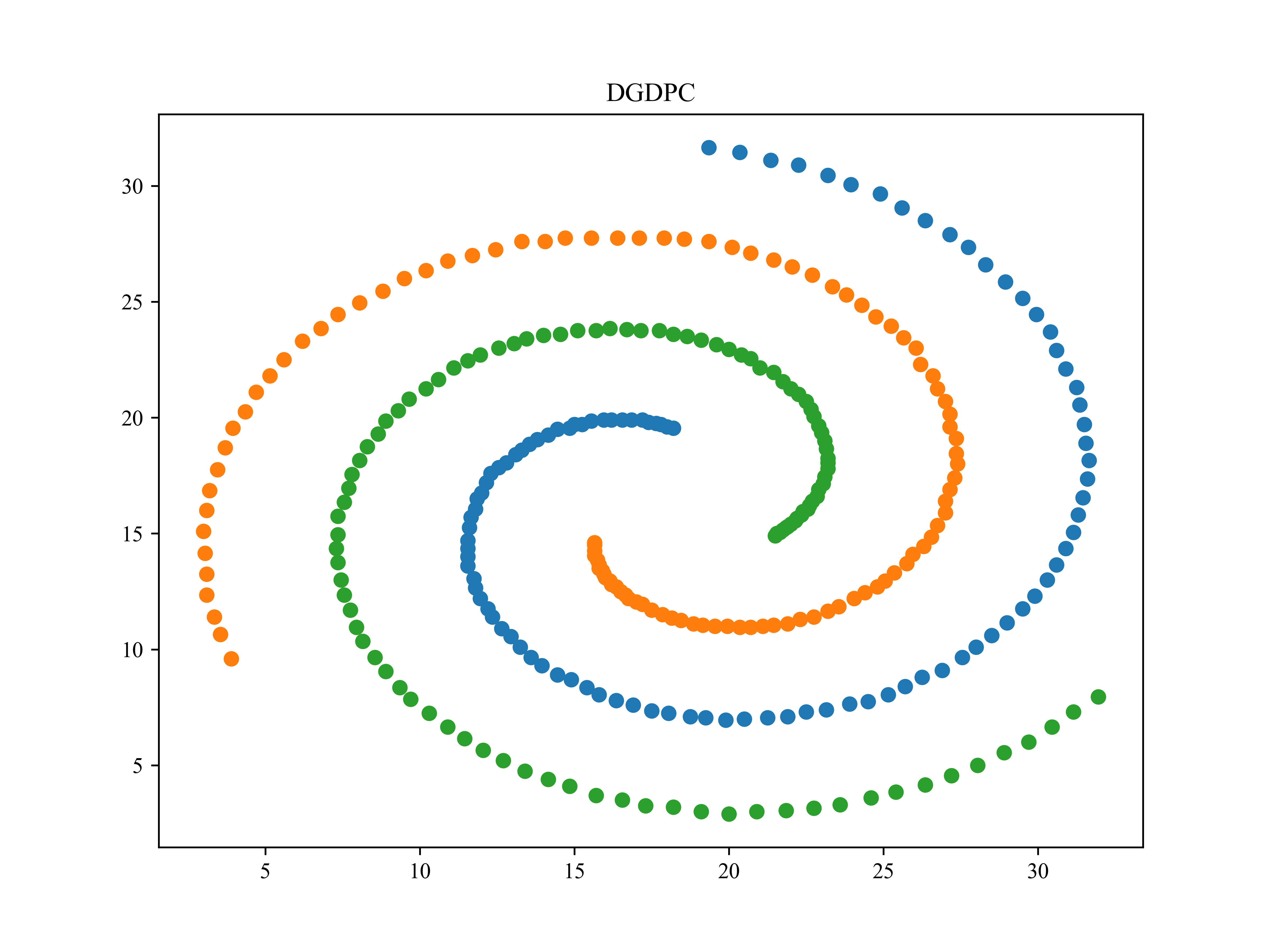}}\\
\subfloat{\includegraphics[width = 0.2\textwidth]{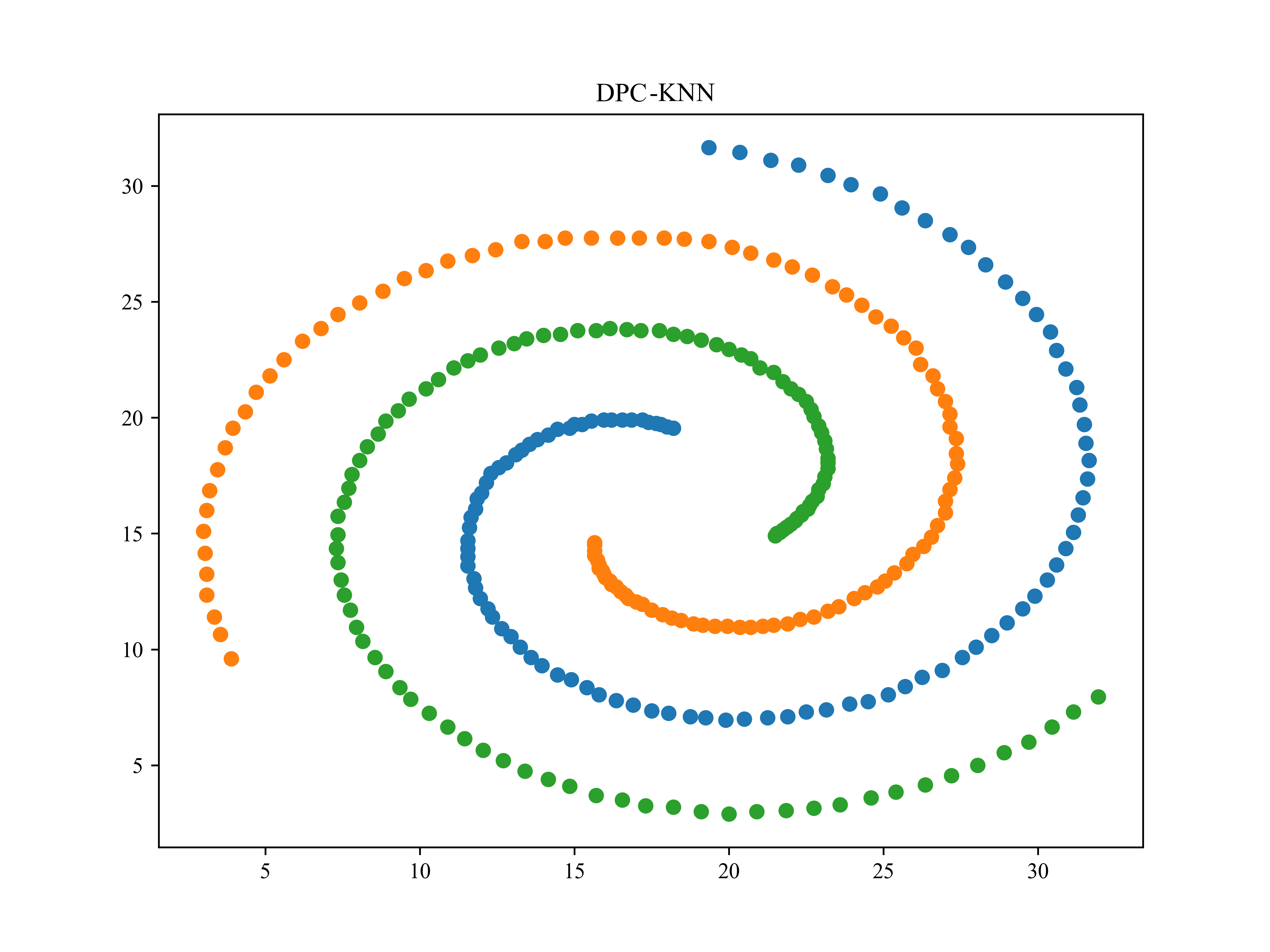}\centering}
\subfloat{\includegraphics[width = 0.2\textwidth]{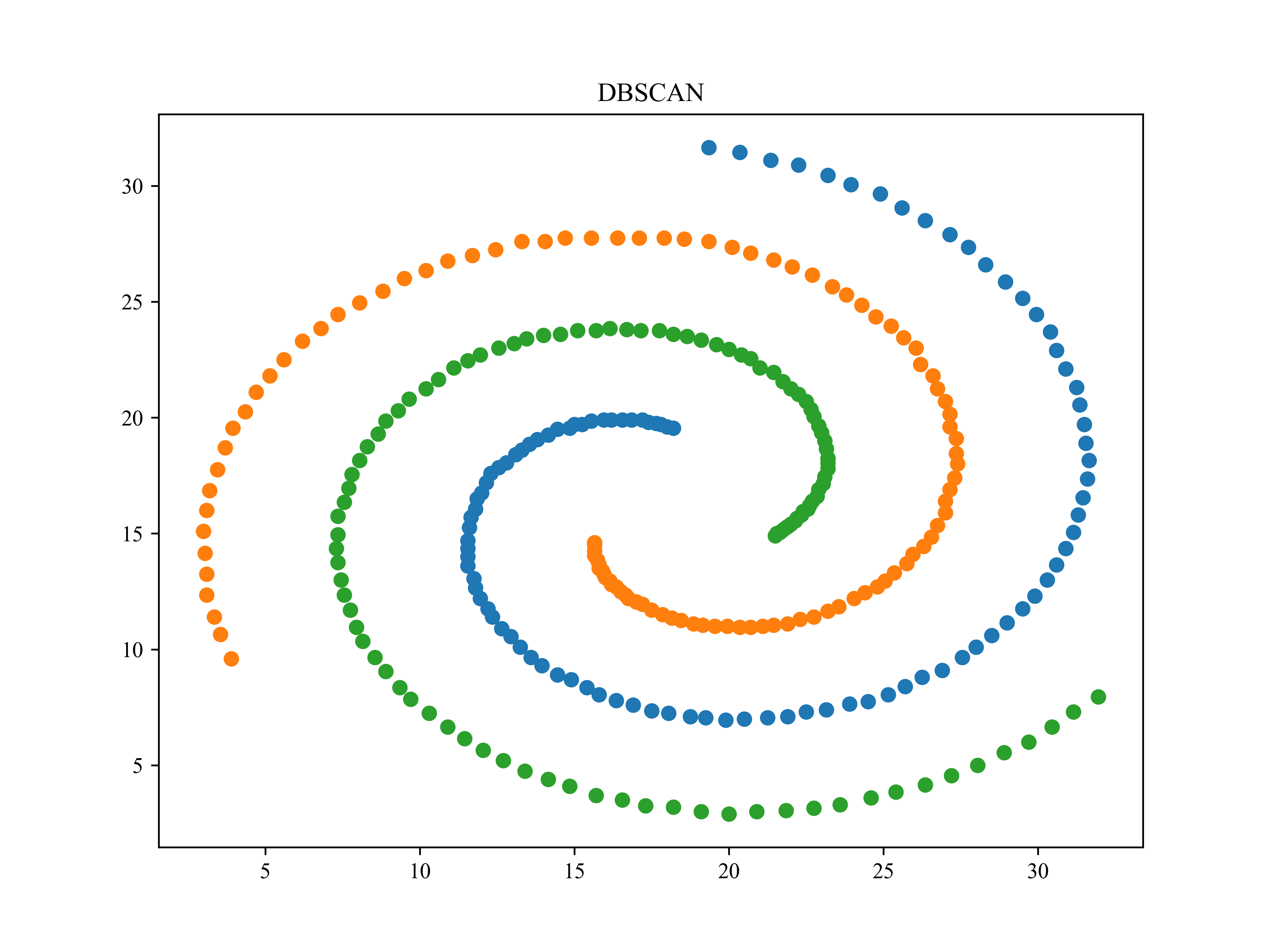}\centering}	
\subfloat{\includegraphics[width = 0.2\textwidth]{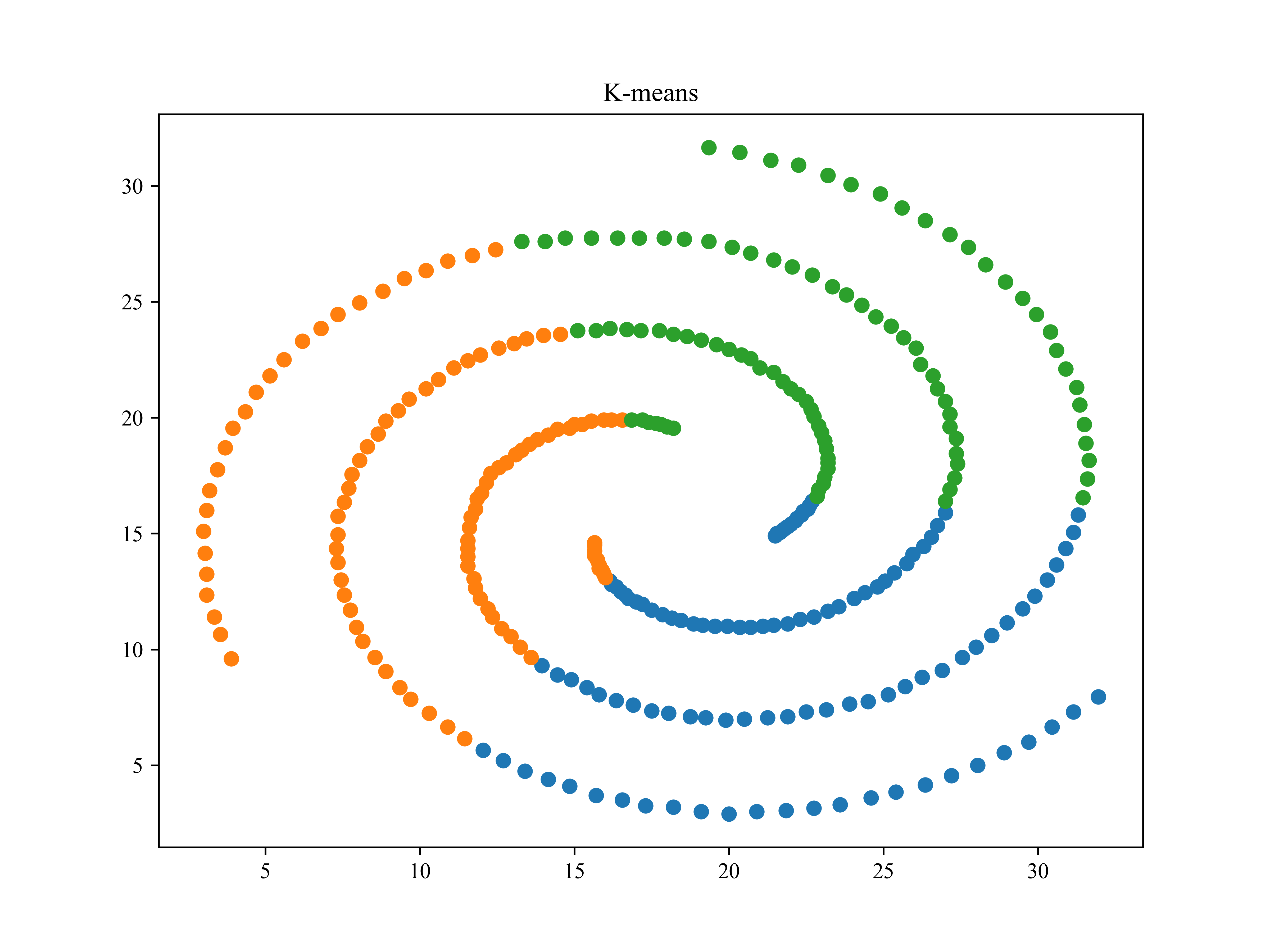}\centering}	
\caption{\textbf {Clustering results of the seven algorithms on Spiral dataset.} }
\label{fig:labe3}
\end{figure*}

\begin{figure*}[ht]
\centering
	\subfloat{\includegraphics[width = 0.2\textwidth]{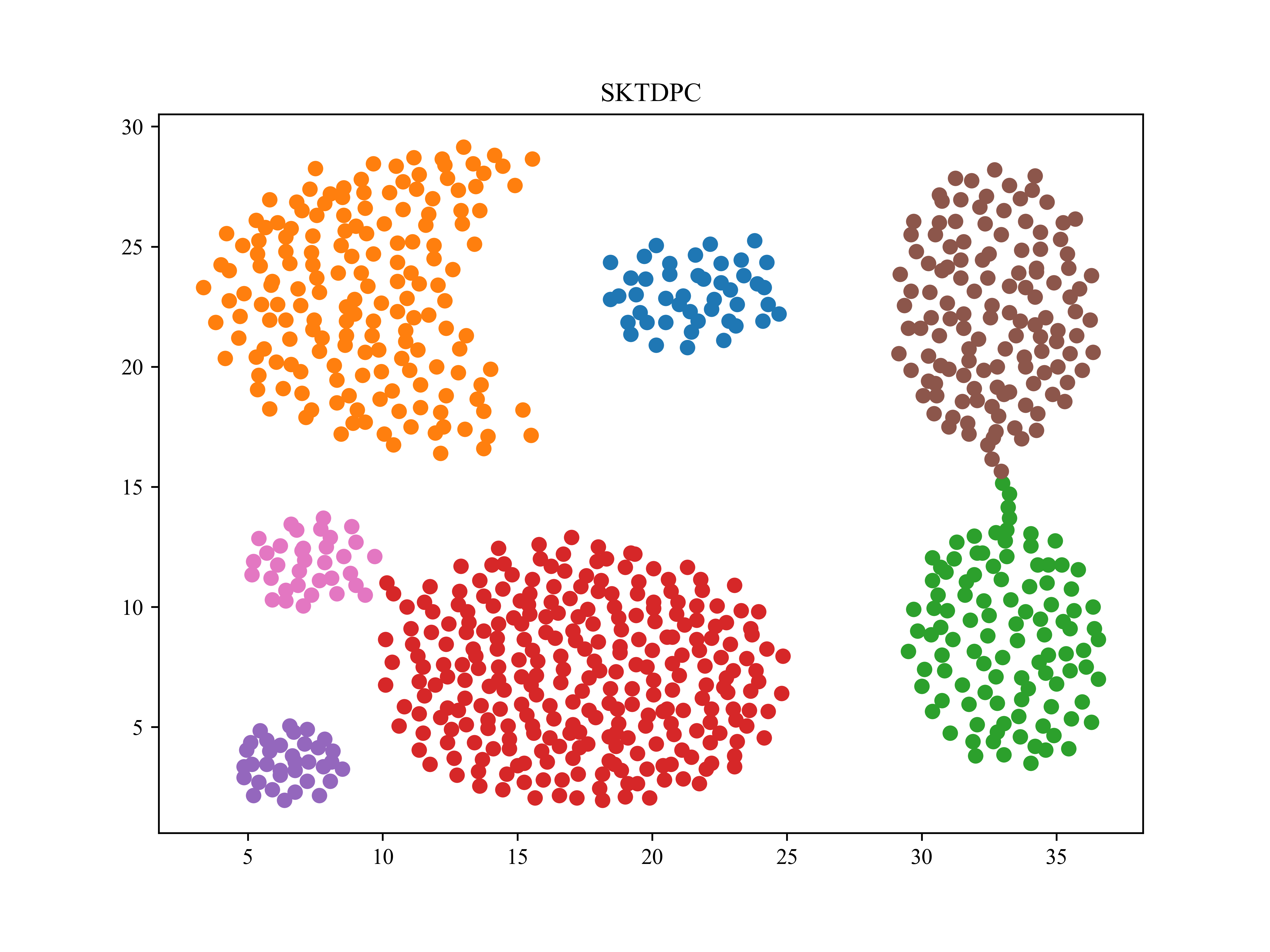}}
	\subfloat{\includegraphics[width = 0.2\textwidth]{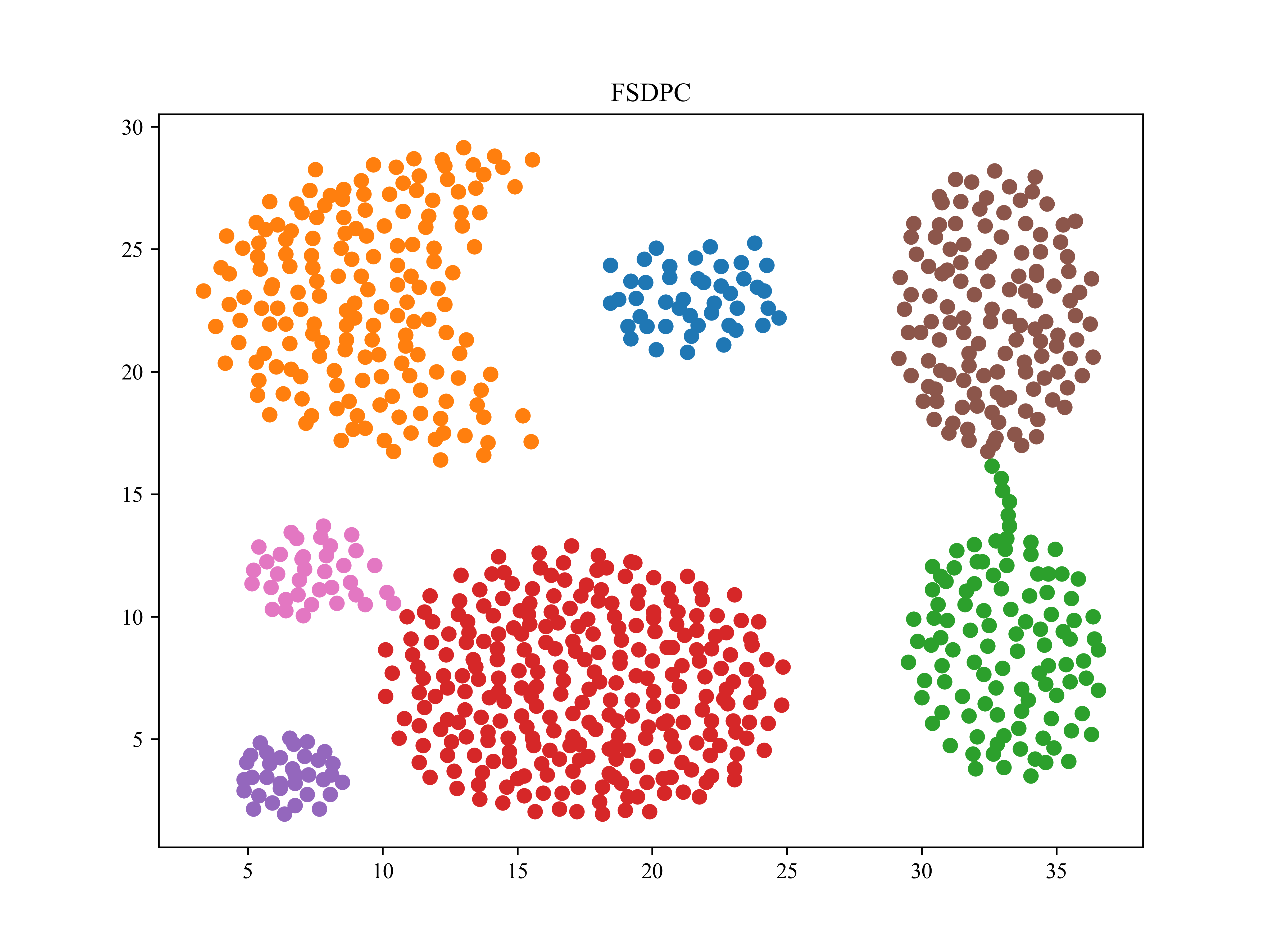}}
\subfloat{\includegraphics[width = 0.2\textwidth]{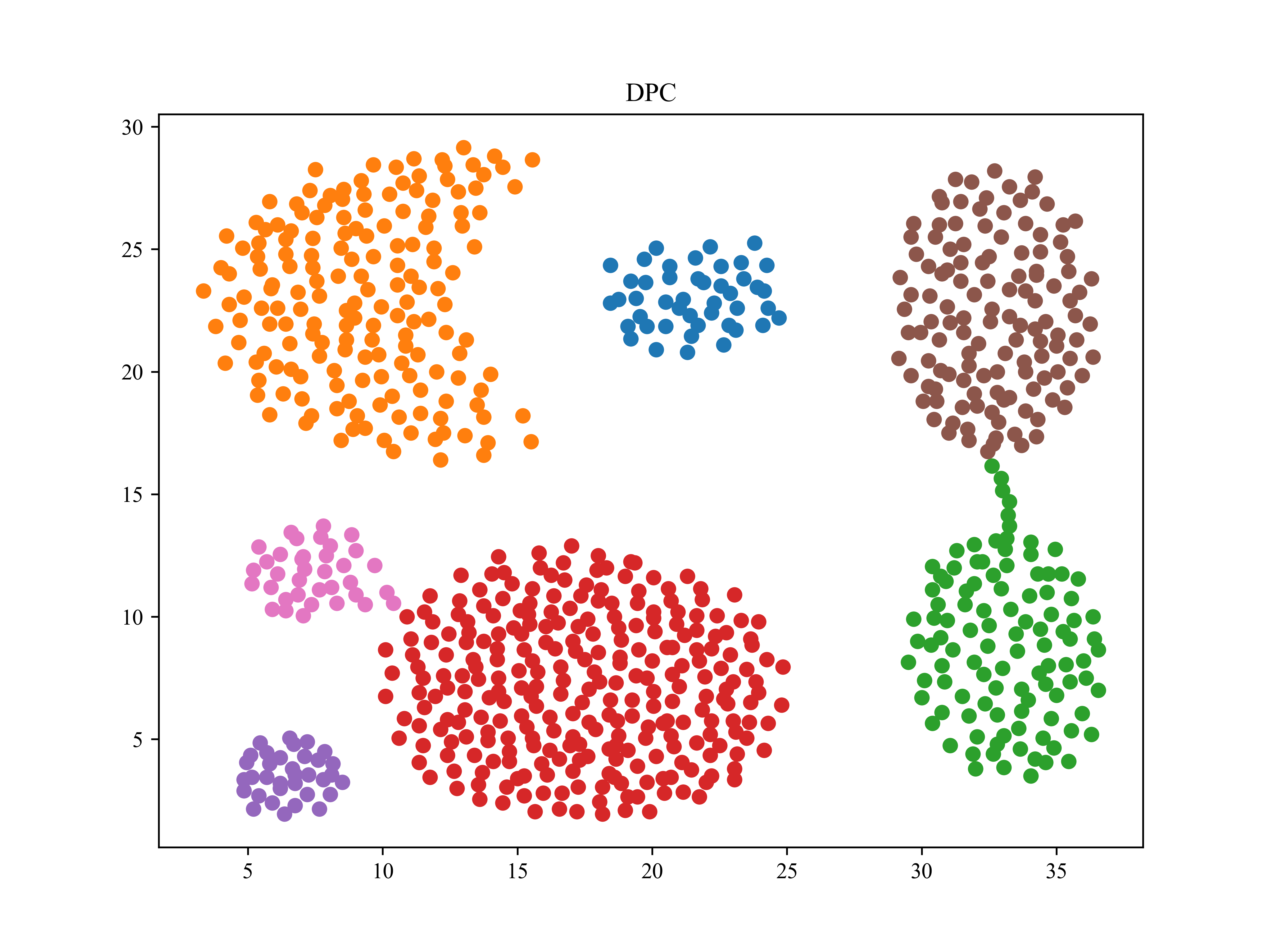}}
\subfloat{\includegraphics[width = 0.2\textwidth]{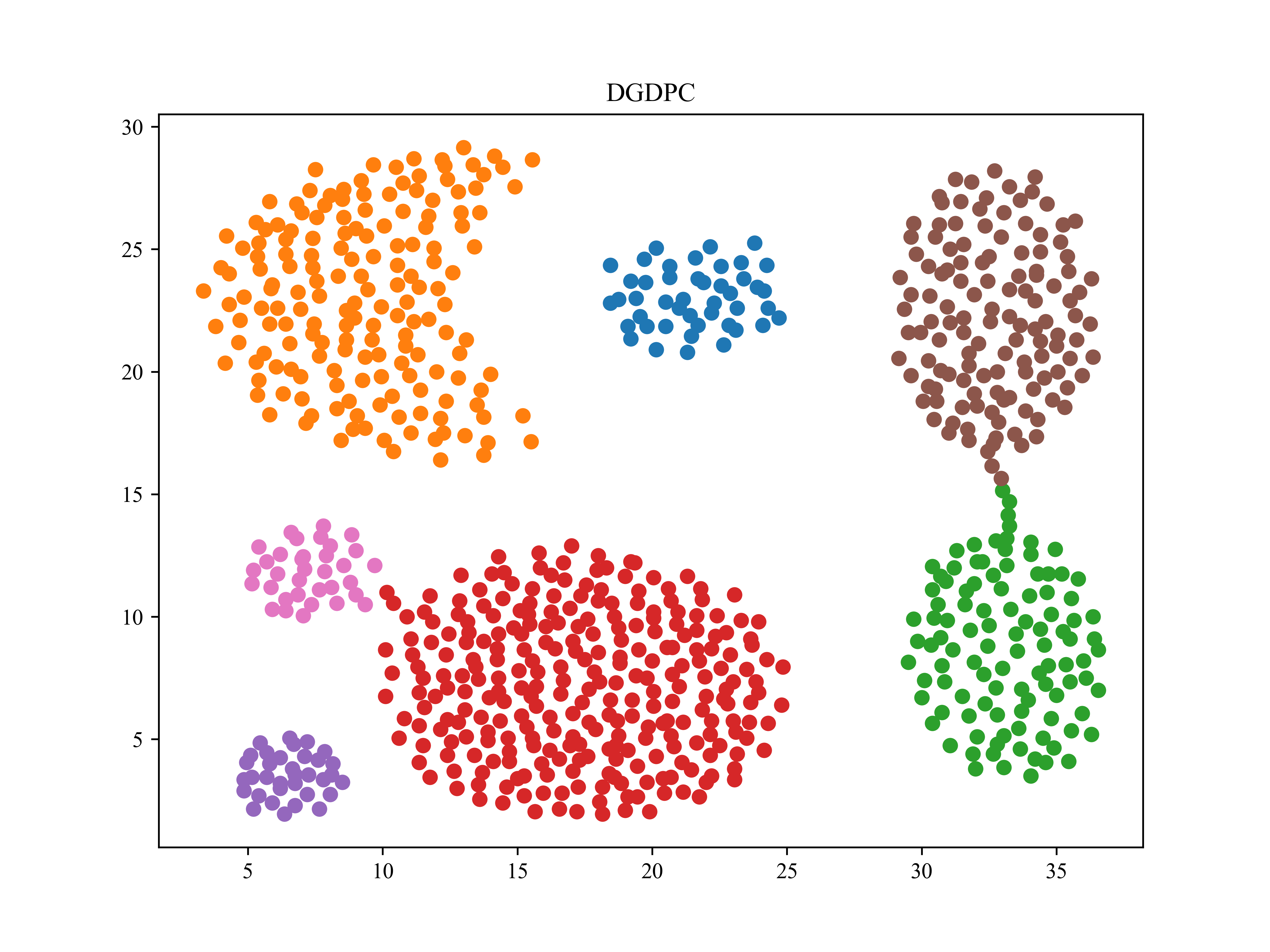}}\\
\subfloat{\includegraphics[width = 0.2\textwidth]{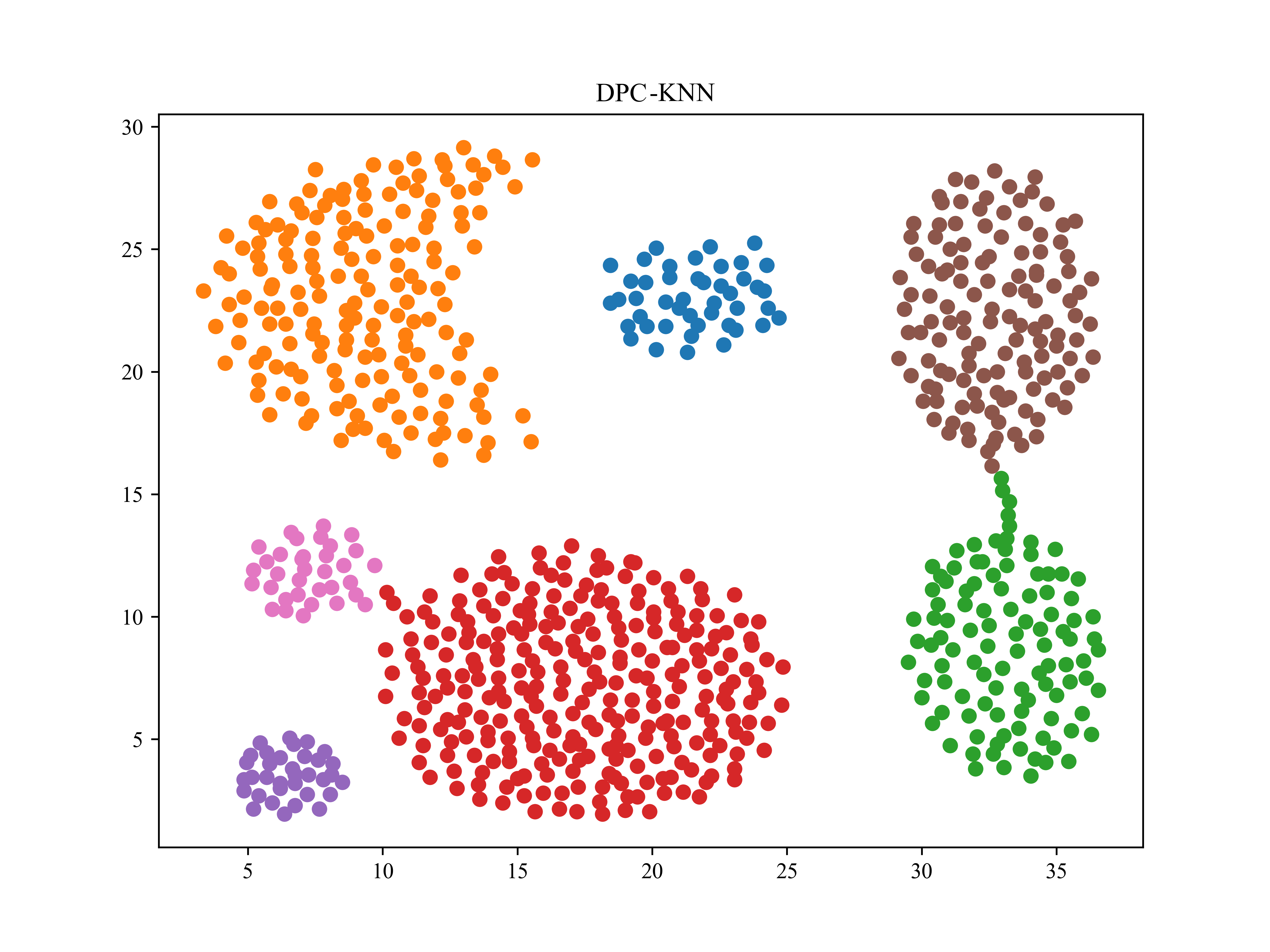}\centering}
\subfloat{\includegraphics[width = 0.2\textwidth]{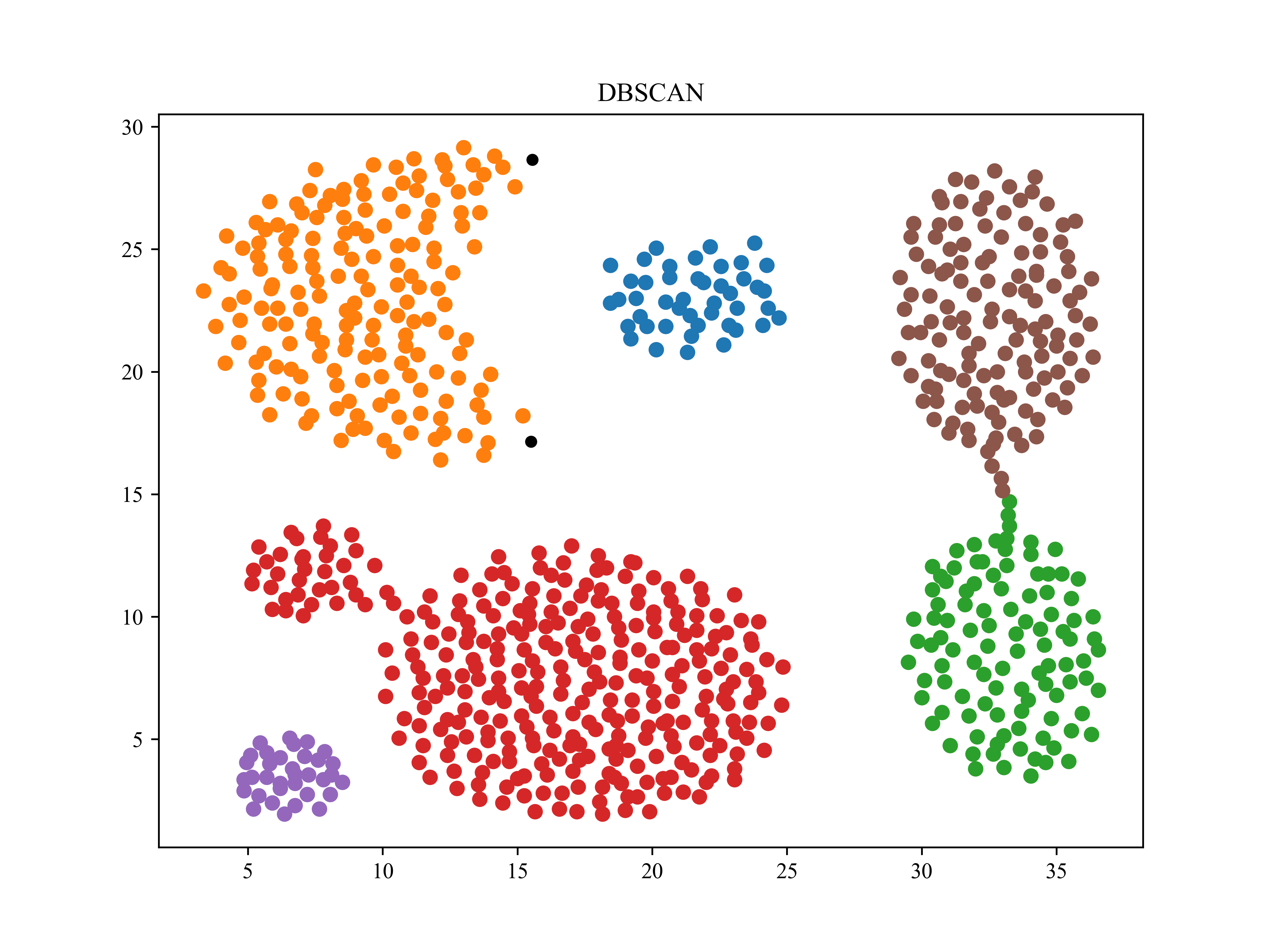}\centering}	
\subfloat{\includegraphics[width = 0.2\textwidth]{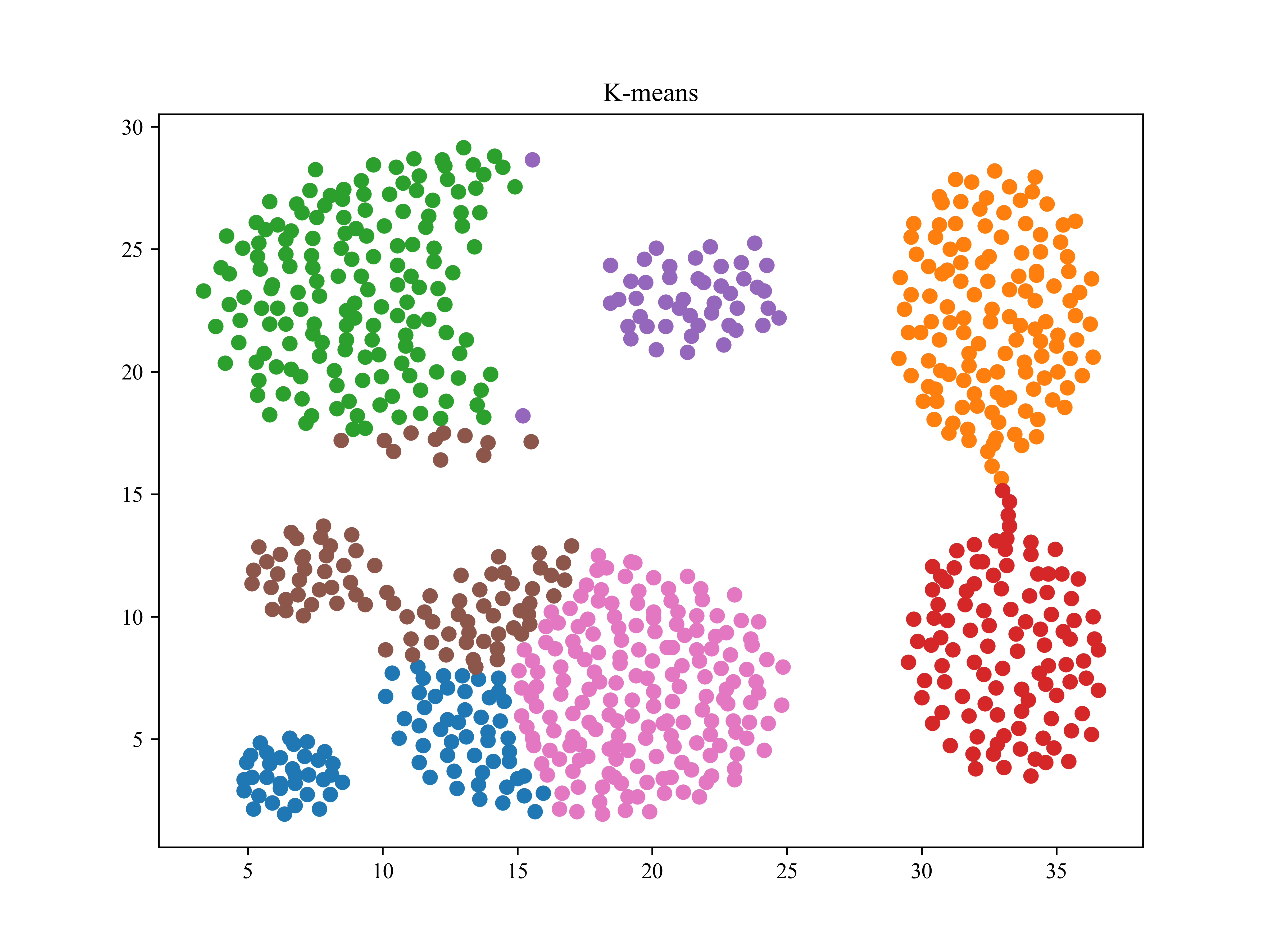}\centering}	
\caption{\textbf {Clustering results of the seven algorithms on Aggregation dataset. }}
\label{fig:labe4}
\end{figure*}

\begin{figure*}[ht]
\centering
	\subfloat{\includegraphics[width = 0.2\textwidth]{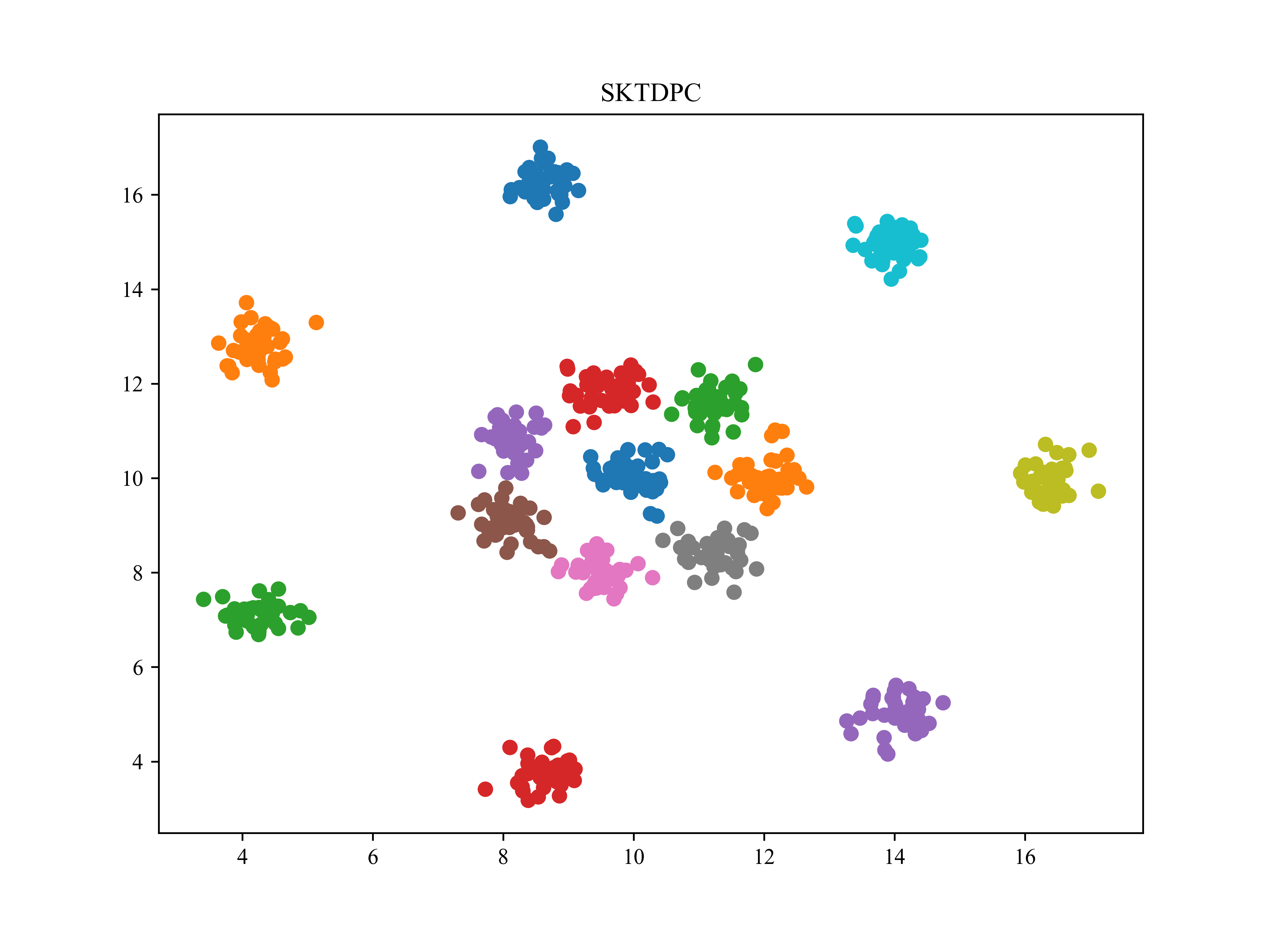}}
	\subfloat{\includegraphics[width = 0.2\textwidth]{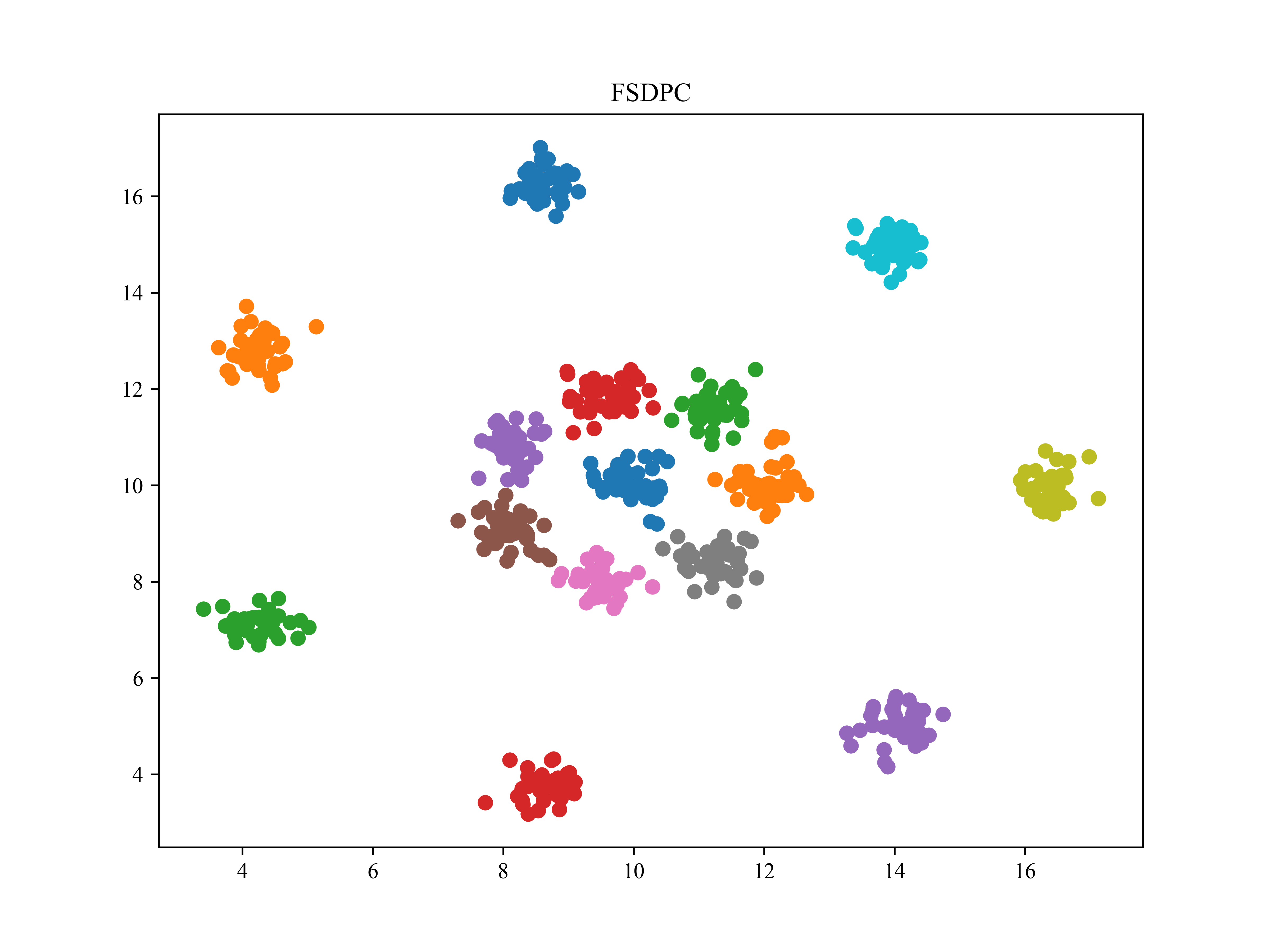}}
\subfloat{\includegraphics[width = 0.2\textwidth]{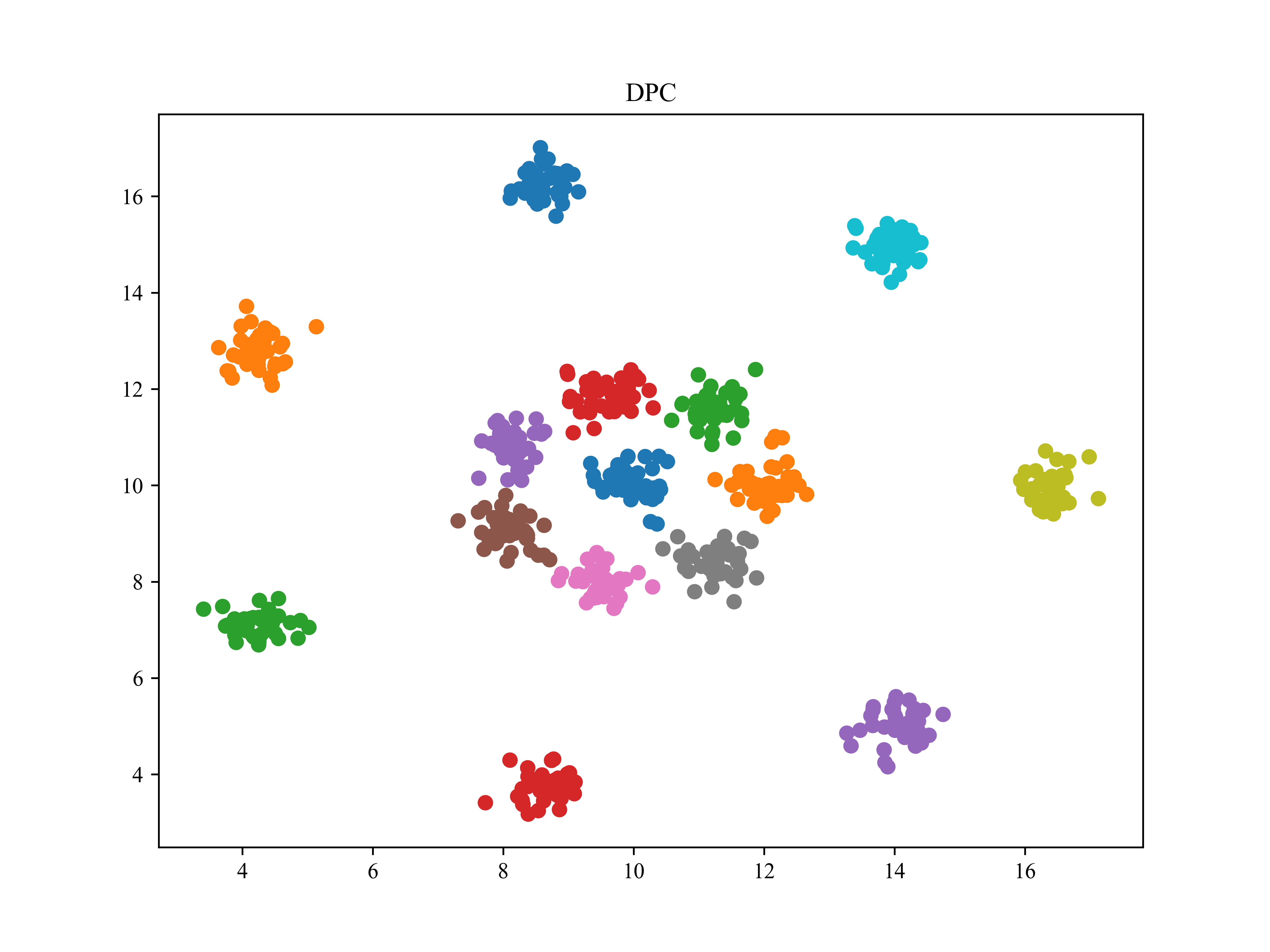}}
\subfloat{\includegraphics[width = 0.2\textwidth]{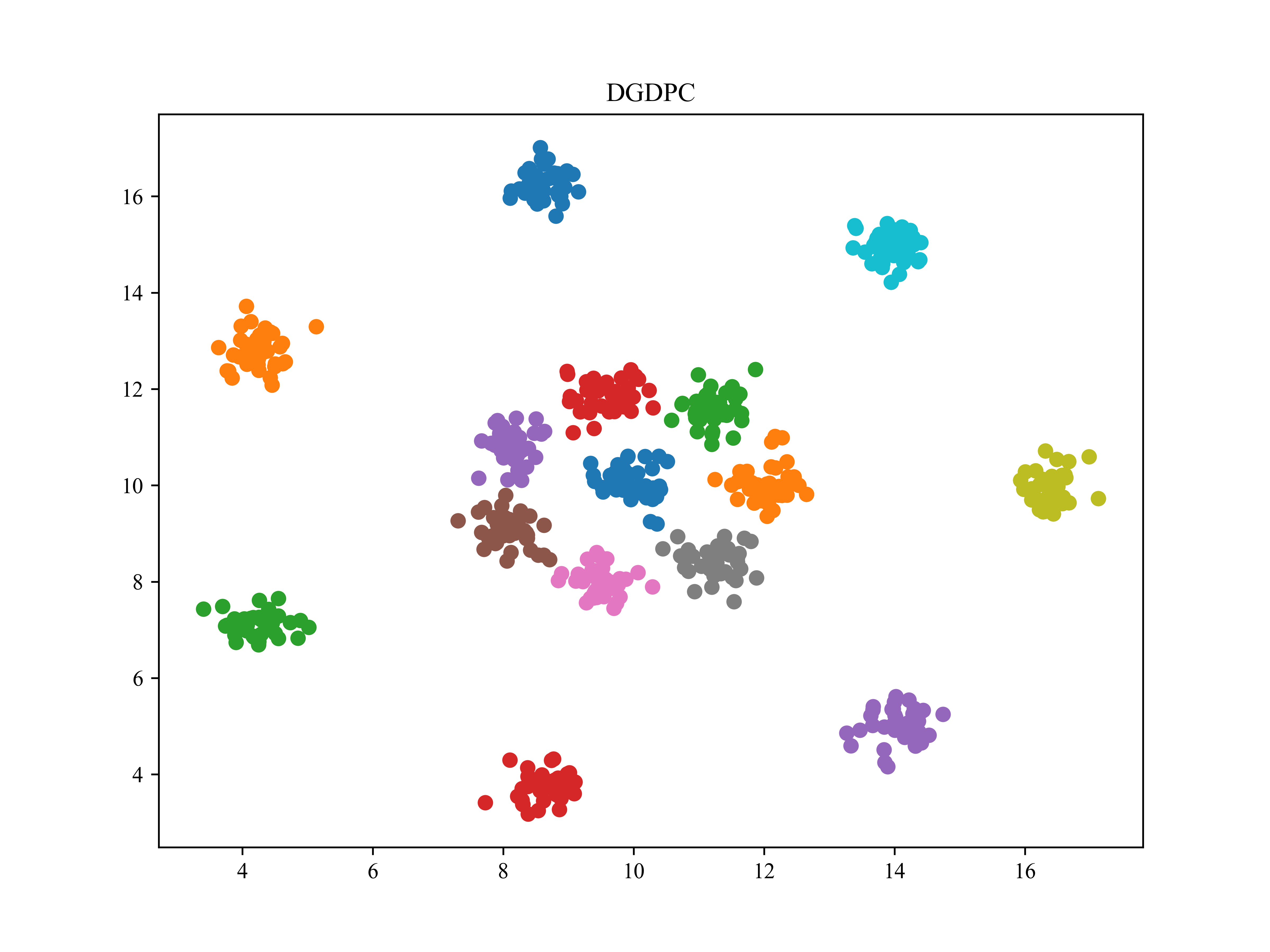}}\\
\subfloat{\includegraphics[width = 0.2\textwidth]{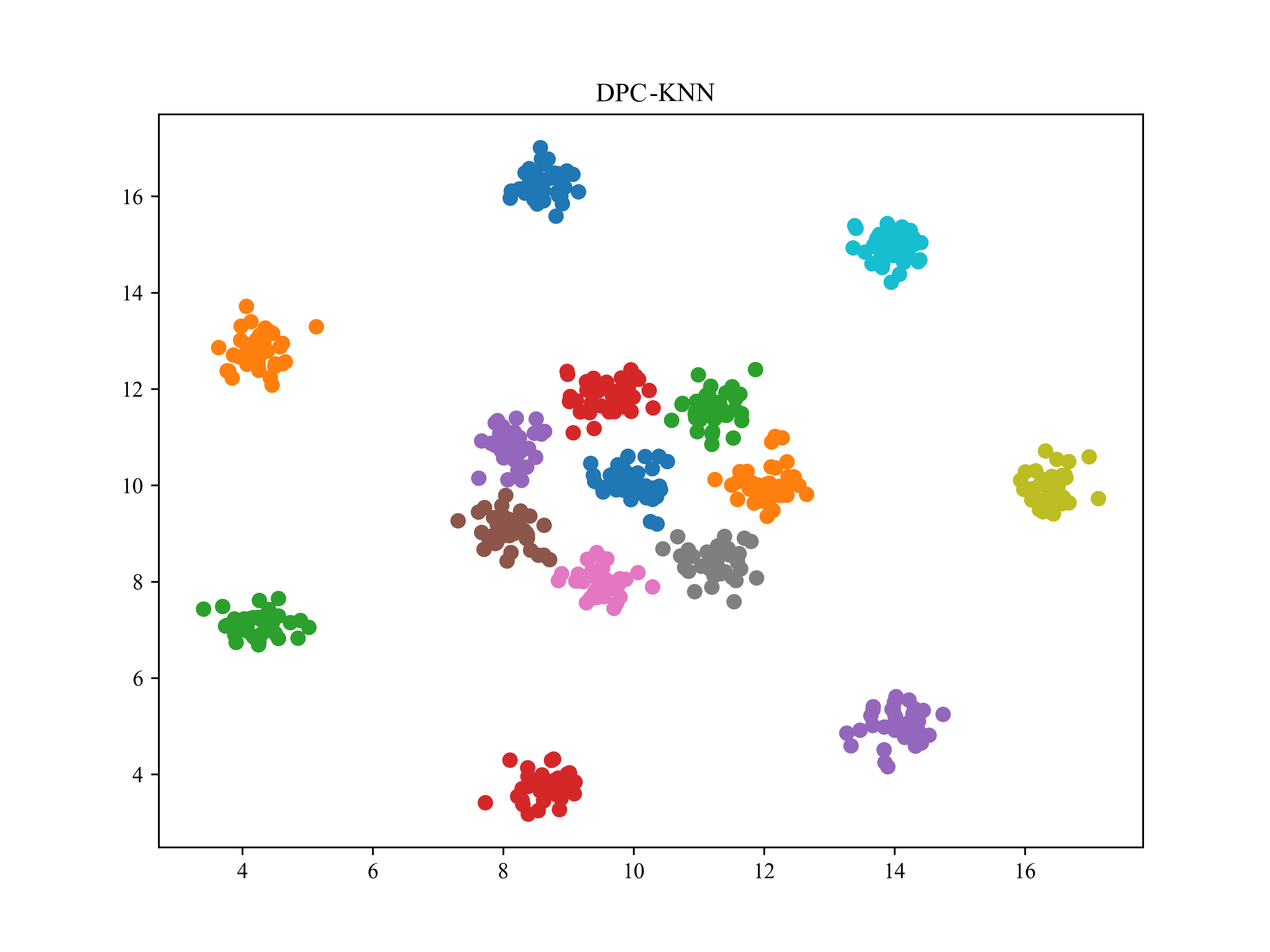}\centering}
\subfloat{\includegraphics[width = 0.2\textwidth]{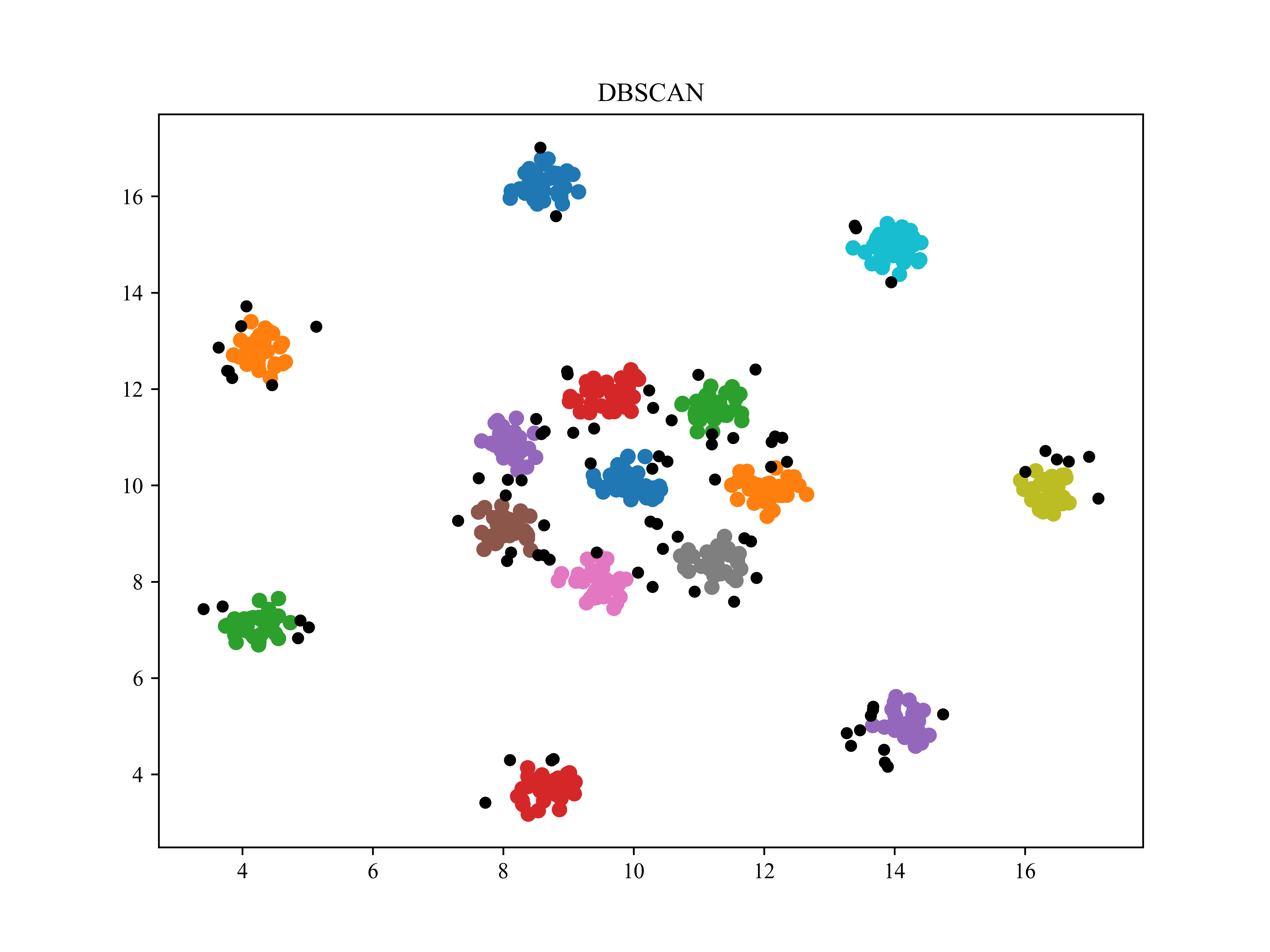}\centering}	
\subfloat{\includegraphics[width = 0.2\textwidth]{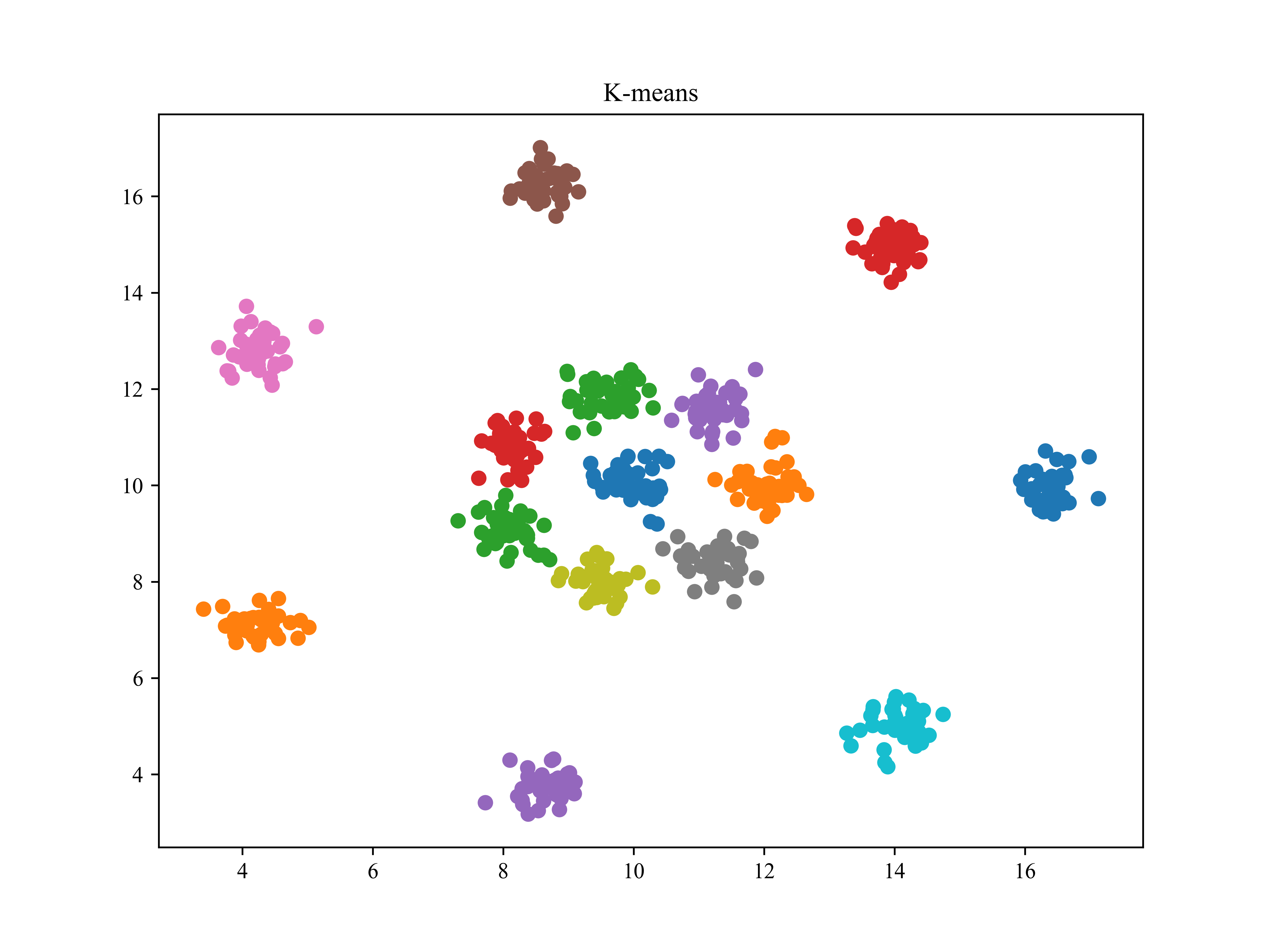}\centering}	
\caption{\textbf {Clustering results of the seven algorithms on R15 dataset.} }
\label{fig:labe5}
\end{figure*}

\begin{figure*}[ht]
\centering
	\subfloat{\includegraphics[width = 0.2\textwidth]{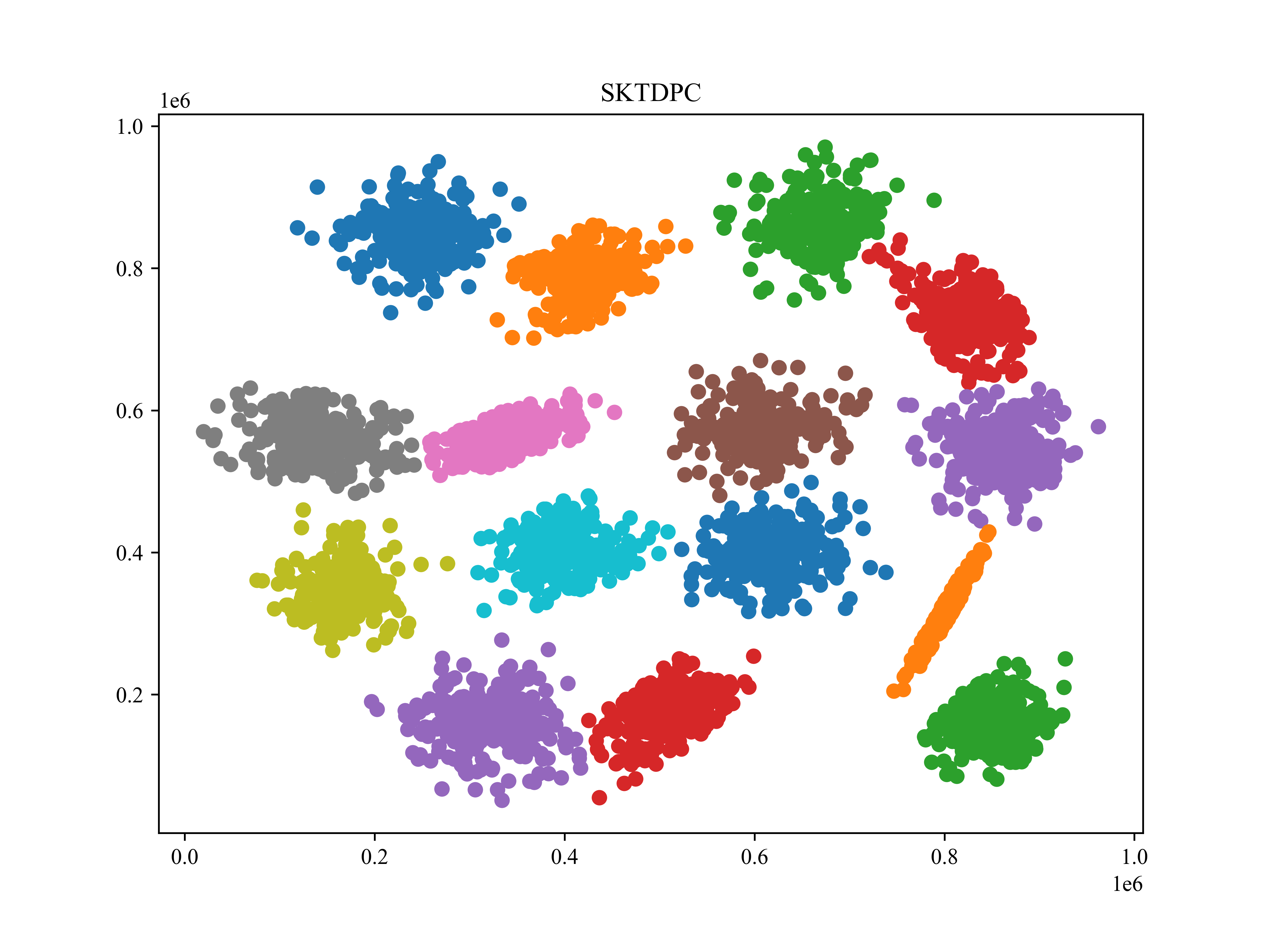}}
	\subfloat{\includegraphics[width = 0.2\textwidth]{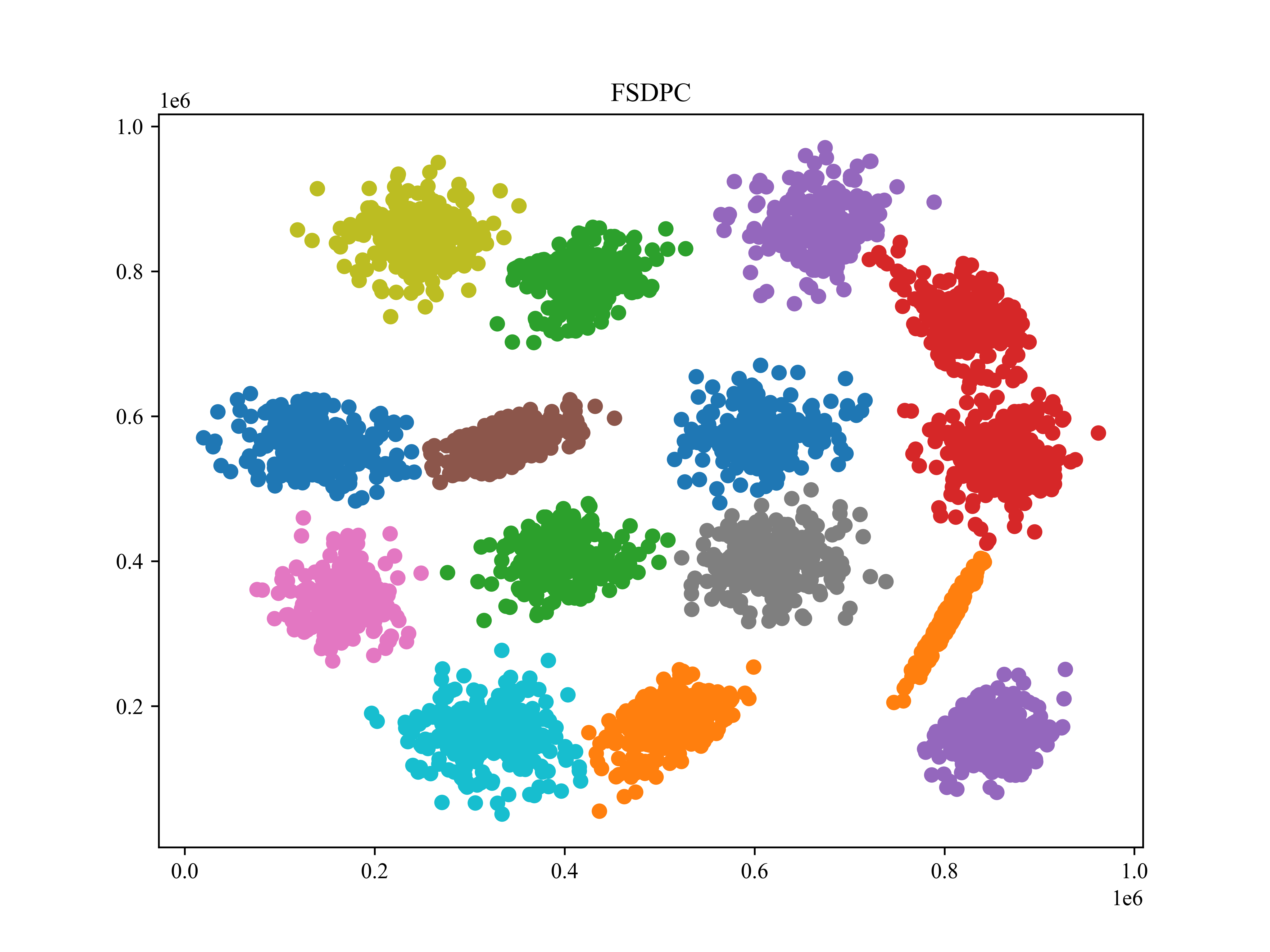}}
\subfloat{\includegraphics[width = 0.2\textwidth]{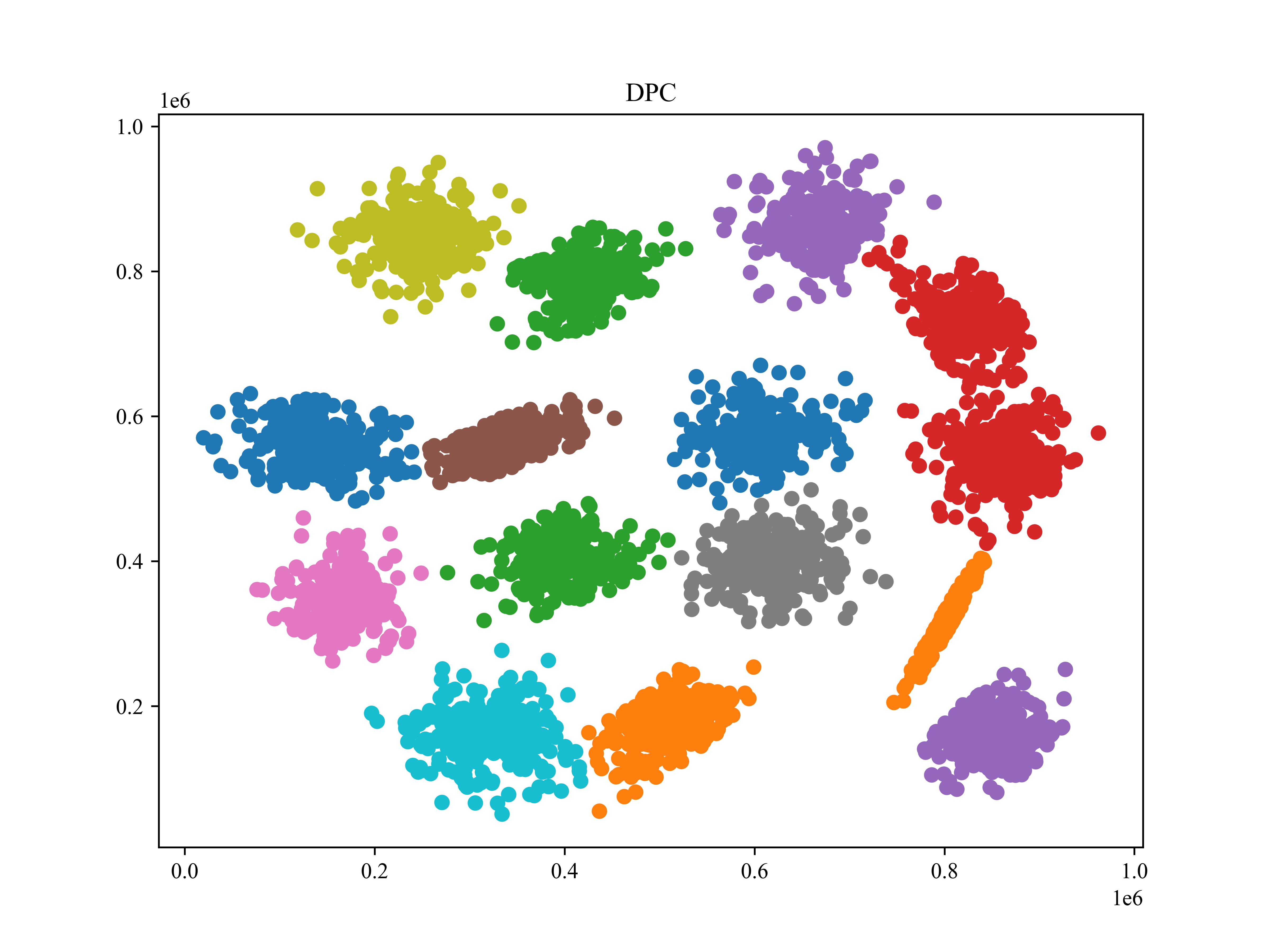}}
\subfloat{\includegraphics[width = 0.2\textwidth]{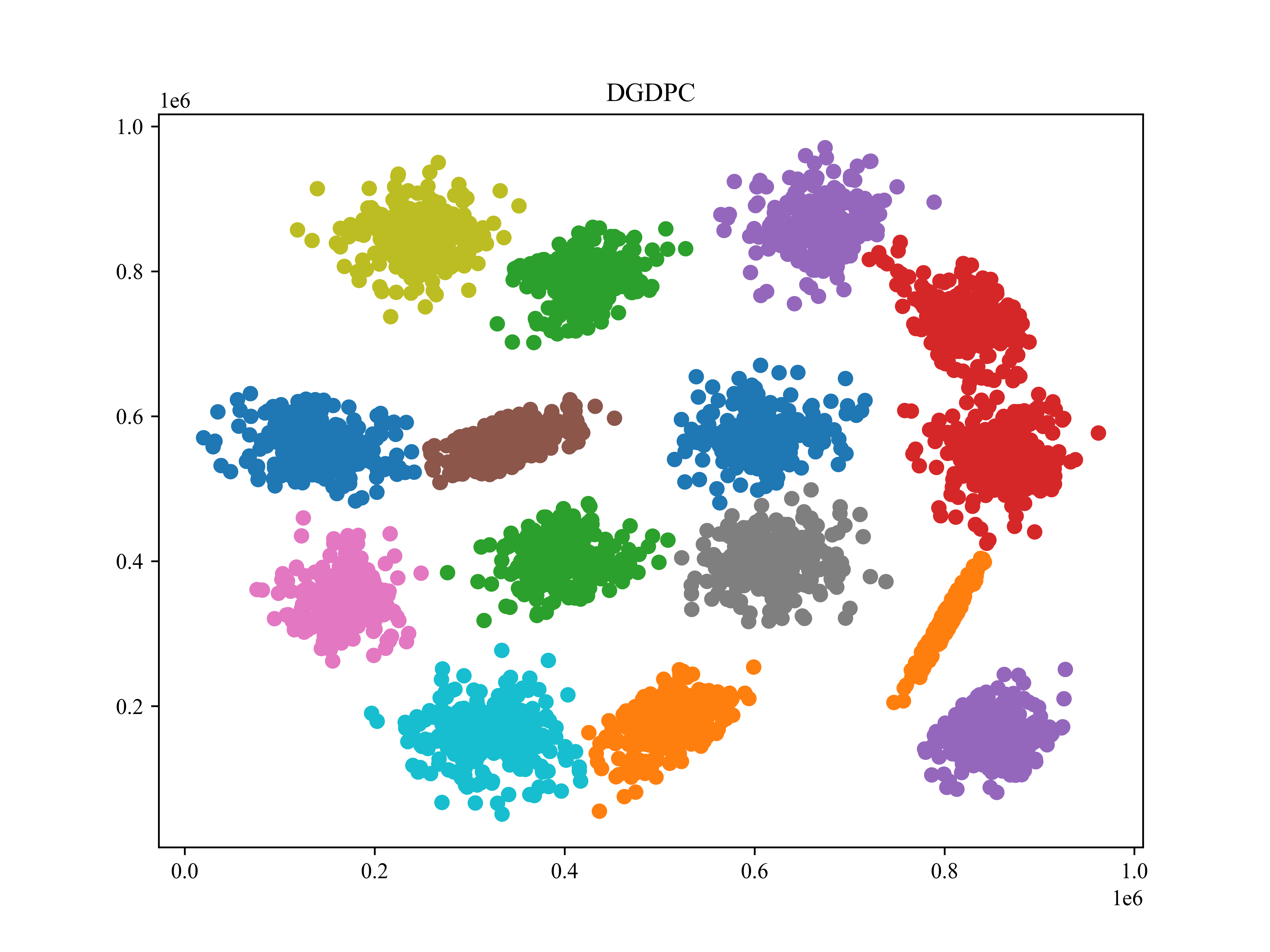}}\\
\subfloat{\includegraphics[width = 0.2\textwidth]{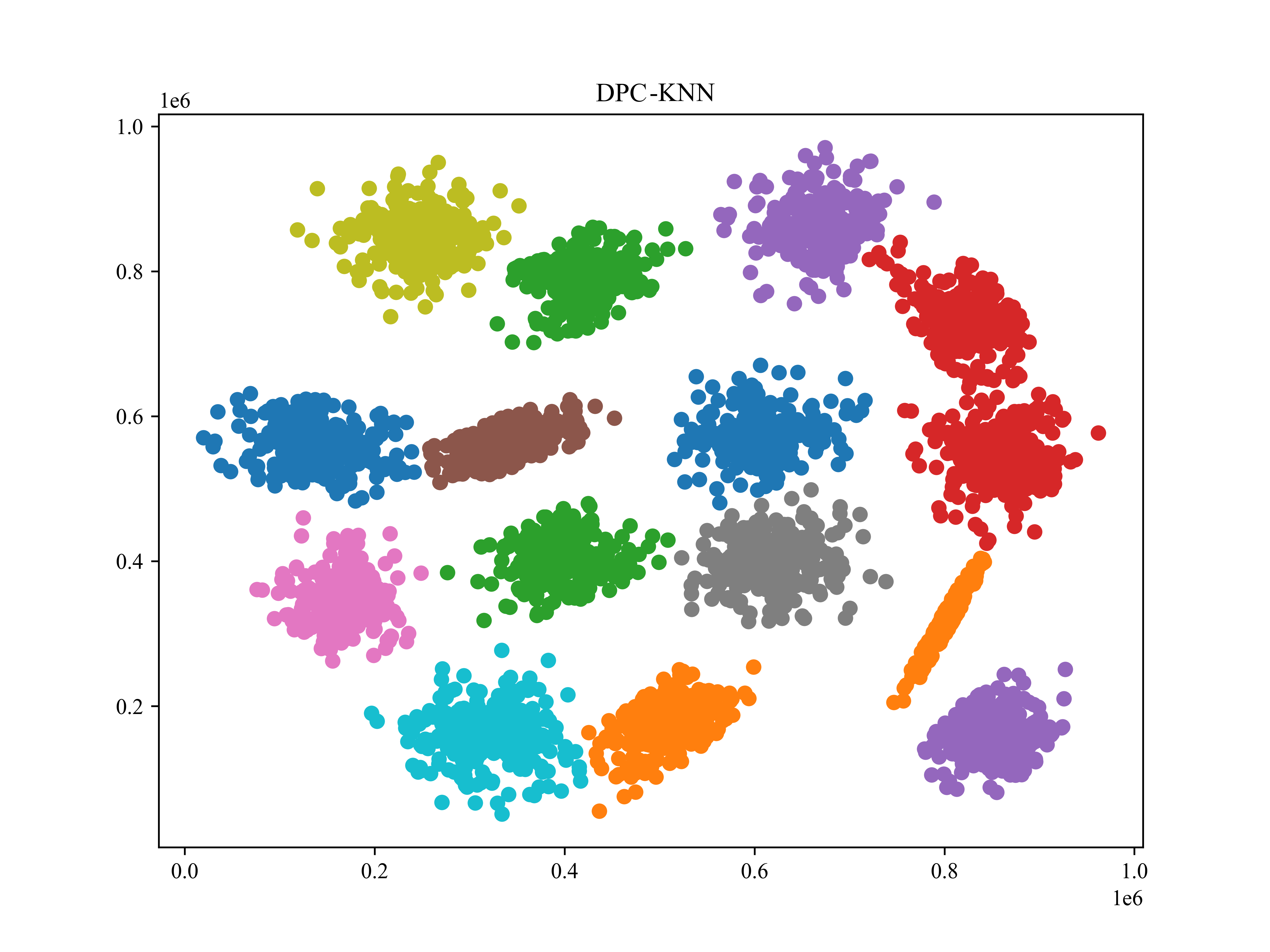}\centering}
\subfloat{\includegraphics[width = 0.2\textwidth]{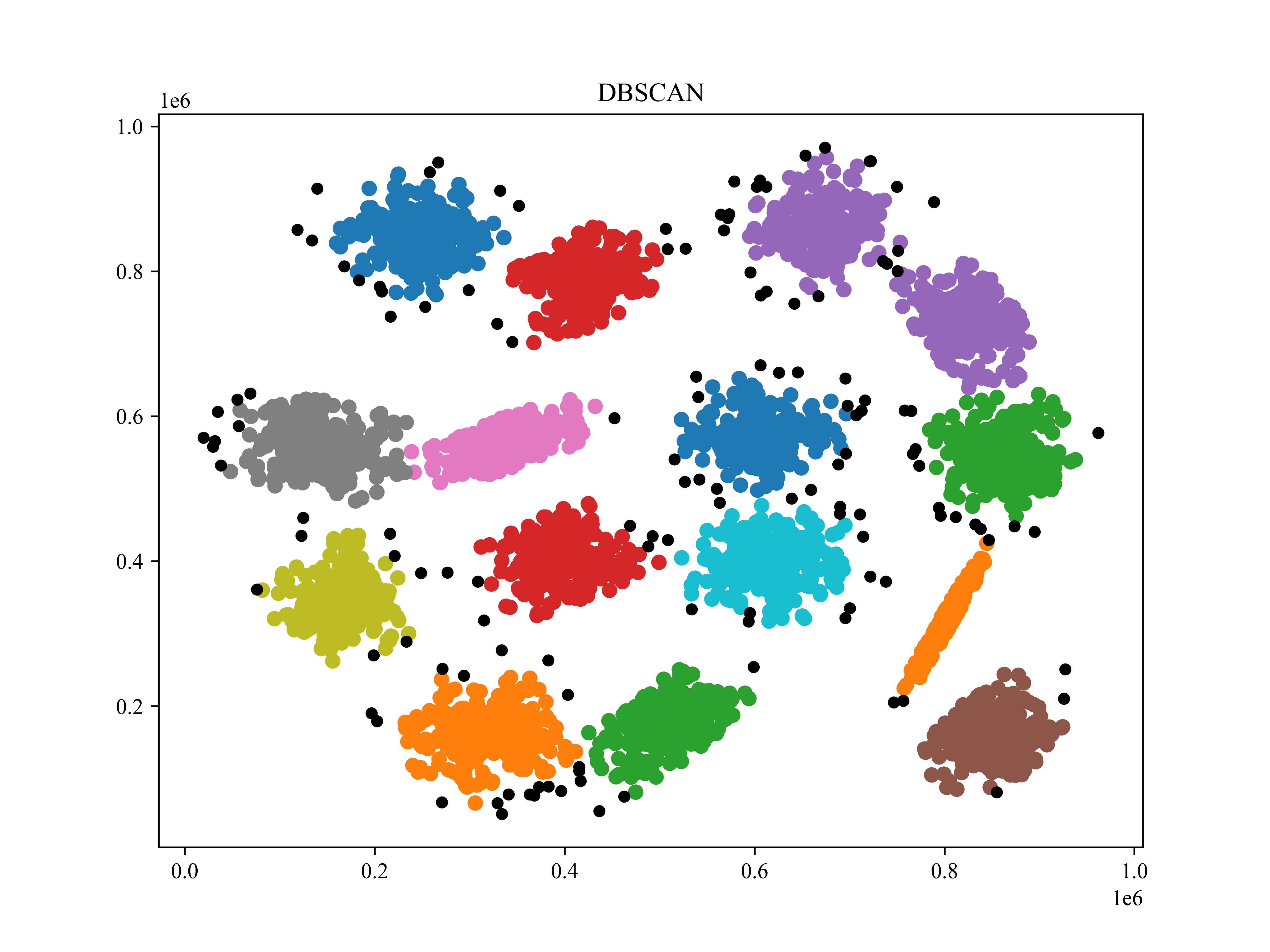}\centering}	
\subfloat{\includegraphics[width = 0.2\textwidth]{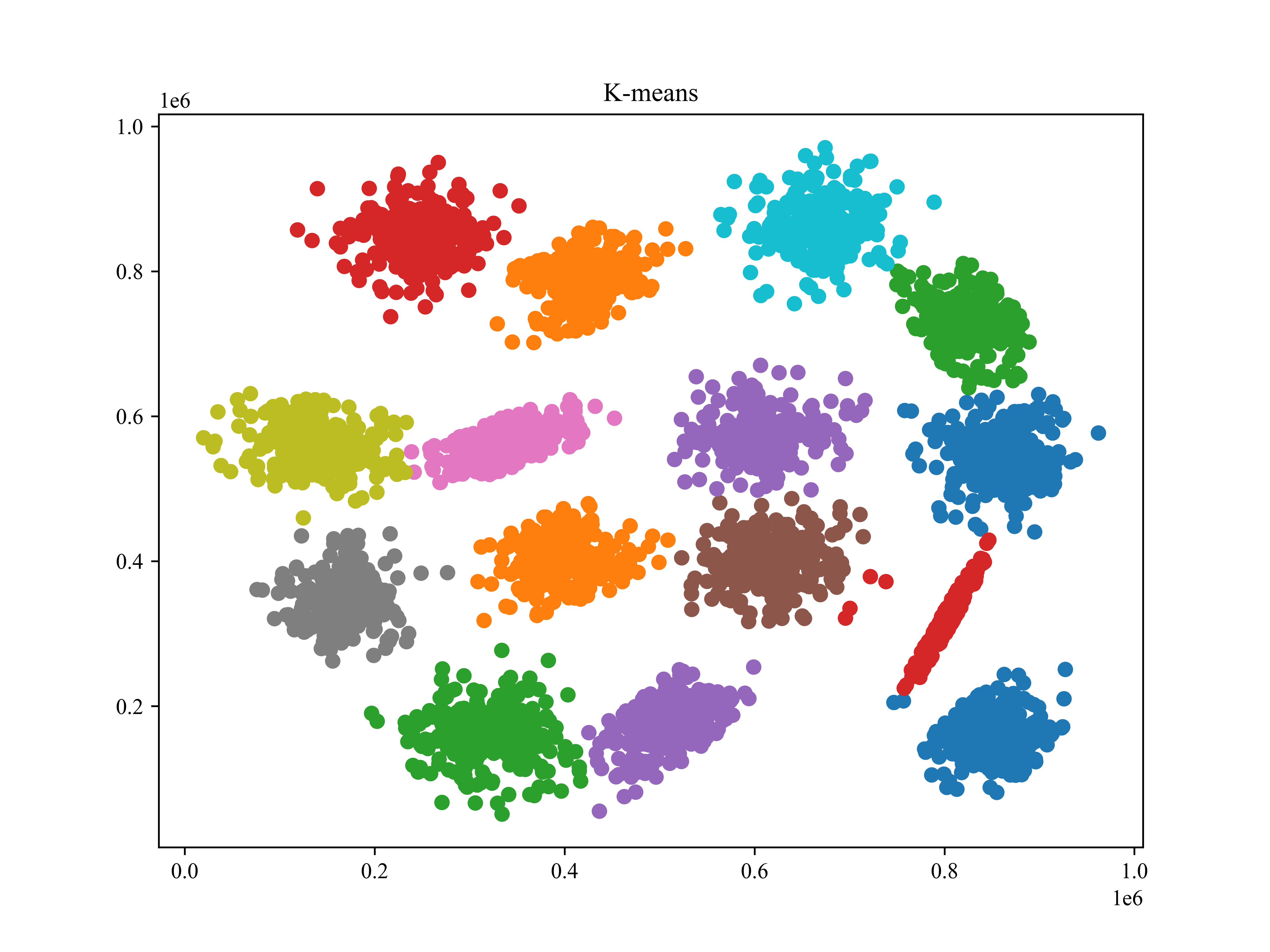}\centering}	
\caption{\textbf {Clustering results of the seven algorithms on S1 dataset.} }
\label{fig:labe6}
\end{figure*}

\begin{figure*}[ht]
\centering
	\subfloat{\includegraphics[width = 0.2\textwidth]{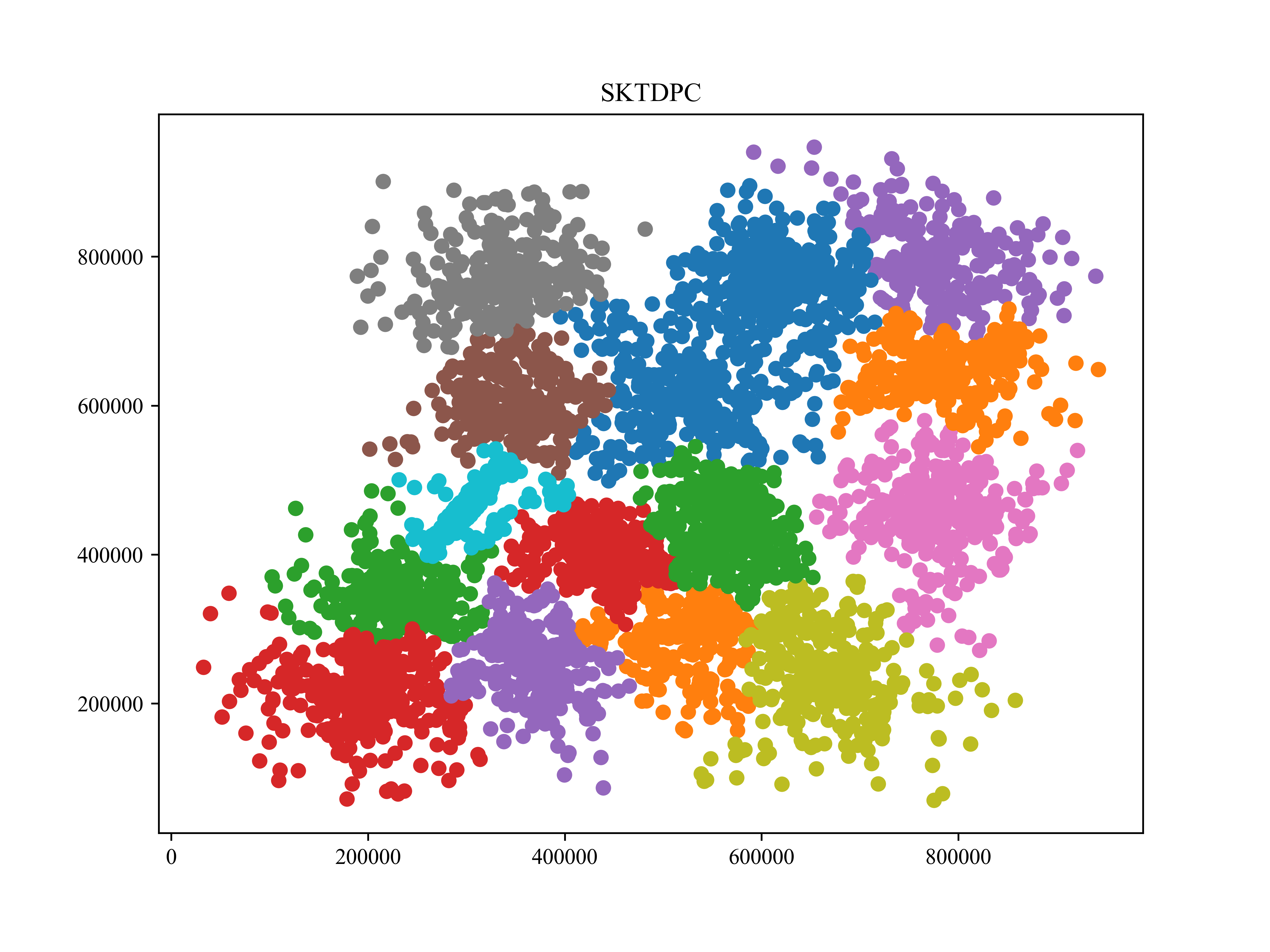}}
	\subfloat{\includegraphics[width = 0.2\textwidth]{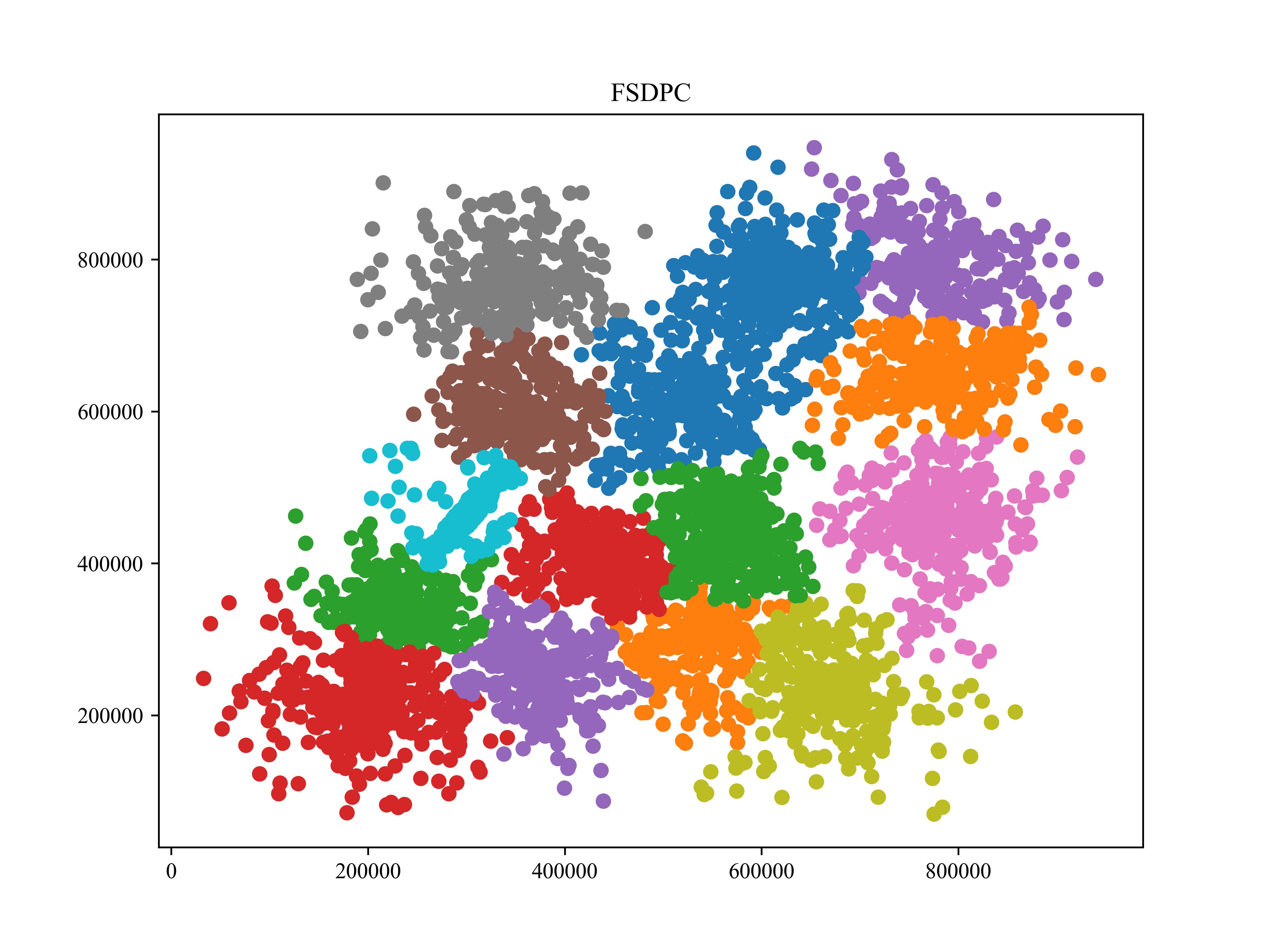}}
\subfloat{\includegraphics[width = 0.2\textwidth]{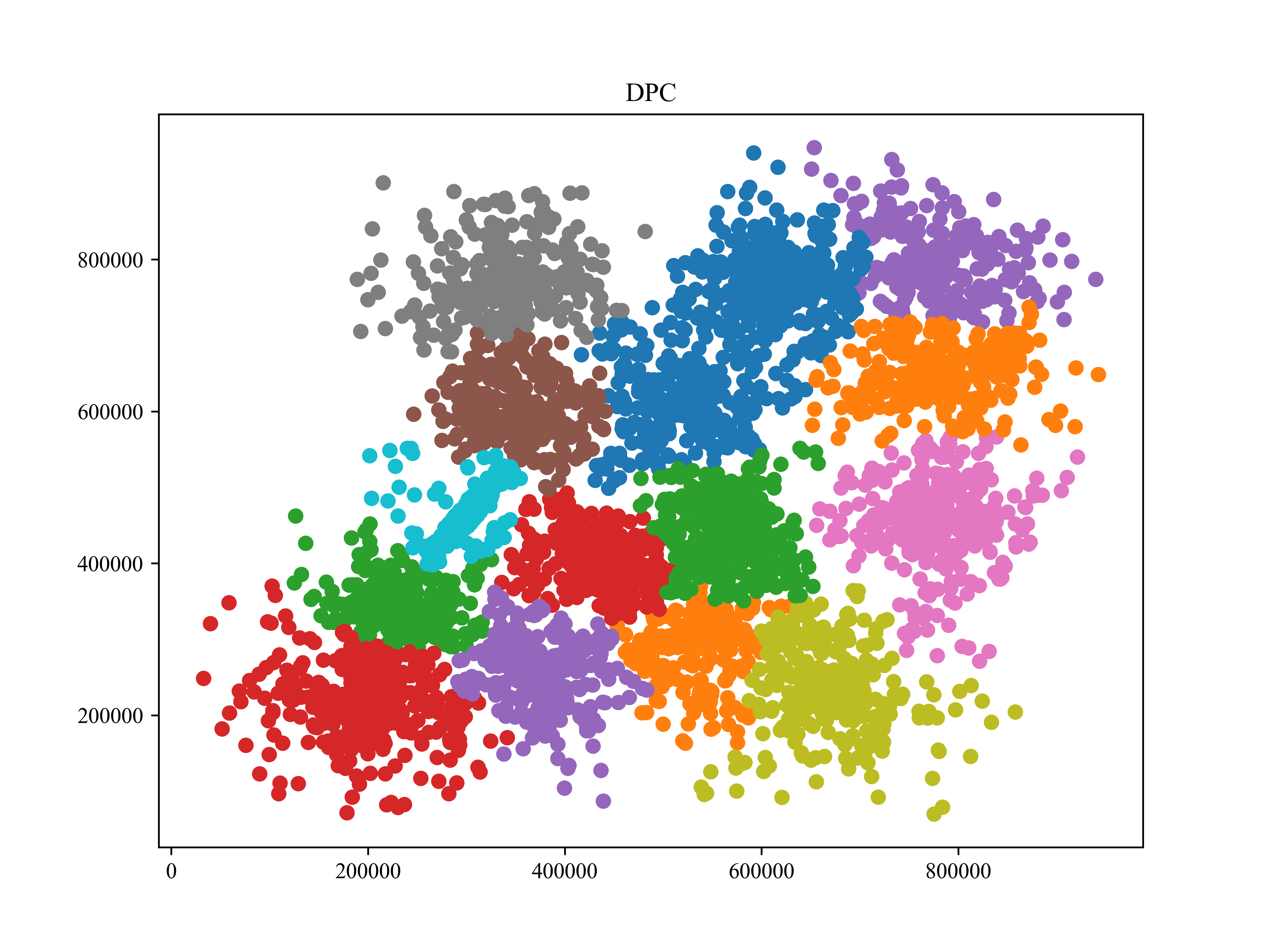}}
\subfloat{\includegraphics[width = 0.2\textwidth]{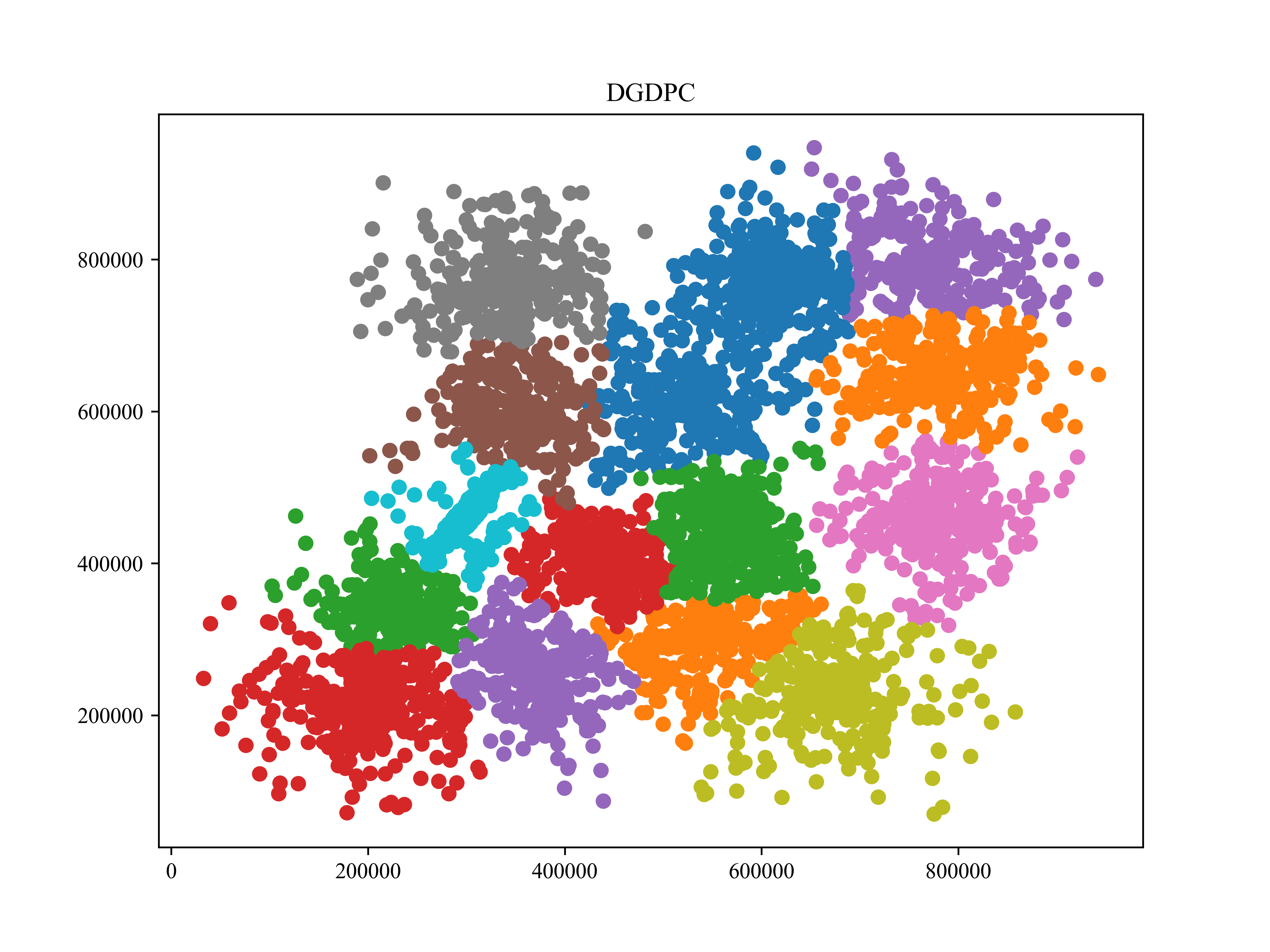}}\\
\subfloat{\includegraphics[width = 0.2\textwidth]{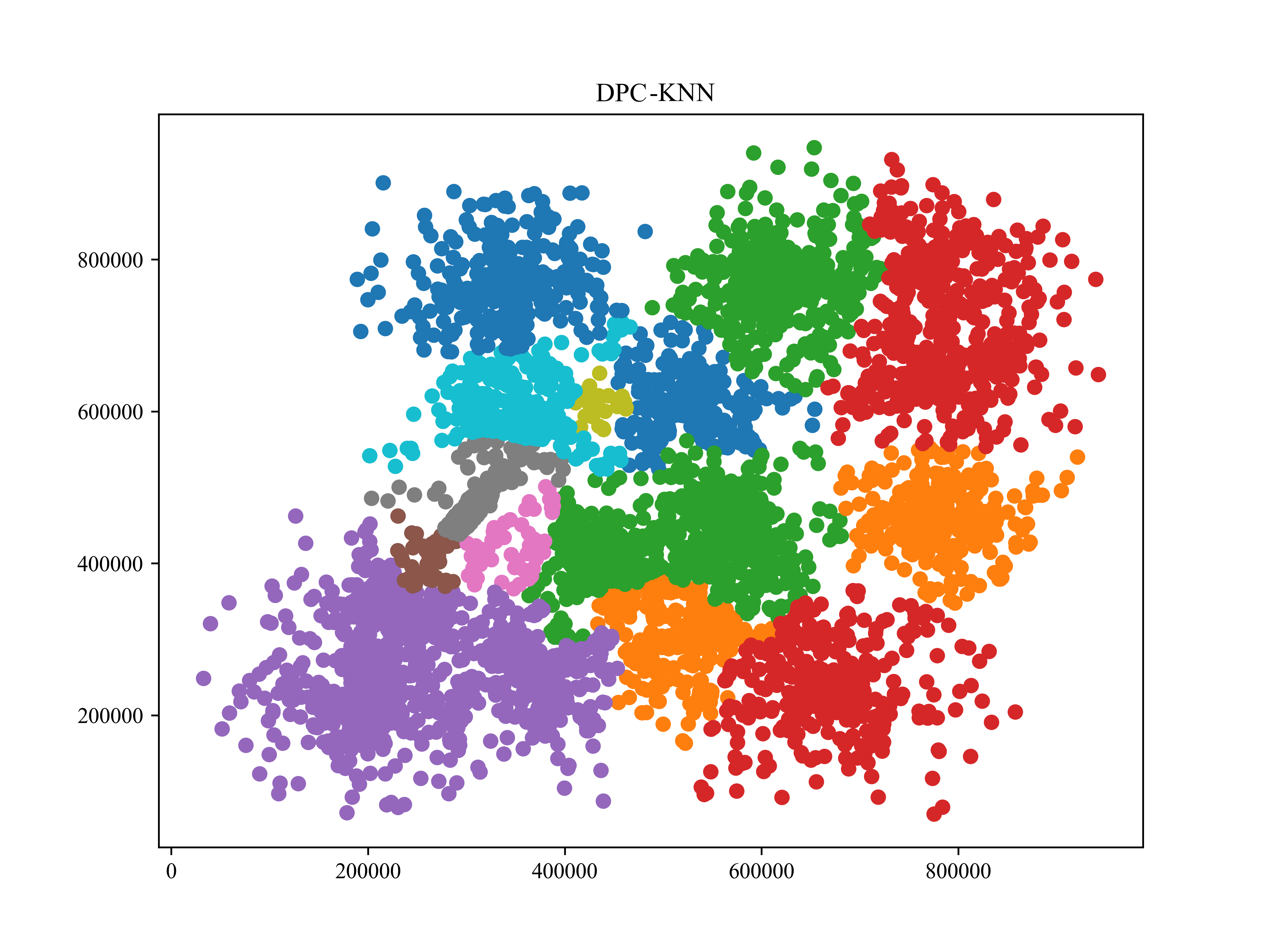}\centering}
\subfloat{\includegraphics[width = 0.2\textwidth]{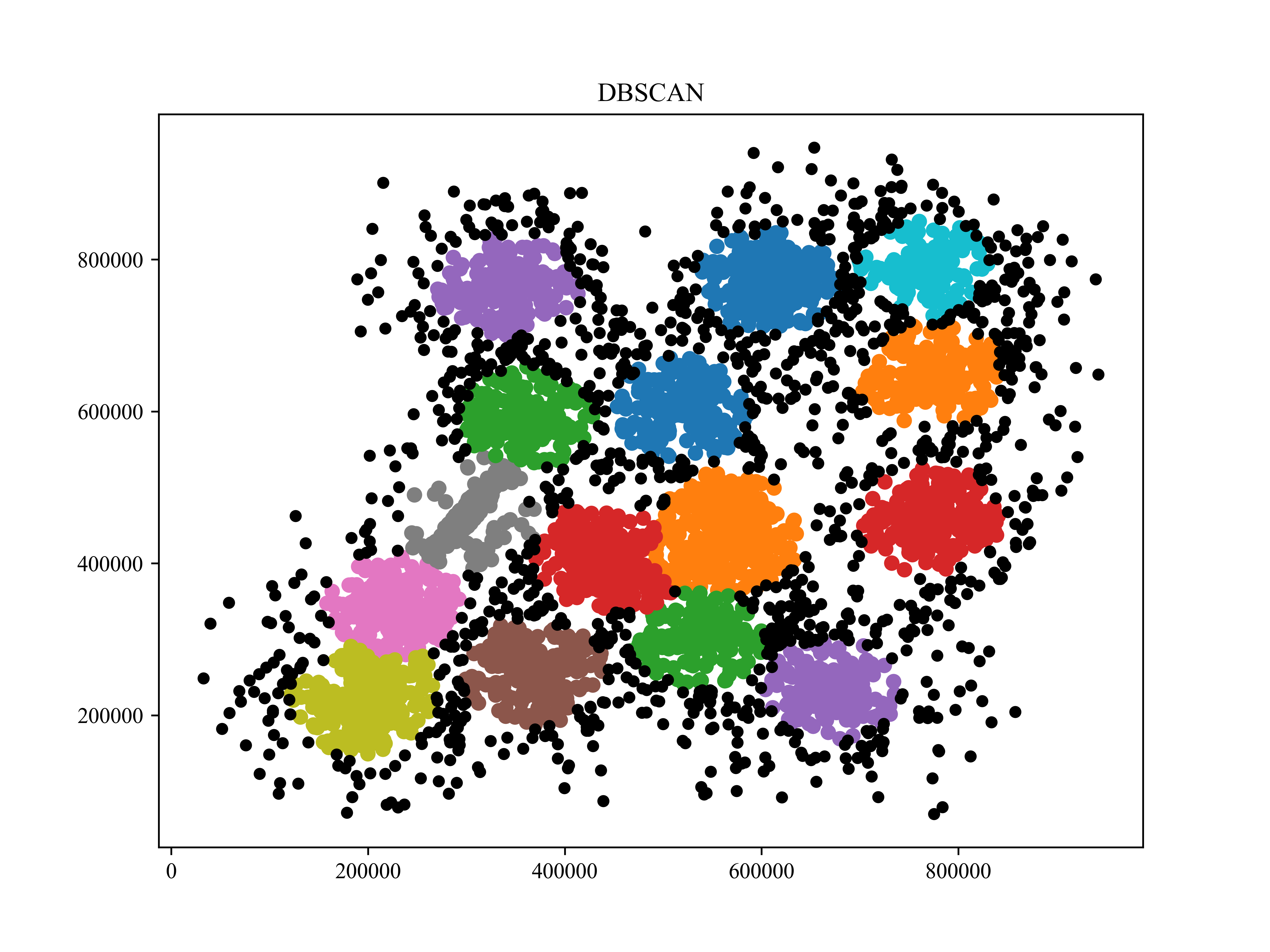}\centering}
\subfloat{\includegraphics[width = 0.2\textwidth]{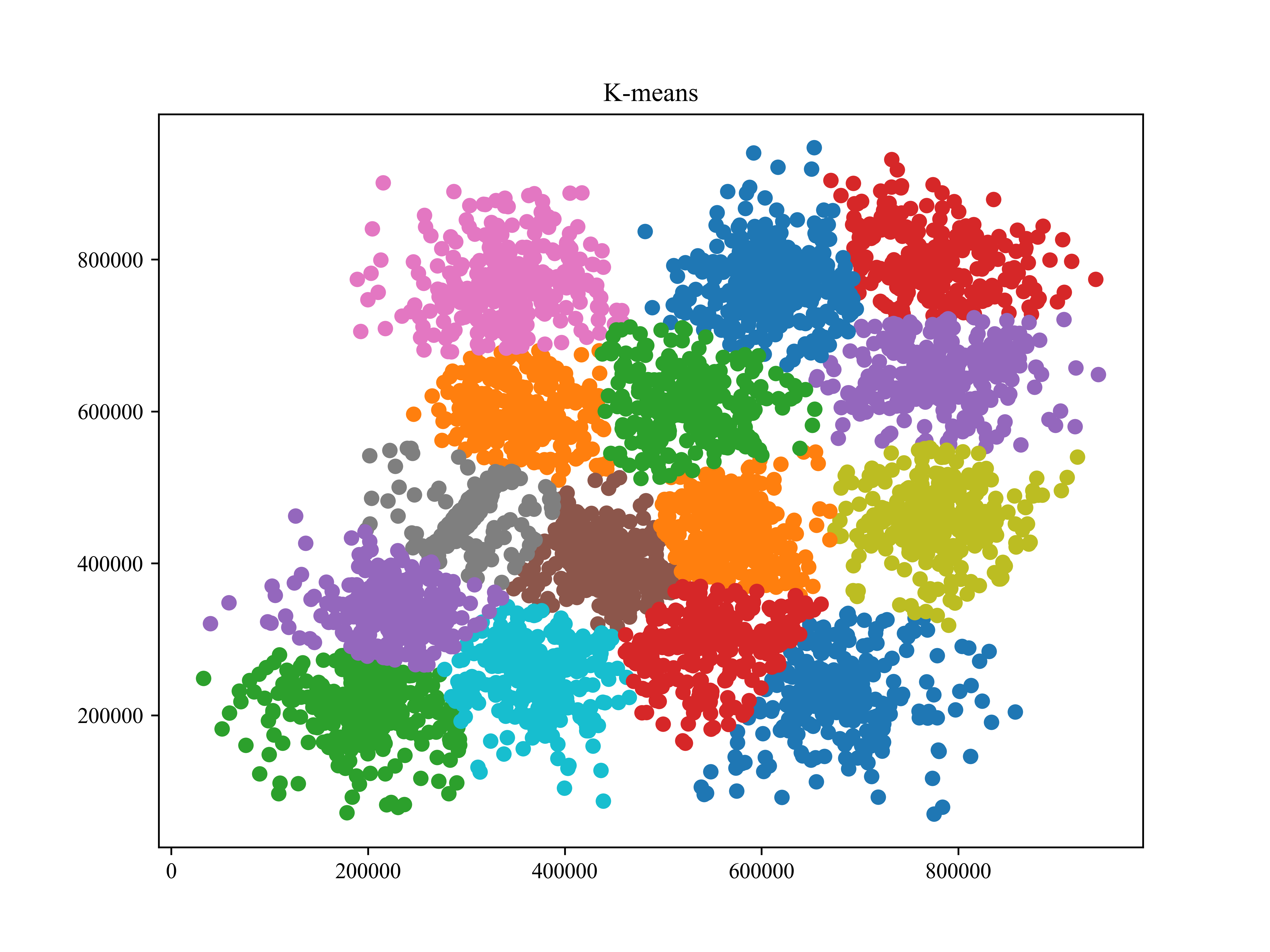}\centering}	
\caption{\textbf {Clustering results of the seven algorithms on S3 dataset.} }
\label{fig:labe7}
\end{figure*}

\begin{figure*}[ht]
\centering
	\subfloat{\includegraphics[width = 0.2\textwidth]{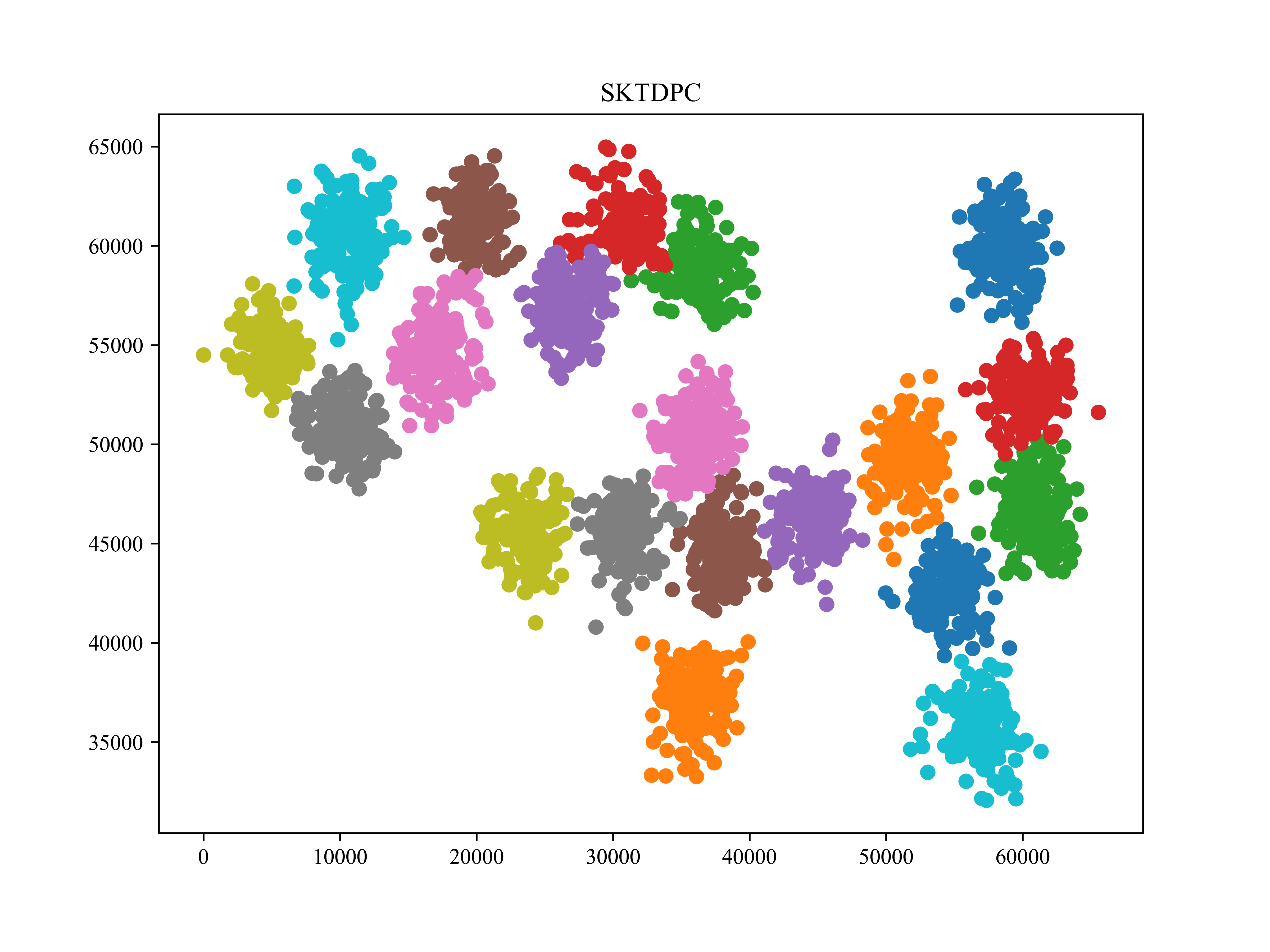}}
	\subfloat{\includegraphics[width = 0.2\textwidth]{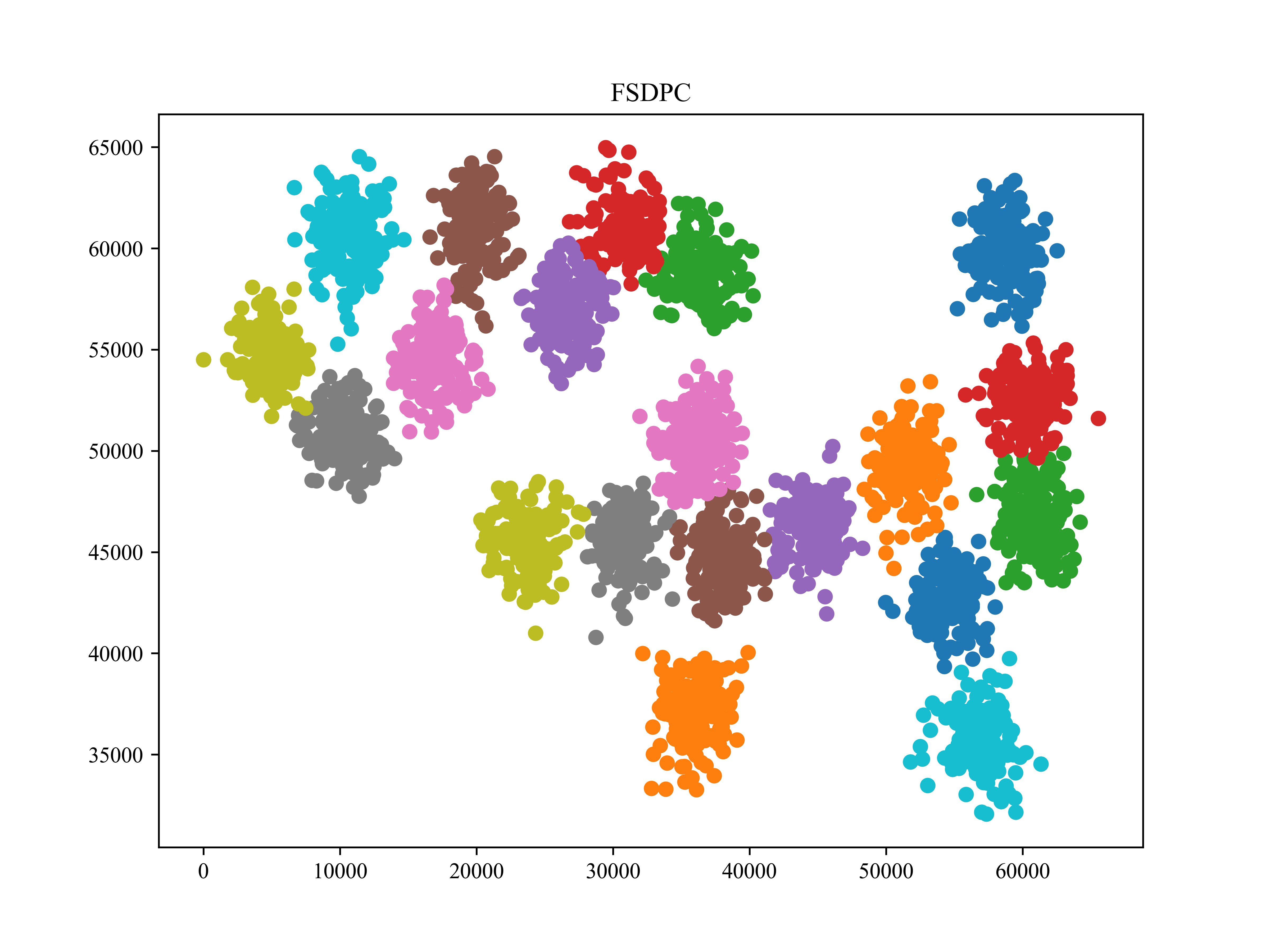}}
\subfloat{\includegraphics[width = 0.2\textwidth]{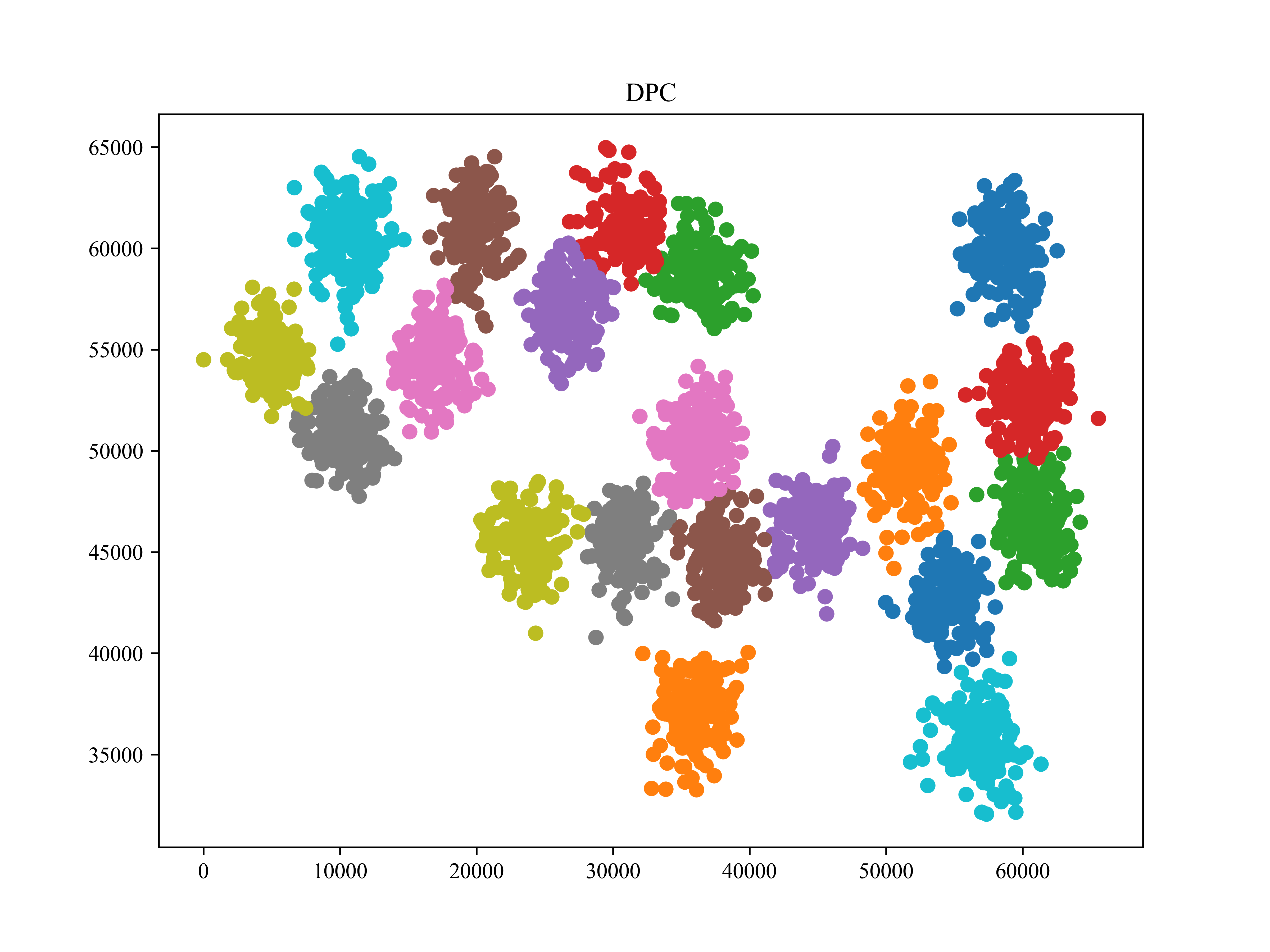}}
\subfloat{\includegraphics[width = 0.2\textwidth]{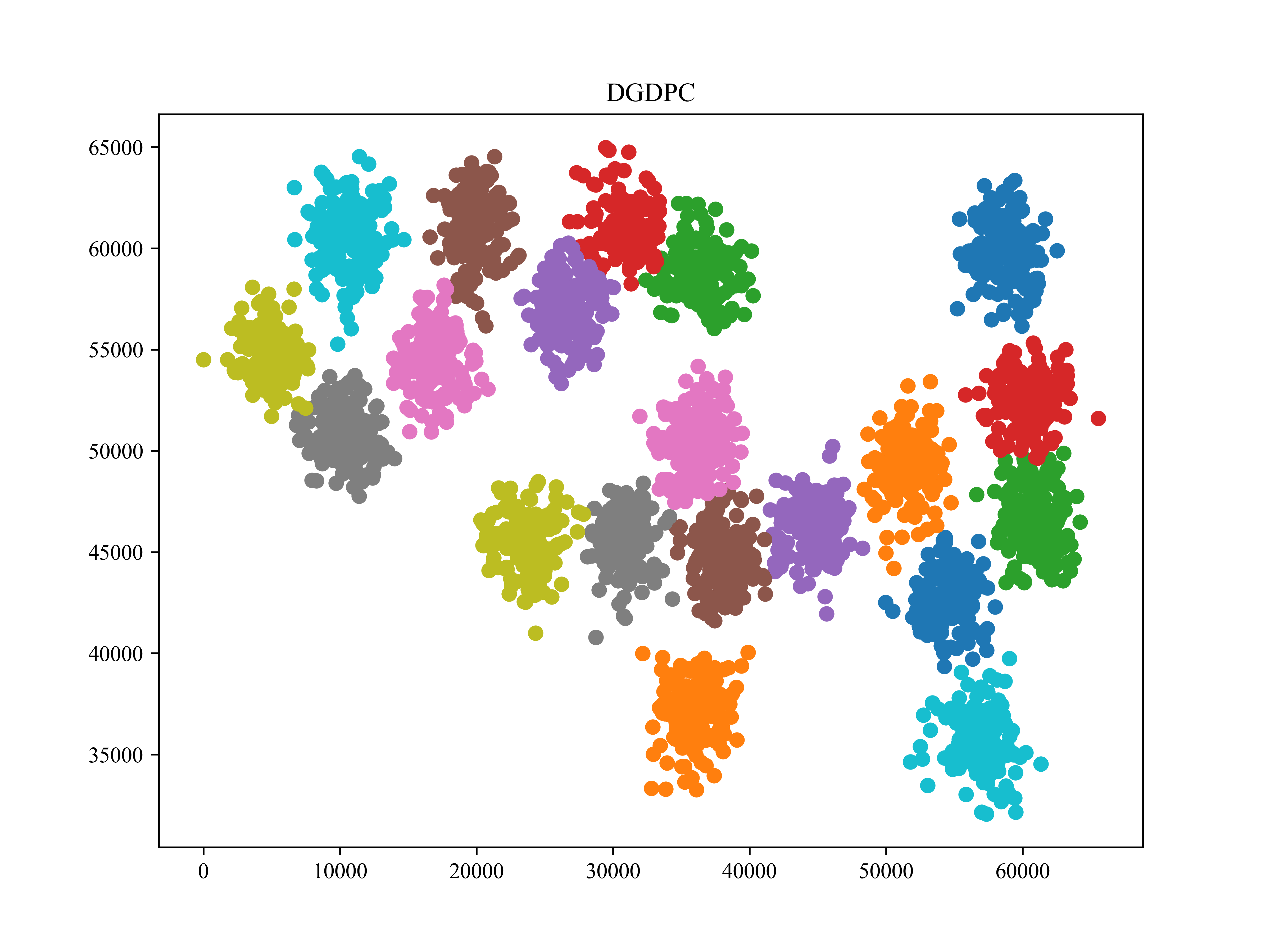}}\\
\subfloat{\includegraphics[width = 0.2\textwidth]{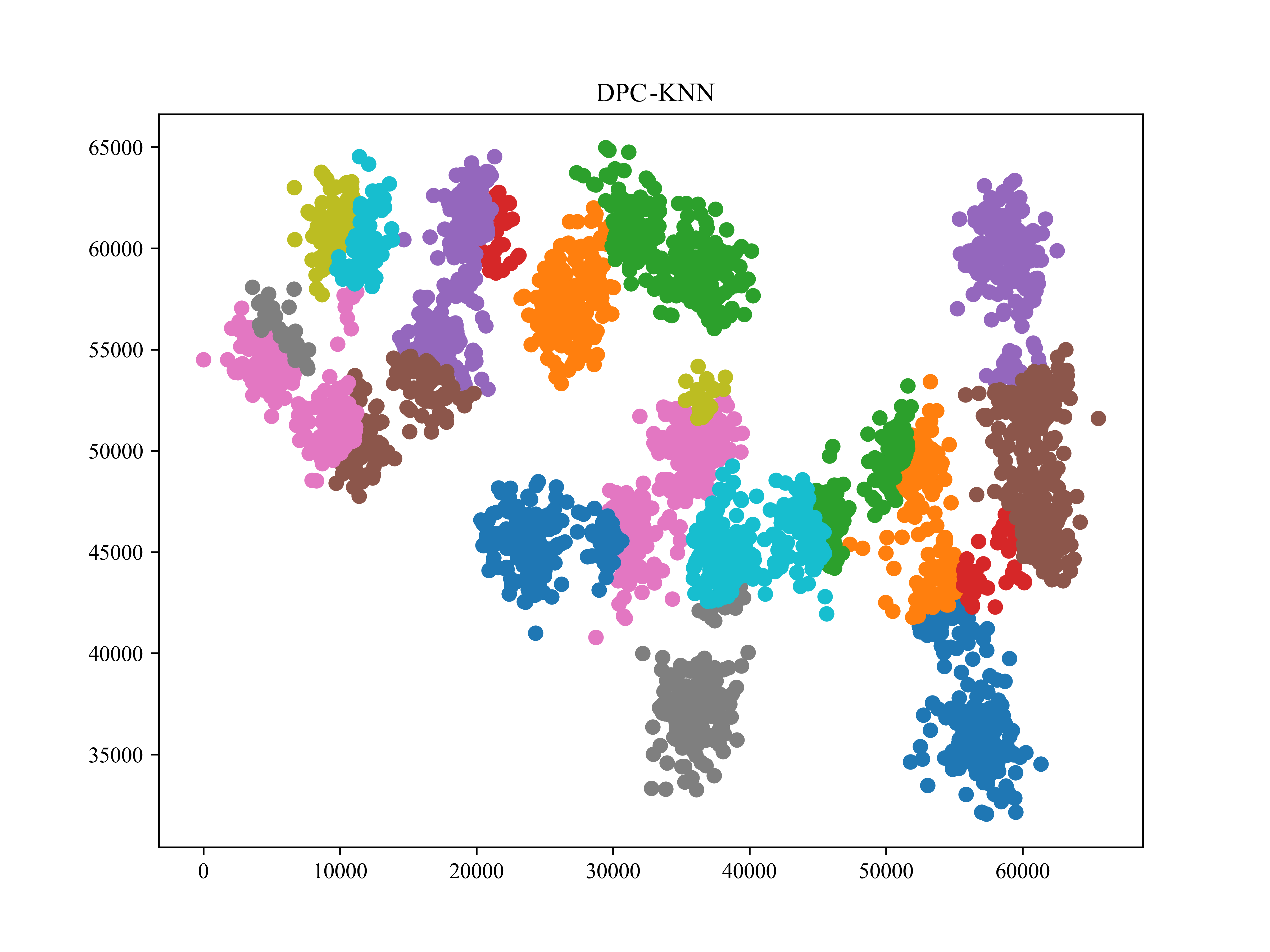}\centering}
\subfloat{\includegraphics[width = 0.2\textwidth]{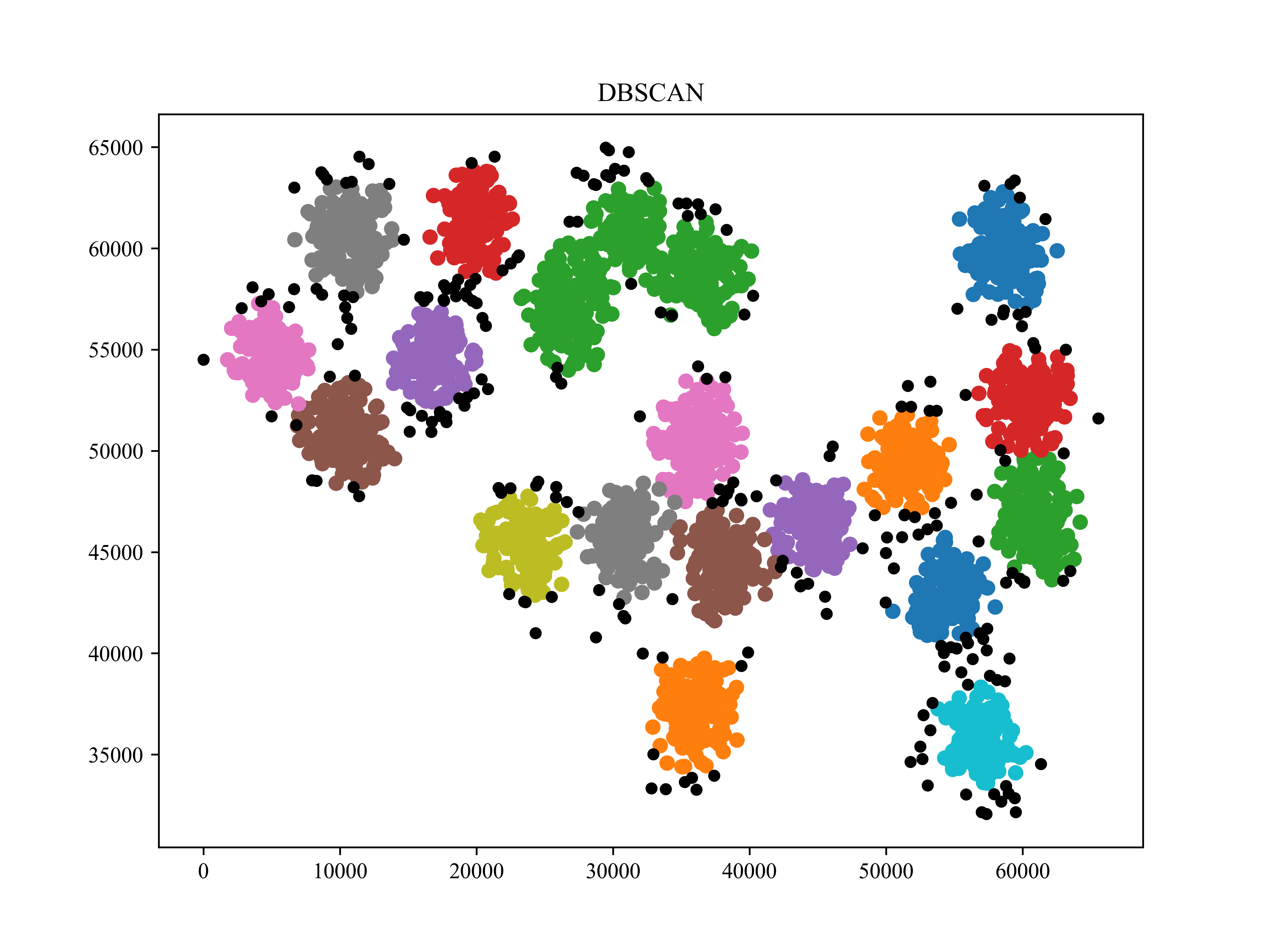}\centering}
\subfloat{\includegraphics[width = 0.2\textwidth]{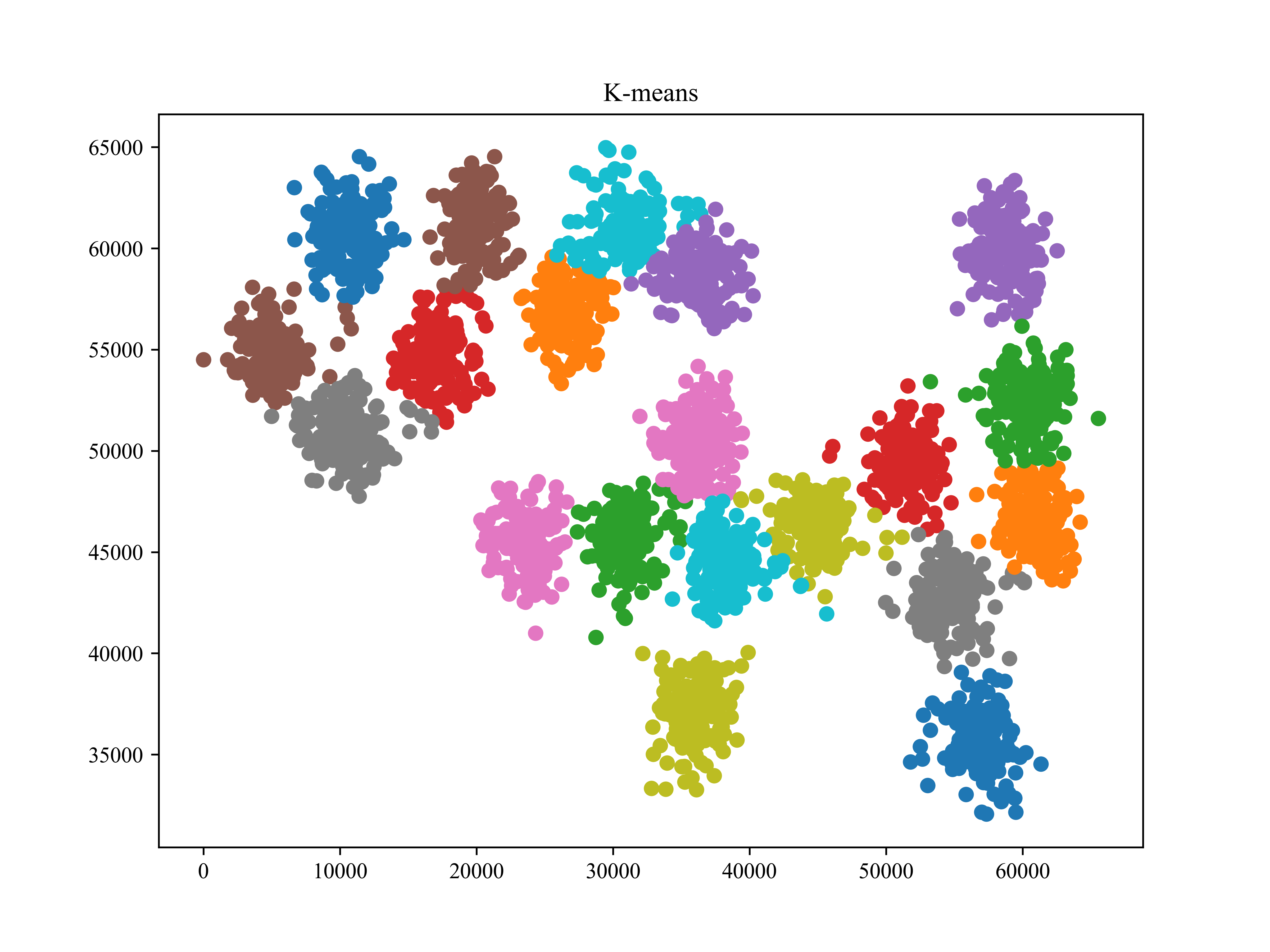}\centering}	
\caption{\textbf {Clustering results of the seven algorithms on A1 dataset. }}
\label{fig:labe8}
\end{figure*}

\begin{figure*}[ht]
\centering
	\subfloat{\includegraphics[width = 0.2\textwidth]{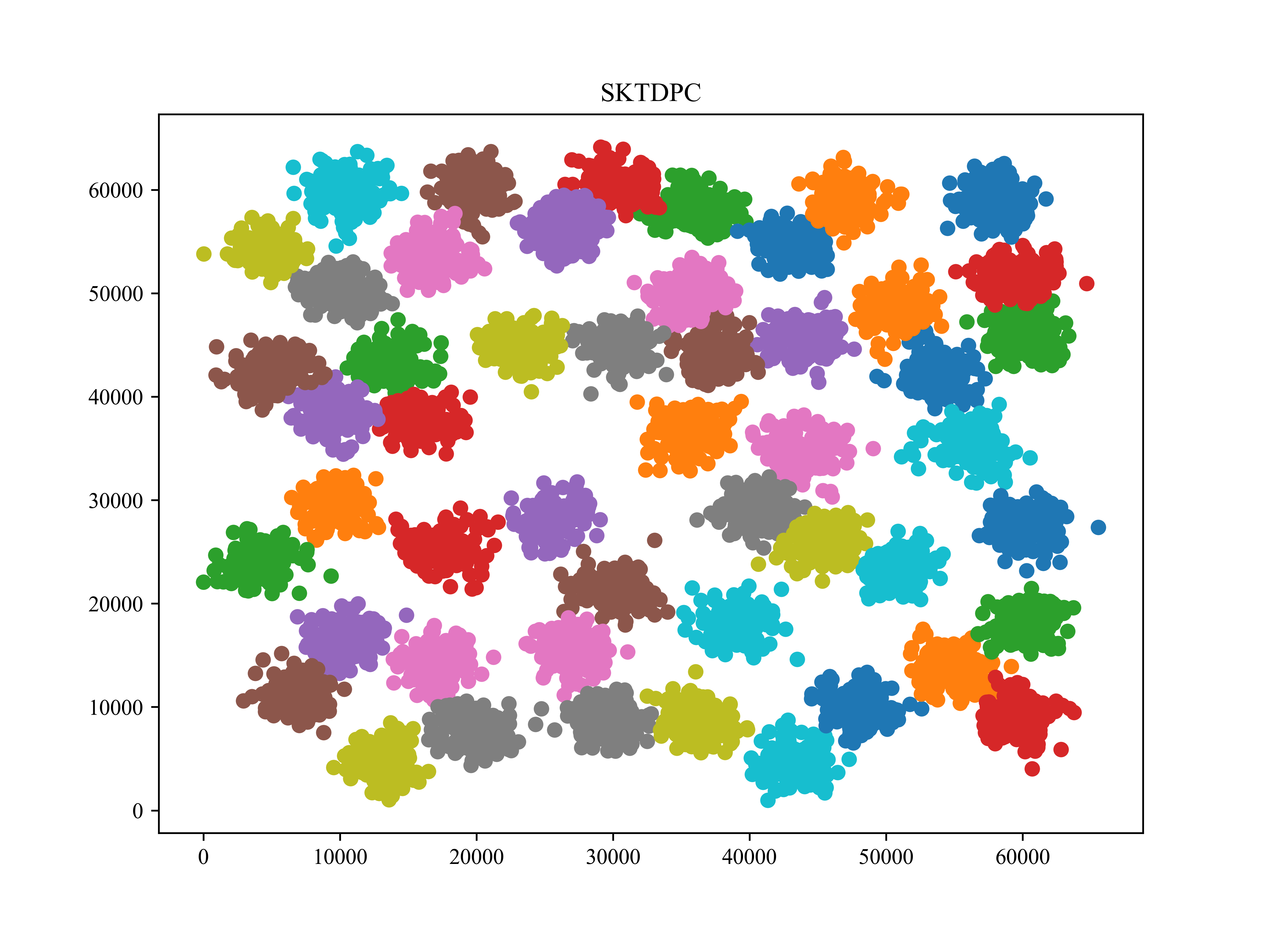}}
	\subfloat{\includegraphics[width = 0.2\textwidth]{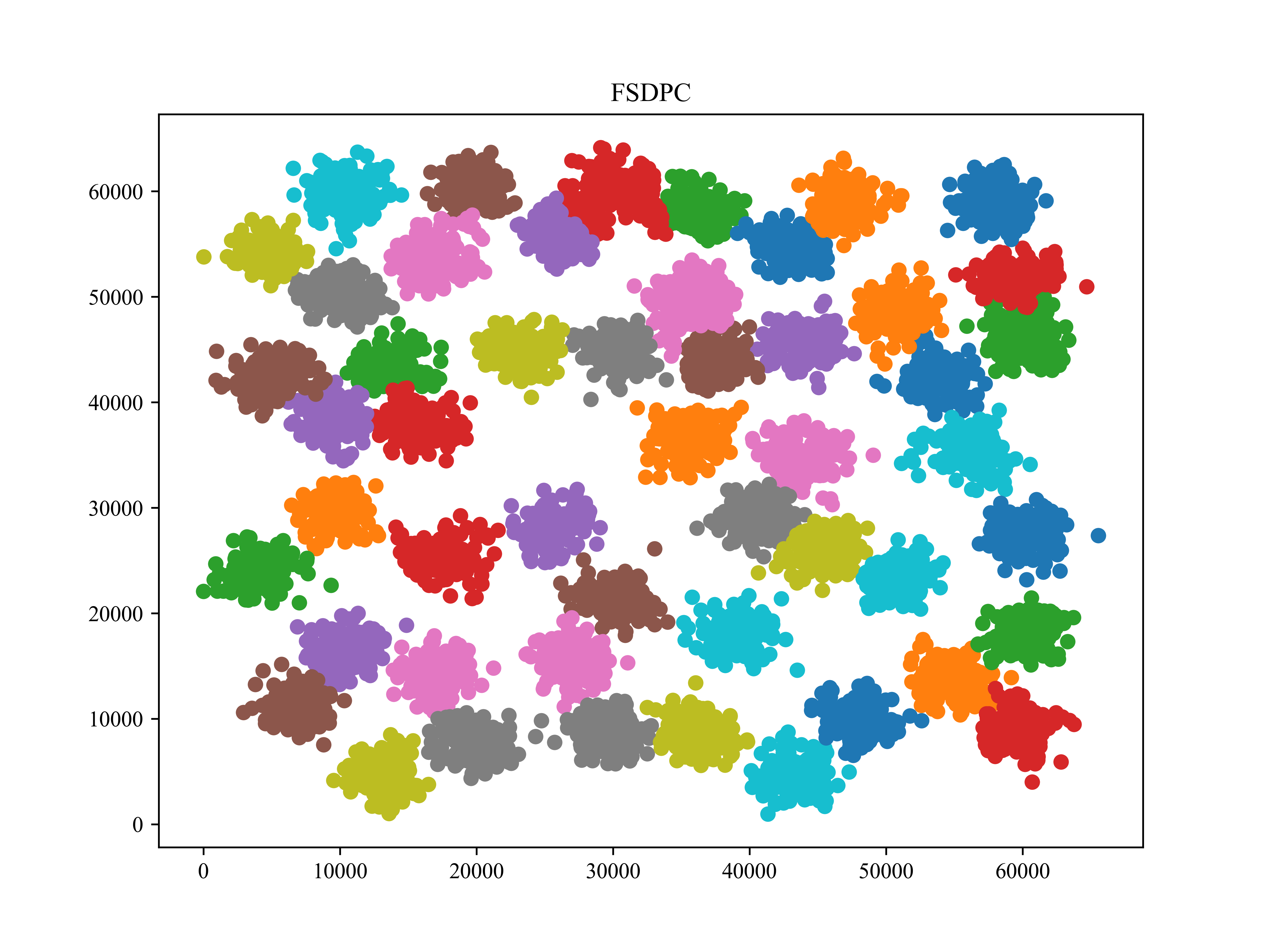}}
\subfloat{\includegraphics[width = 0.2\textwidth]{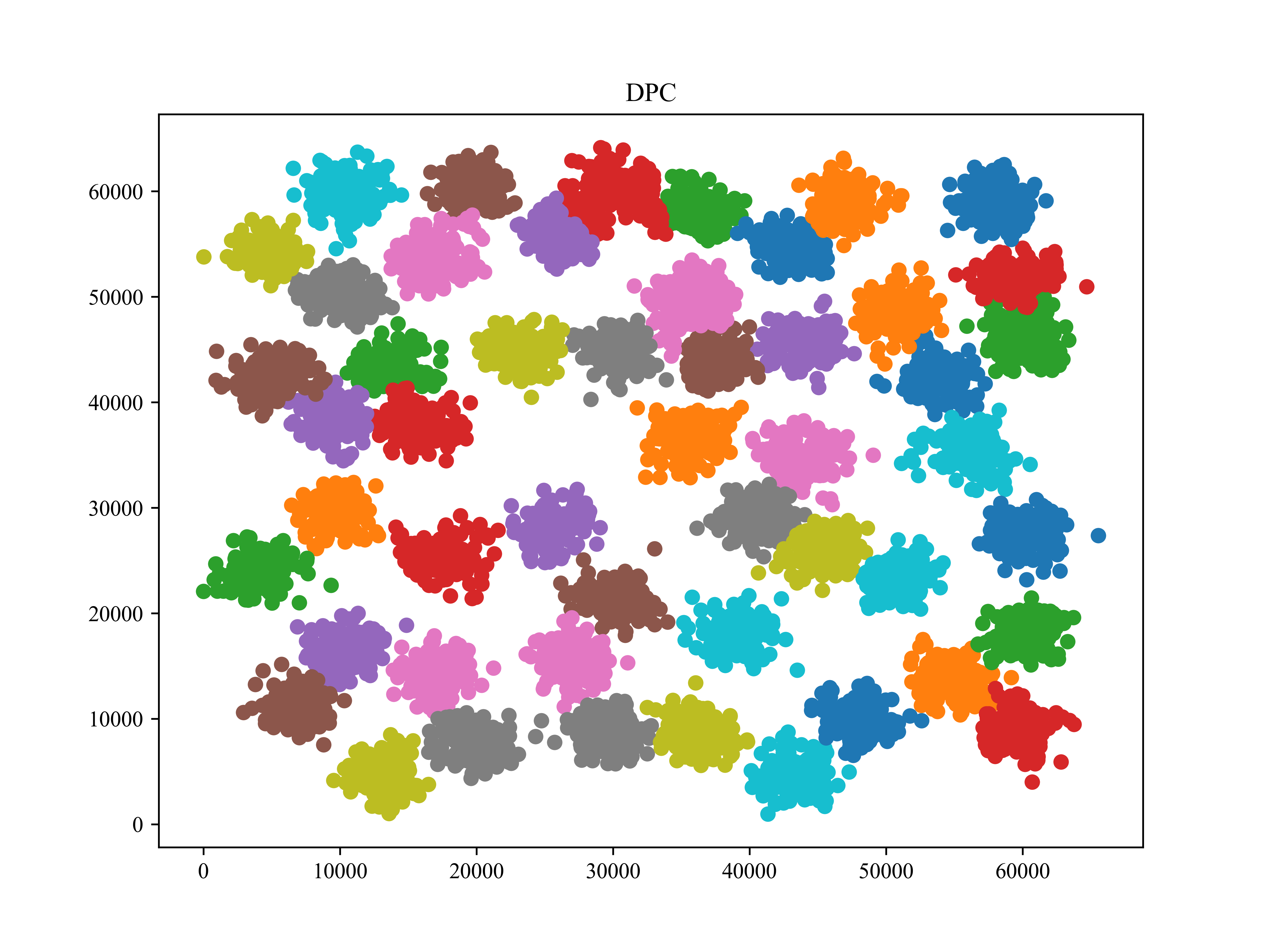}}
\subfloat{\includegraphics[width = 0.2\textwidth]{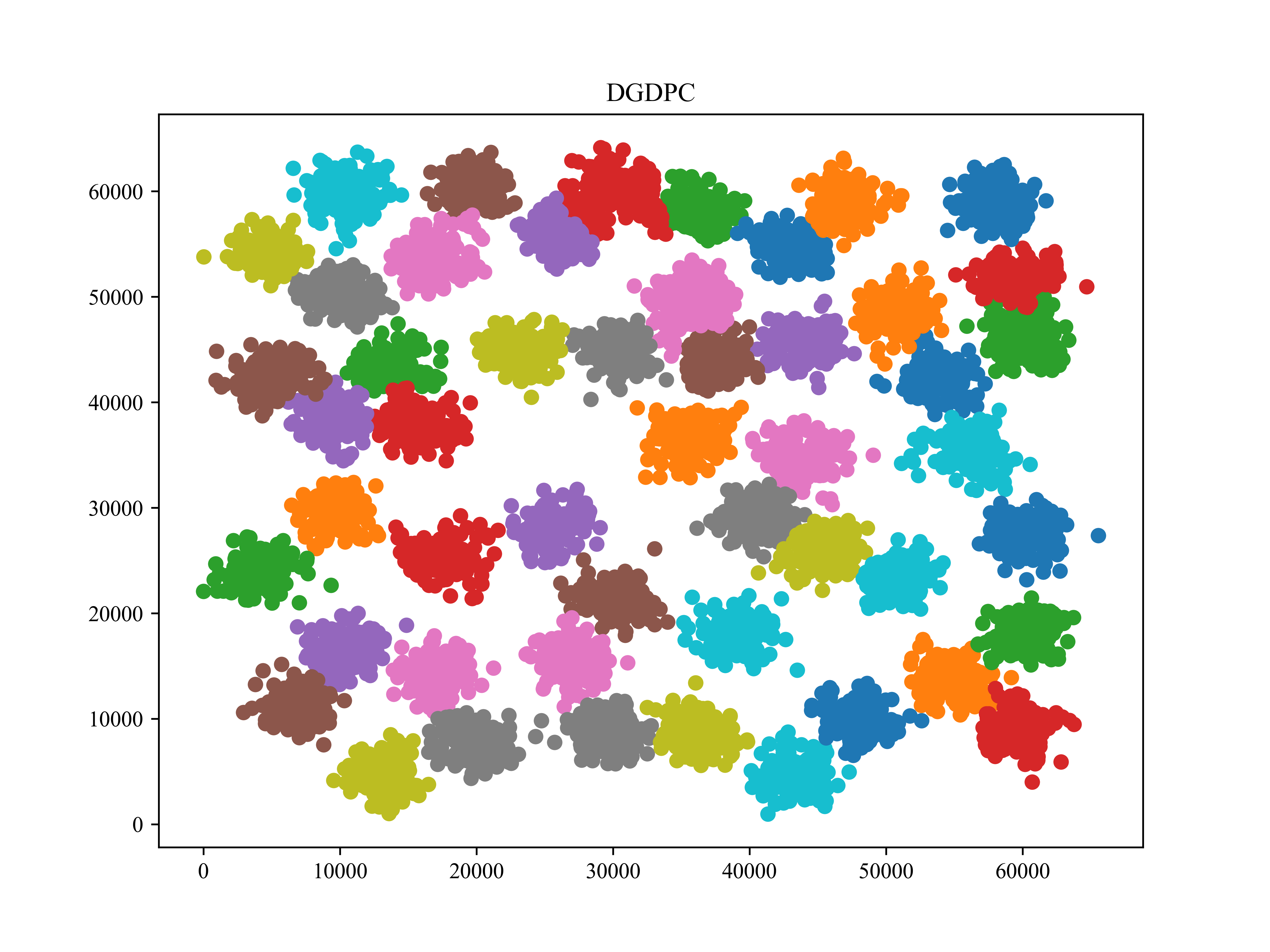}}\\
\subfloat{\includegraphics[width = 0.2\textwidth]{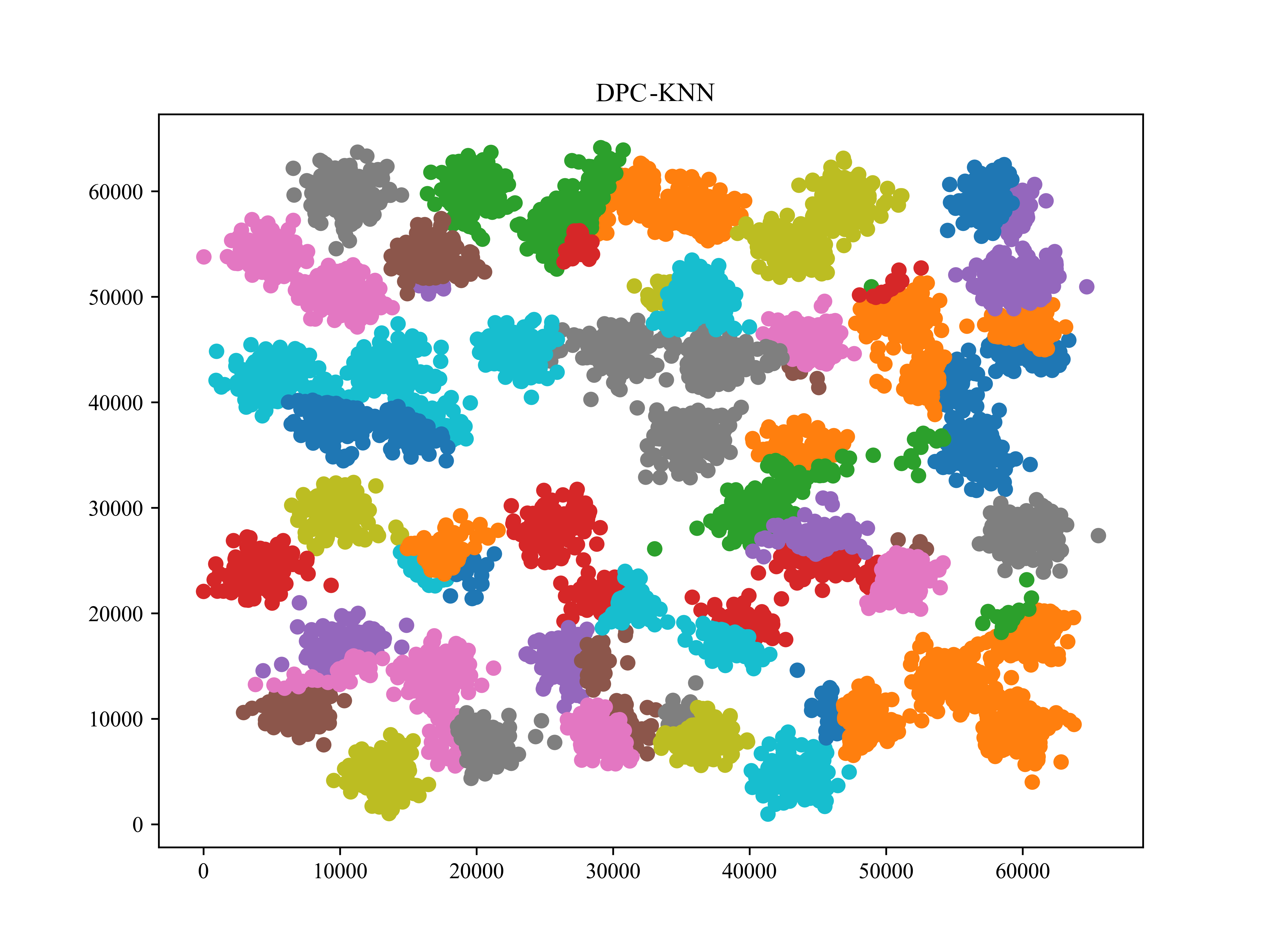}\centering}
\subfloat{\includegraphics[width = 0.2\textwidth]{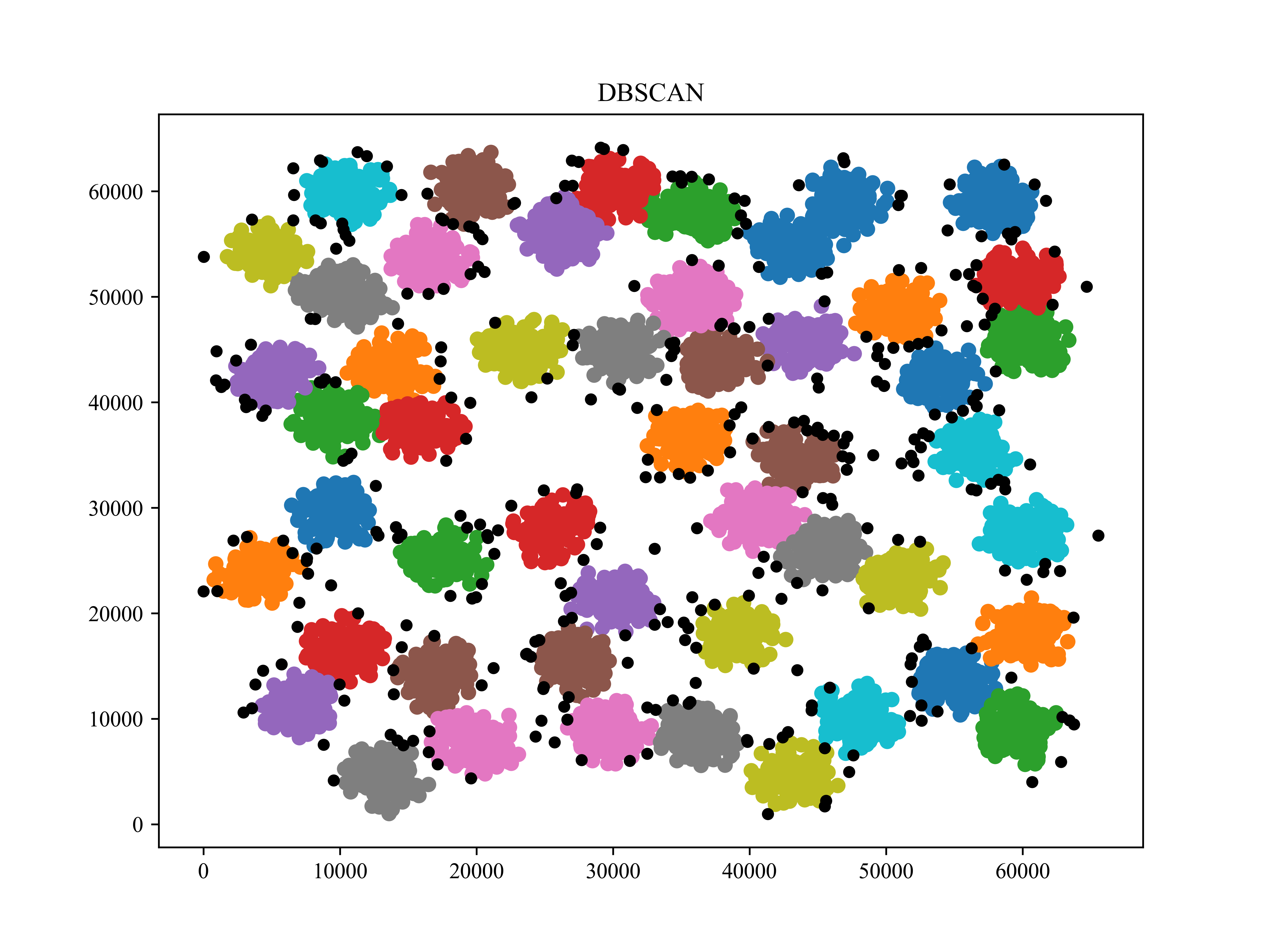}\centering}
\subfloat{\includegraphics[width = 0.2\textwidth]{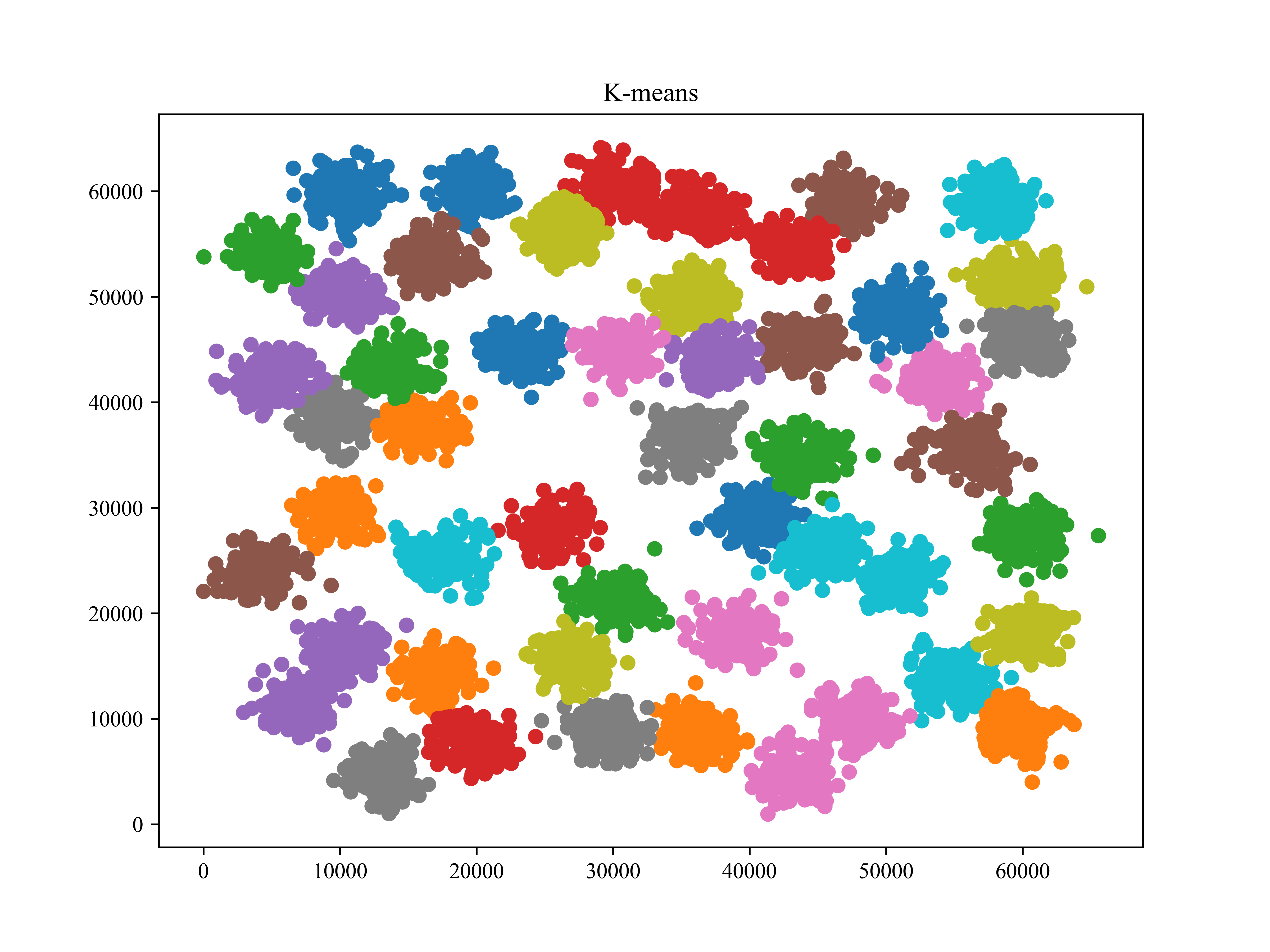}\centering}	
\caption{\textbf {Clustering results of the seven algorithms on A3 dataset.} }
\label{fig:labe9}
\end{figure*}

\begin{table*}
\centering
\caption{\textbf{CLUSTERING EFFICIENCY OF ALGORITHMS ON SYNTHETIC DATASETS.}}
\begin{tabular}{lccccccc}
\hline\hline
\multirow{2}{*}{Data} &\multicolumn{7}{c}{Running time (s)} \\
\cline{2-8} & SKTDPC&	FSDPC&	DPC&	DGDPC&	DPC-KNN&	DBSCAN&	K-means\\
\hline
Flame&	0.1&	0.24&	0.31&	0.34&	0.28&	0.17&	0.15\\
Spiral&	0.14&	0.29&	0.33&	0.39&	0.37&	0.14&	0.15\\
Aggregation&	0.53&	0.73&	0.82&	0.88&	0.74&	0.19&0.32\\
R15&	0.35&	0.54&	0.61&	0.86&	0.79&	0.70&	0.27\\
S1&	8.16&	14.38&	18.70&	22.13&	20.29&	1.10&	0.46\\
S3&	10.58&	16.78&	19.61&	23.27&	21.07&	8.63&	0.42\\
A1&	4.04&	6.91&	7.81&	8.02&	7.75&	1.41&	0.43\\
A3&	12.86&	30.41&	40.06&	49.63&	45.93&	2.72&	1.62\\
\hline\hline
\end{tabular}
\label{tab6}
\end{table*}

Among the other six clustering algorithms compared with SKTDPC, the code for original DPC and DPC-KNN are obtained by retranslating the MATLAB source code, provided by the authors of original DPC and DPC-KNN, into Python language form. The FSDPC and DGDPC codes are written in accordance with the original reference. The algorithms for DBSCAN and K-means are programmed by the sklearn.cluster library in Python. In order to make a fair comparison, for each algorithm experiments are conducted with its optimal hyper-parameters. At the same time, the values of all the indicators below are average by the 20 times independently repeated experiments for each algorithm, to avoid the occurrence of contingency. In addition, all algorithms are implemented by Python 3.8.0. The experiments are carried out in a computer environment with a core i7 2.3 GHz processor, Windows 10 operating system and 16GB RAM. The relevant code for this article is published in https://github.com/Nutshe/code.

\begin{table*}
\centering
\caption{\textbf{ACC OF ALGORITHMS ON DIFFERENT UCI DATASETS.}}
\begin{tabular}{lcccccccccccccc}
\hline\hline
\multirow{2}{*}{Data} &\multicolumn{2}{c}{SKTDPC} &\multicolumn{2}{c}{FSDPC}&\multicolumn{2}{c}{DPC}&\multicolumn{2}{c}{DGDPC}&\multicolumn{2}{c}{DPC-KNN}&\multicolumn{2}{c}{DBSCAN}&\multicolumn{2}{c}{K-means}\\
\cline{2-15} & Acc & Par & Acc & Par & Acc & Par & Acc & Par & Acc & Par & Acc& Par & Acc & Par    \\
\hline
Seeds	&\bf 0.914&	3&	0.895&	1&	0.904&	2&	0.909	&1/1	&0.895&	6&	0.348&	0.09/3&	0.890	&3\\
Iris&\bf	0.960	&2&	0.957&	4	&\bf 0.960&	4	&0.892&	1/3&\bf	0.960	&7	&0.740&	0.12/5&	0.887	&3\\
Banknote authentication	&0.683&	2	&0.680&	2	&0.687&	2	&0.685&	1/2	&0.517&	5	&\bf 0.742&	0.09/2&	0.576	&2\\
Wine&	0.893&	6&	0.853	&2	&0.854&	2	&0.762&	1/1&	0.859	&6&	0.815&	0.5/21&\bf	0.949	&3\\
Ecoli&\bf	0.807	&2&	0.542&	2&	0.542&	2&	0.586&	1/2&	0.631&	3&	0.622&	0.2/10&	0.595&	8\\
Parking Birmingham&\bf	0.575&	4&	0.541	&1&	0.543	&1&	0.557&	1/1&	0.556&	4&	0.432	&0.08/67&	0.336&	3\\
\hline\hline
\end{tabular}
\label{tab7}
\end{table*}

\begin{table*}
\centering
\caption{\textbf{AMI, ARI OF ALGORITHMS ON DIFFERENT UCI DATASETS.}}
\label{table}
\begin{tabular}{lcccccccccccccc}
\hline\hline
\multirow{2}{*}{Data} &\multicolumn{2}{c}{SKTDPC} &\multicolumn{2}{c}{FSDPC}&\multicolumn{2}{c}{DPC}&\multicolumn{2}{c}{DGDPC}&\multicolumn{2}{c}{DPC-KNN}&\multicolumn{2}{c}{DBSCAN}&\multicolumn{2}{c}{K-means}\\
\cline{2-15} & AMI & ARI & AMI & ARI & AMI & ARI & AMI & ARI & AMI & ARI & AMI& ARI & AMI & ARI    \\
\hline
Seeds	&\bf 0.732&\bf 	0.766&	0.674&	0.717&	0.701&	0.741&	0.726&	0.753&	0.674&	0.717&	0.057&	0.001&	0.671&	0.705\\
Iris&\bf	0.861&\bf	0.886&	0.857&	0.879	&\bf 0.861&	0.885&	0.787&	0.781&	0.858&	0.885&	0.634&	0.612&	0.733&	0.716\\
Banknote authentication&	0.222&	0.130&	0.168&	0.132&	0.173&	0.133&	0.185	&0.135&	0.068	&0.010&	\bf 0.534	&\bf 0.589&	0.017&	0.022\\
Wine&	0.723&	0.699&	0.679&	0.613&	0.678&	0.613&	0.621&	0.458&	0.681&	0.615&	0.586&	0.529	&\bf 0.840&\bf	0.854\\
Ecoli&\bf	0.665&\bf	0.740&	0.554&	0.416&	0.555	&0.416&	0.597&	0.463&	0.606	&0.499&	0.487&	0.494	&0.601&	0.425\\
Parking Birmingham&\bf	0.078	&\bf 0.075&	0.052&	0.046	&0.052&	0.049	&0.062&	0.059	&0.061&	0.058	&0.017&	0.042&	0.010&	0.007\\
\hline\hline
\end{tabular}
\label{tab8}
\end{table*}

\begin{table*}
\centering
\caption{\textbf{NMI, FMI OF ALGORITHMS ON DIFFERENT UCI DATASETS.}}
\label{table}
\begin{tabular}{lcccccccccccccc}
\hline\hline
\multirow{2}{*}{Data} &\multicolumn{2}{c}{SKTDPC} &\multicolumn{2}{c}{FSDPC}&\multicolumn{2}{c}{DPC}&\multicolumn{2}{c}{DGDPC}&\multicolumn{2}{c}{DPC-KNN}&\multicolumn{2}{c}{DBSCAN}&\multicolumn{2}{c}{K-means}\\
\cline{2-15} & NMI & FMI & NMI & FMI & NMI & FMI & NMI & FMI & NMI & FMI & NMI& FMI & NMI & FMI    \\
\hline
Seeds&\bf	0.734&\bf	0.844&	0.675&	0.785	&0.704	&0.826&	0.728&	0.831&	0.675&	0.785	&0.090	&0.542&	0.674	&0.803\\
Iris&\bf	0.862&\bf	0.923&	0.858&	0.911	&\bf 0.862&	\bf 0.923&	0.789&	0.839	&0.861&\bf	0.923&	0.640&	0.729&	0.742&	0.811\\
Banknote authentication&	0.222&	0.673&	0.225&	0.655&	0.230&	0.678&	0.228&	0.676	&0.108&	0.589&	\bf 0.537&\bf	0.770&	0.017&	0.514\\
Wine&	0.726&	0.801&	0.680&	0.745&	0.682&	0.746&	0.619&	0.683&	0.685&	0.749&	0.590&	0.712&	\bf 0.842&\bf	0.902\\
Ecoli&\bf	0.675&\bf	0.815&	0.574&	0.559&	0.573&	0.559&	0.601&	0.605&	0.620&	0.617&	0.496&	0.676&	0.618	&0.559\\
Parking Birmingham&\bf	0.078&\bf	0.548&	0.052&	0.443&	0.052&	0.443&	0.063&	0.526&	0.061&	0.477&	0.017&	0.382&	0.010&	0.376\\
\hline\hline
\end{tabular}
\label{tab9}
\end{table*}

\subsection{EXPERIMENTS ON SYNTHETIC DATASETS}
Among the eight synthetic datasets with different distributions, the four classic datasets Flame, Spiral, Aggregation and S3 were used in the reference on original DPC \cite{b14}. The 15 class clusters of dataset R15 are distributed in the ring and have similar Gaussian distribution. The datasets S1, A1 and A3 are the three commonly used clustering datasets with different characteristics of overlapping, complexity and number of class clusters. Applying to the eight two-dimensional synthetic datasets, the clustering results for the seven algorithms are shown in Figs. \ref{fig:labe2}-\ref{fig:labe9}, visually. Meanwhile, the five clustering evaluation indexes are shown in Tables \ref{tab3}-\ref{tab5}, in which DPC refers to the original DPC \cite{b14}. In addition, the values of optimal hyper-parameter, abbreviated by Par, suitable for each algorithm are also given in Table \ref{tab3}.

For the Acc, AMI, ARI, NMI and FMI index values of the seven algorithms shown in Tables \ref{tab3}-\ref{tab5} and the clustering effect shown in Figs. \ref{fig:labe2}-\ref{fig:labe9}, we can see that the SKTDPC algorithm shows the best performance on all of the eight synthetic datasets. This is because SKTDPC introduces the idea of k nearest neighbors to define the local density and obtains the correct cluster centers adaptively by second-order difference method. Through these two methods, SKTDPC can capture more comprehensive information for data distribution and avoid the wrong selection of cluster centers, which also plays an important role in improving the cluster effect. By further comparing SKTDPC with other algorithms, it is demonstrated that there are relative weaknesses with different degree for these six algorithms. Firstly, FSDPC, the original DPC and DGDPC are slightly inferior to SKTDPC, even though they perform well on most datasets. This is because these two algorithms only consider the global structure of data, resulting in partial information loss. By contrast, the local density of SKTDPC is calculated based on the distances from k nearest neighbors, which can deal with local data information well. In addition, FSDPC and the original DPC need to determine the clustering center manually. Thus the possible wrong selection of cluster centers would also lead to poor clustering effect. Secondly, for DPC-KNN and DBSCAN, both of them show obviously poor clustering effect on S3, A1 and A3 datasets, as shown in Figs. \ref{fig:labe7}-\ref{fig:labe9}. This implies that the processing ability of the two algorithms is obviously weak for datasets with a high degree of overlap relatively. Moreover, the accuracy of DBSCAN is slightly lower than that of other algorithms on Flame, Aggregation, R15 and S1 datasets. This is caused by its recognition of noise points, as shown in related figures. Finally, the clustering effect of K-means algorithm is very poor on Flame, Spiral and Aggregation datasets, as shown in Figs. \ref{fig:labe2}-\ref{fig:labe4}. This is attributed to the defects of K-means itself, which cannot capture the structural characteristics of non-convex dataset. In contrast, the other six density-based algorithms show good performance on these datasets.

\begin{table*}
\centering
\caption{\textbf{CLUSTERING EFFICIENCY OF ALGORITHMS ON DIFFERENT UCI DATASETS.}}
\begin{tabular}{lccccccc}
\hline\hline
\multirow{2}{*}{Data} &\multicolumn{7}{c}{Running time (s)} \\
\cline{2-8} & SKTDPC&	FSDPC&	DPC&	DGDPC&	DPC-KNN&	DBSCAN&	K-means\\
\hline
Seeds&	0.16&	0.17&	0.19&	0.27&	0.39&	0.26&	0.11\\
Iris&	0.09	&0.12&	0.15&	0.20&	0.27&	0.09&	0.08\\
Banknote authentication&	1.63&	2.97&	3.85&	6.12&	11.58&	0.10&	0.19\\
Wine&	0.17&	0.19&	0.22&	0.31&	0.40&	0.19&	0.08\\
Ecoli&	0.31&	0.43&	0.48&	0.61&	0.76&	0.14&	0.14\\
Parking Birmingham&	685.91&	1316.23&	–&	–	&–	&9.22	&0.43\\
\hline\hline
\end{tabular}
\label{tab10}
\end{table*}

The running time of SKTDPC algorithm and the other six algorithms is shown in Table \ref{tab6} to evaluate the algorithmic clustering efficiency. It can be found that among the five algorithms belonging to the DPC series, including SKTDPC, FSDPC, the original DPC, DGDPC and DPC-KNN, the clustering efficiency of SKTDPC algorithm newly proposed is much higher than that for the other four algorithms. Especially when the dataset size is large enough, the acceleration effect is relatively more obvious. This is precisely because SKTDPC accelerates the calculation of local density and relative-separation by K-d tree and the sparse search strategy, and further requires less distance calculation and storage space compared with the other four algorithms. It can also be found that the running speed of K-means and DBSCAN is relatively faster compared with the DPC series. As discussed above, however, K-means is only applicable to globular clusters. For non-convex clusters, the clustering effect is very poor, even though the efficiency is very high. As for DBSCAN, it is sensitive to the setting of two hyper-parameters and is difficult to use when the data density is not evenly distributed. In addition, as shown above, the processing ability of this algorithm is weak obviously for datasets with a high degree of overlap, such as S3, A1 and A3.

\subsection{EXPERIMENTS ON REAL-WORLD DATASETS}
In order to further verify the clustering effect and efficiency of SKTDPC algorithm, a comparative analysis is made on six UCI real datasets (Seeds, Iris, Banknote, Wine, Ecoli, Parking Birmingham) with different dataset size, number of class clusters and data dimension. In contrast with the eight synthetic datasets, which are two-dimensional, the UCI real datasets are relatively high-dimensional and complex. The performance of SKTDPC and other six algorithms is compared. The index values and parameter values of all algorithms are shown in Tables \ref{tab7}-\ref{tab10}, in which “-” indicates that the value for running time at this position is very large or even cannot be obtained.

From the Acc, AMI, ARI, NMI and FMI values shown in Tables \ref{tab7}-\ref{tab9} it can be found that: (1) SKTDPC algorithm achieves the best clustering results compared with the DPC series and K-means algorithm in all real datasets, and the clustering accuracy is improved to a certain degree. For DBSCAN algorithm, it gives the best clustering accuracy on Banknote authentication, better than SKTDPC slightly. However, DBSCAN shows much lower accuracy than that given by SKTDPC on other types of datasets. (2) The overall clustering effect of DPC series algorithm is better than K-means and DBSCAN algorithm, indicating that the general applicability of DPC series algorithm is stronger relatively. (3) FSDPC and DGDPC algorithms can maintain similar clustering accuracy as the original DPC, but there is still distinction in some extent for these two algorithms compared with SKTDPC algorithm. This is because FSDPC, DPC and DGDPC ignore local data information. In addition, SKTDPC can obtain the location of mutation-point adaptively in the clustering process to determine the cluster centers without interrupting the algorithm. Furthermore, due to the complexity and sparsity of the Banknote and Parking datasets, the clustering results of the seven algorithms is not satisfactory. For this kind of dataset, it is necessary to further study its inherent data structure characteristics, to find a more effective clustering method.

By analyzing Table \ref{tab10}, it can be found that the proposed SKTDPC algorithm obviously improves the clustering efficiency. In small datasets, the clustering efficiencies of the seven algorithms are roughly the same, but the distinction becomes more obvious with the increase of data volume. For the dataset Parking Birmingham with the largest data volume, the clustering efficiency of SKTDPC is 1.92 times and more times higher than that of FSDPC, DPC, DGDPC  and DPC-KNN, respectively. At this time, DPC, DGDPC and DPC-KNN are on the verge of collapse and almost lost their execution ability because they require huge amount of computation and storage space. In addition, the running time of these two algorithms is seriously time-consuming, which is already meaningless. The high efficiency for SKTDPC is mainly attributed to the dual acceleration strategies to deal with large datasets, as analyzed in Section III.F. One is the acceleration for calculation of the local density by K-d tree. Thus a sparse distance matrix $\widetilde D$ is calculated instead of a full-rank matrix $D$ to find the $k$ nearest neighbors. Another is the acceleration for calculation of the relative-separation by the sparse search strategy with intersection between $N\!N_k(x_i)$ and $B(x_i)$. Therefore, the dual accelerations strategies play an important role in reducing the algorithmic complexity, which makes SKTDPC algorithm show more prominent on large datasets. It should also be noted that although the running speed of DBSCAN and K-means algorithms is fast, due to their own limitations, the universality of the algorithms are too weak to produce an ideal clustering results.

\begin{figure*}[ht]
\centering
	\subfloat[Acc of SKTDPC on synthetic datasets.]{\includegraphics[width = 0.5\textwidth]{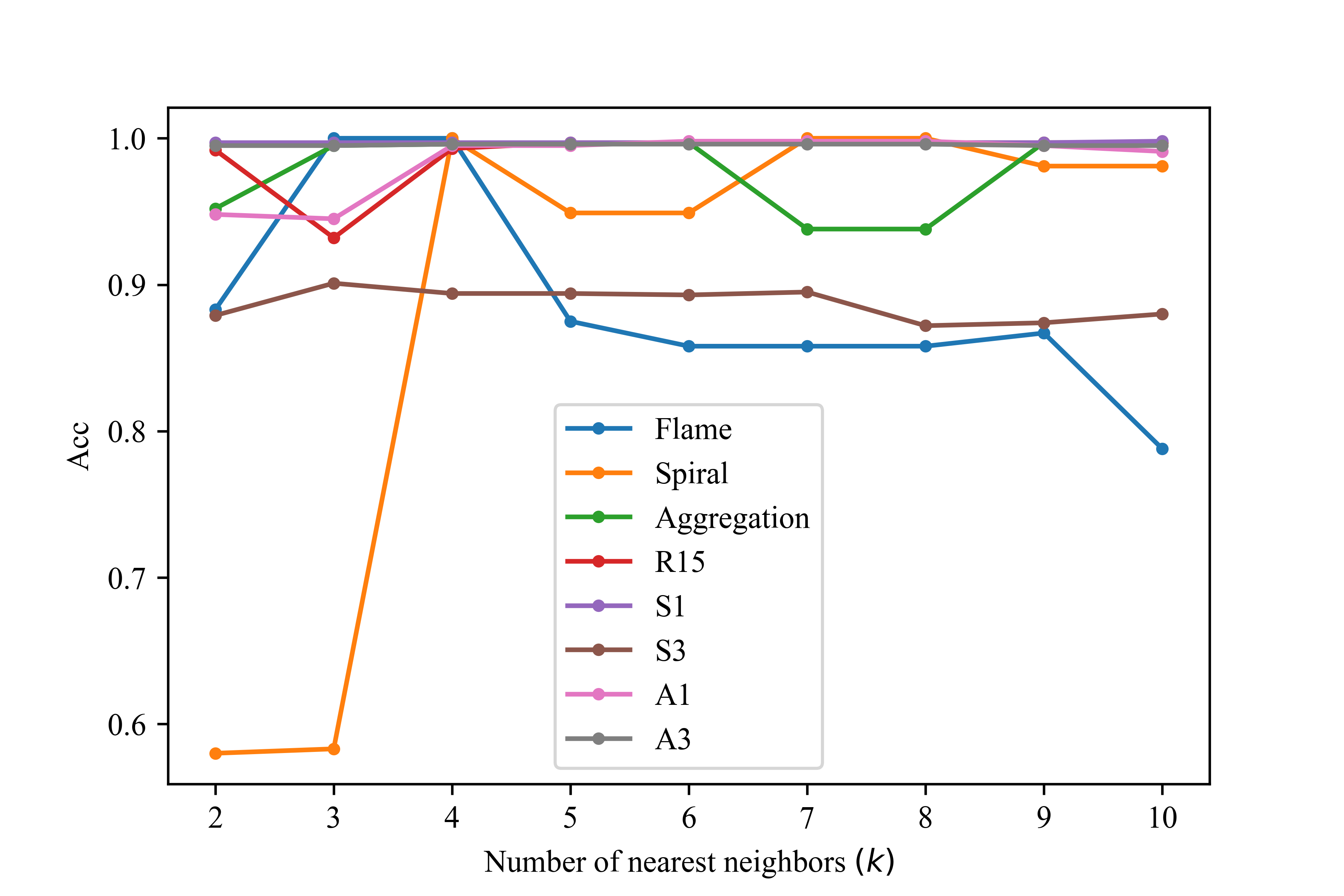}\centering}
	\hfill
	\subfloat[Acc of SKTDPC on UCI datasets.]{\includegraphics[width = 0.5\textwidth]{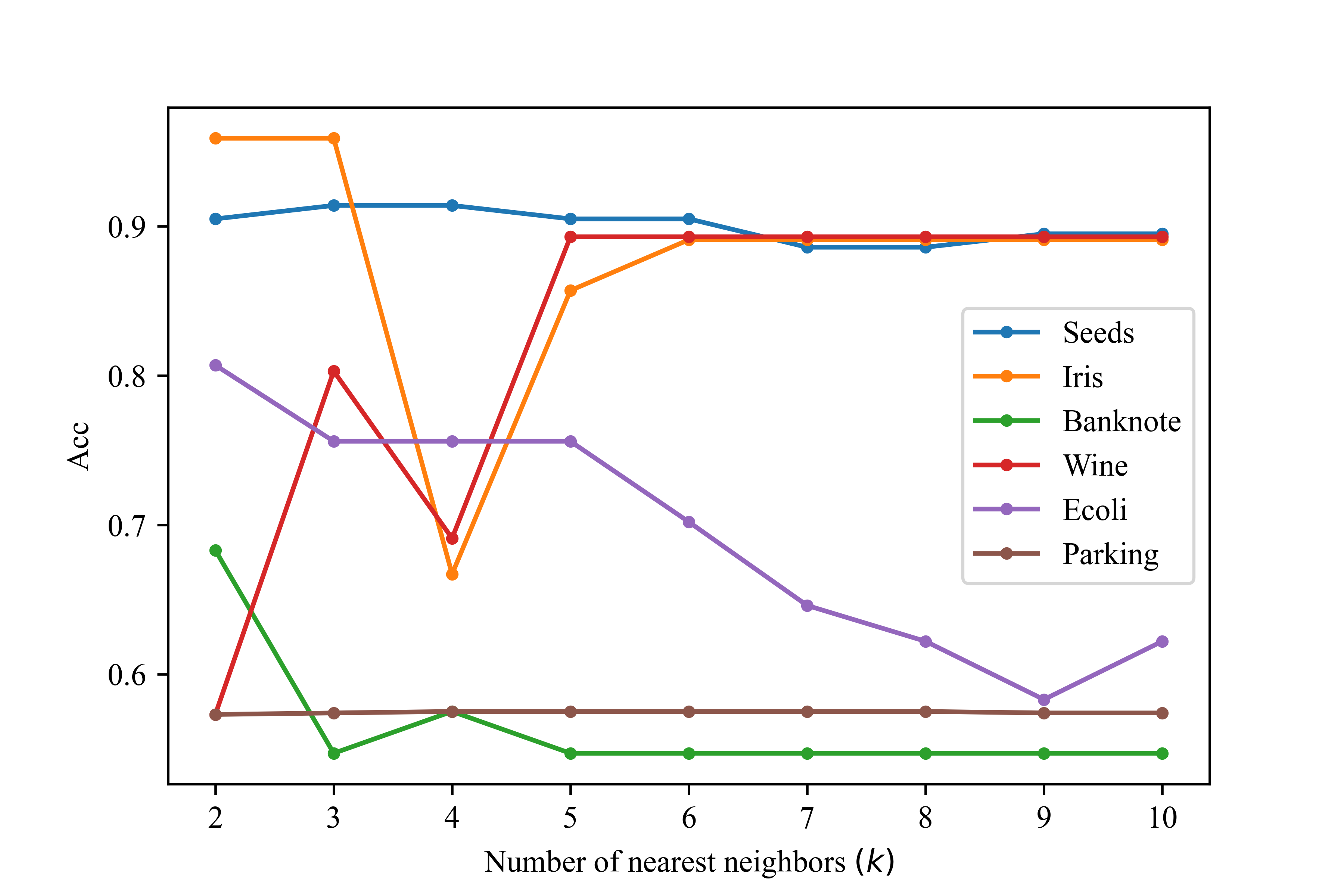}}
	\hfill
\caption{\textbf {Acc of SKTDPC under different parameters $k$. }}
\label{fig:labe10}
\end{figure*}

\subsection{ROBUSTNESS TO NUMBER OF NEAREST NEIGHBORS K}
In this section, this paper will analyze and discuss the influence of the proposed algorithm SKTDPC on the Acc under different parameter $k$ settings. The number of nearest neighbors $k$ is the only hyper-parameter of SKTDPC, so it is crucial to analyze the change of clustering effect under different settings. The adjustment of parameters is to accommodate datasets with different distribution characteristics. At the same time, within the scope of the regulation of as small as possible to avoid algorithm into local optimum and can achieve ideal clustering result is what we are after. If the parameters of the algorithm need to be optimized in a wide range, it will not only affect the search efficiency of the optimal parameters, but also make the robustness and stability of the algorithm worse and it is difficult to achieve the ideal clustering effect. Therefore, we focus on analyzing the change of Acc within a reasonable range of $k = [2, 10]$ for the SKTDPC algorithm, as shown in Fig. \ref{fig:labe10}. 

\begin{figure*}[ht]
\centering
	\subfloat{\includegraphics[width = 0.35\textwidth]{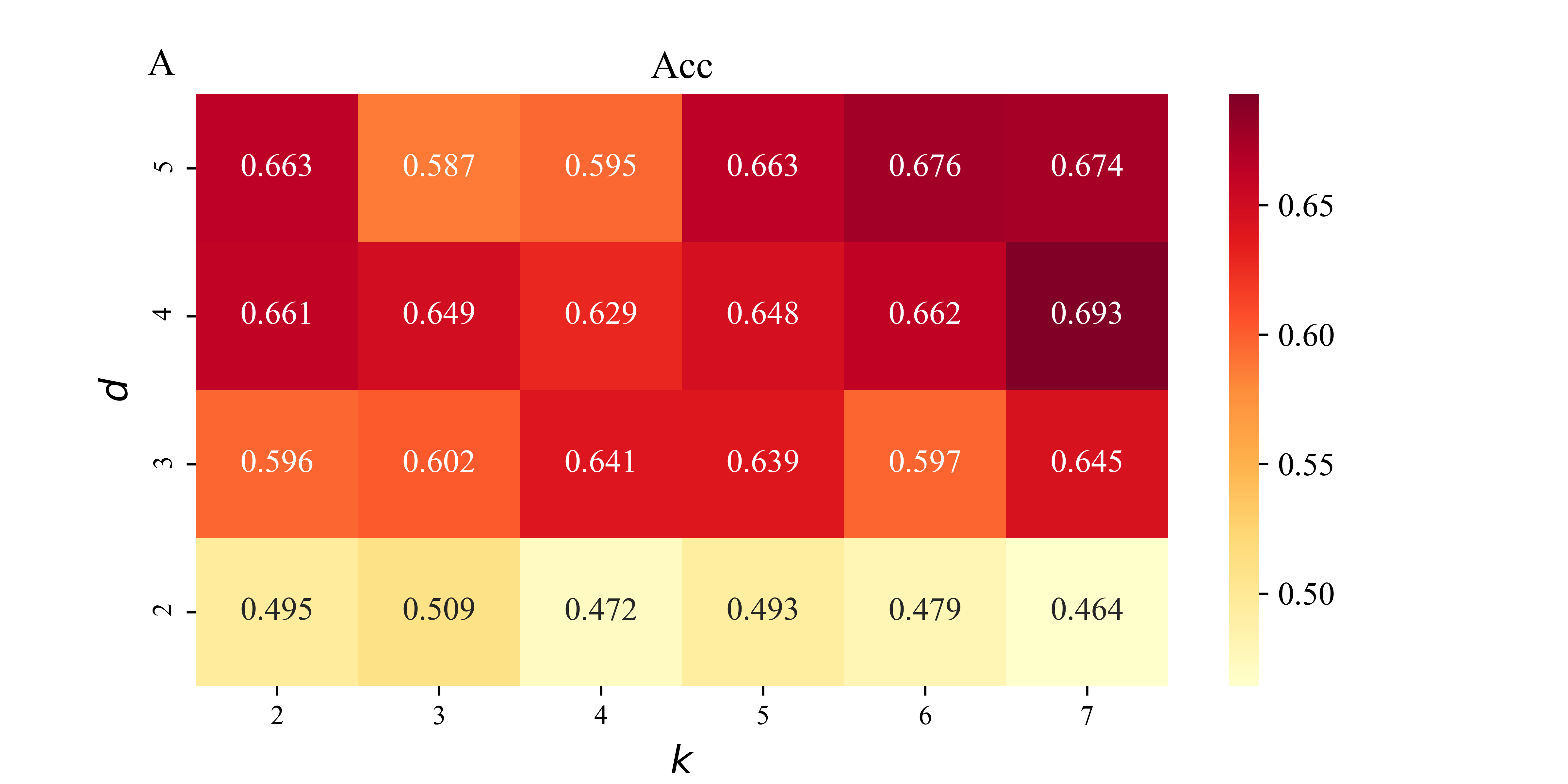}}
	\subfloat{\includegraphics[width = 0.35\textwidth]{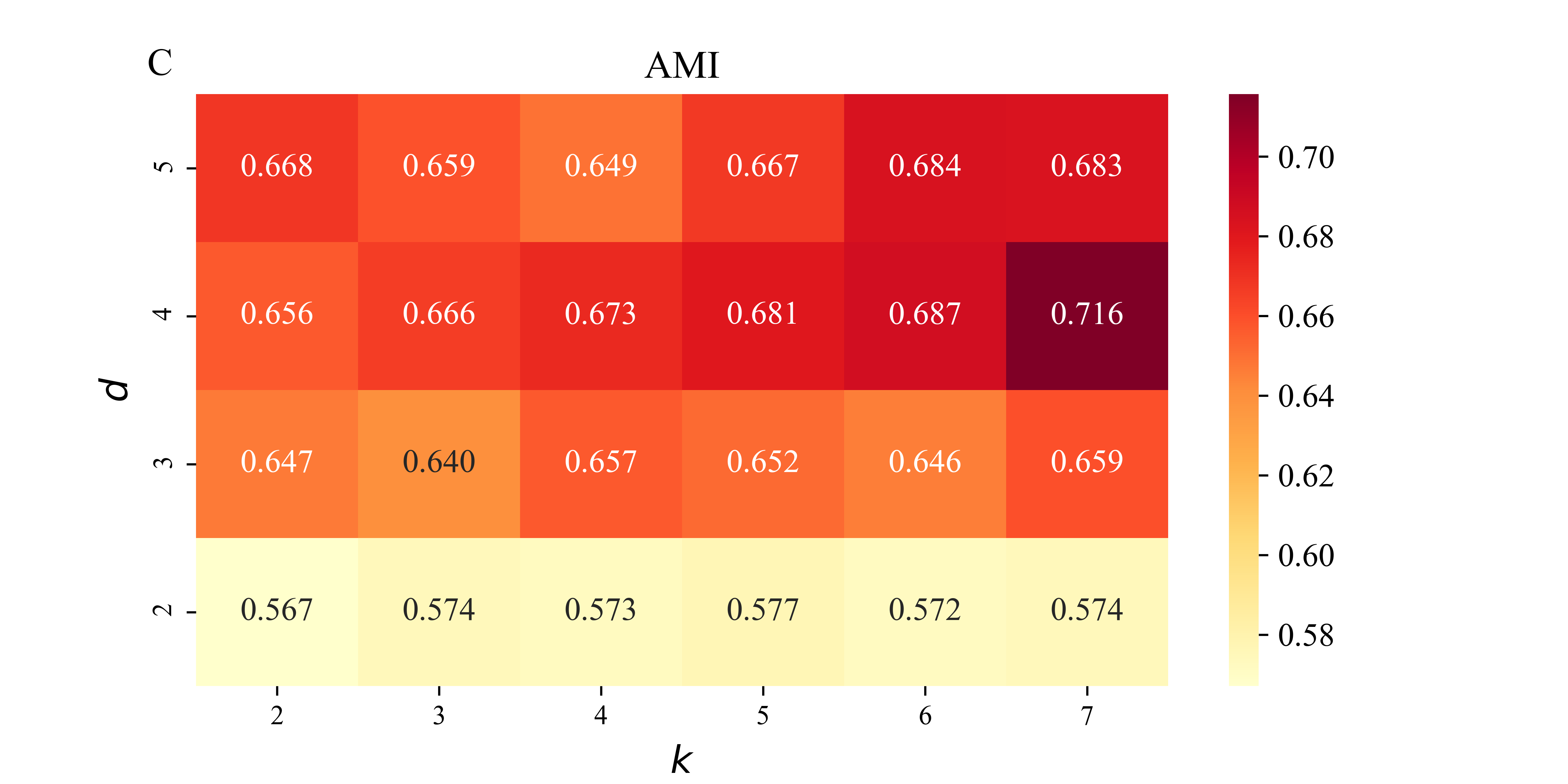}}
\subfloat{\includegraphics[width = 0.35\textwidth]{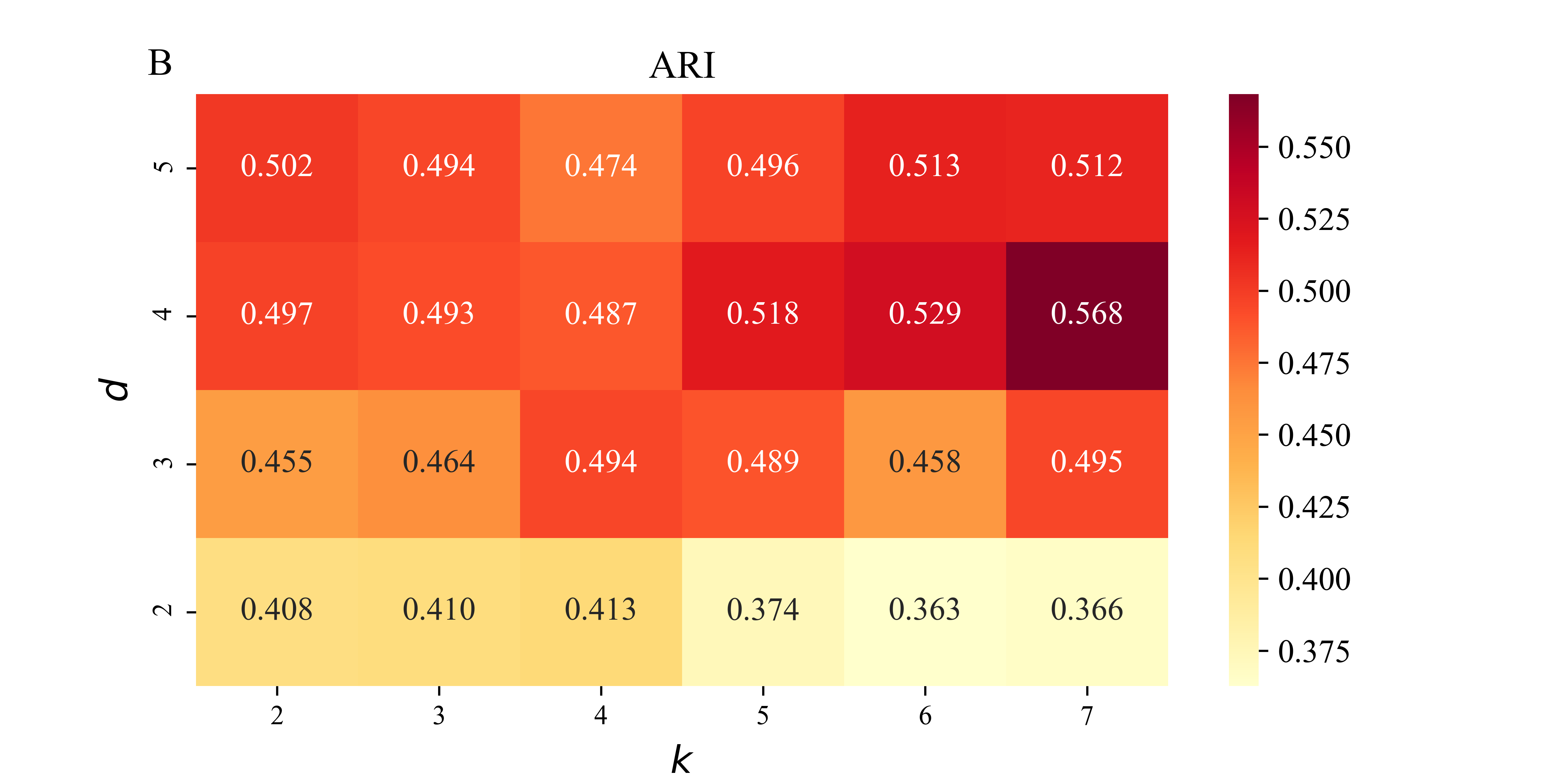}}\\
\subfloat{\includegraphics[width = 0.35\textwidth]{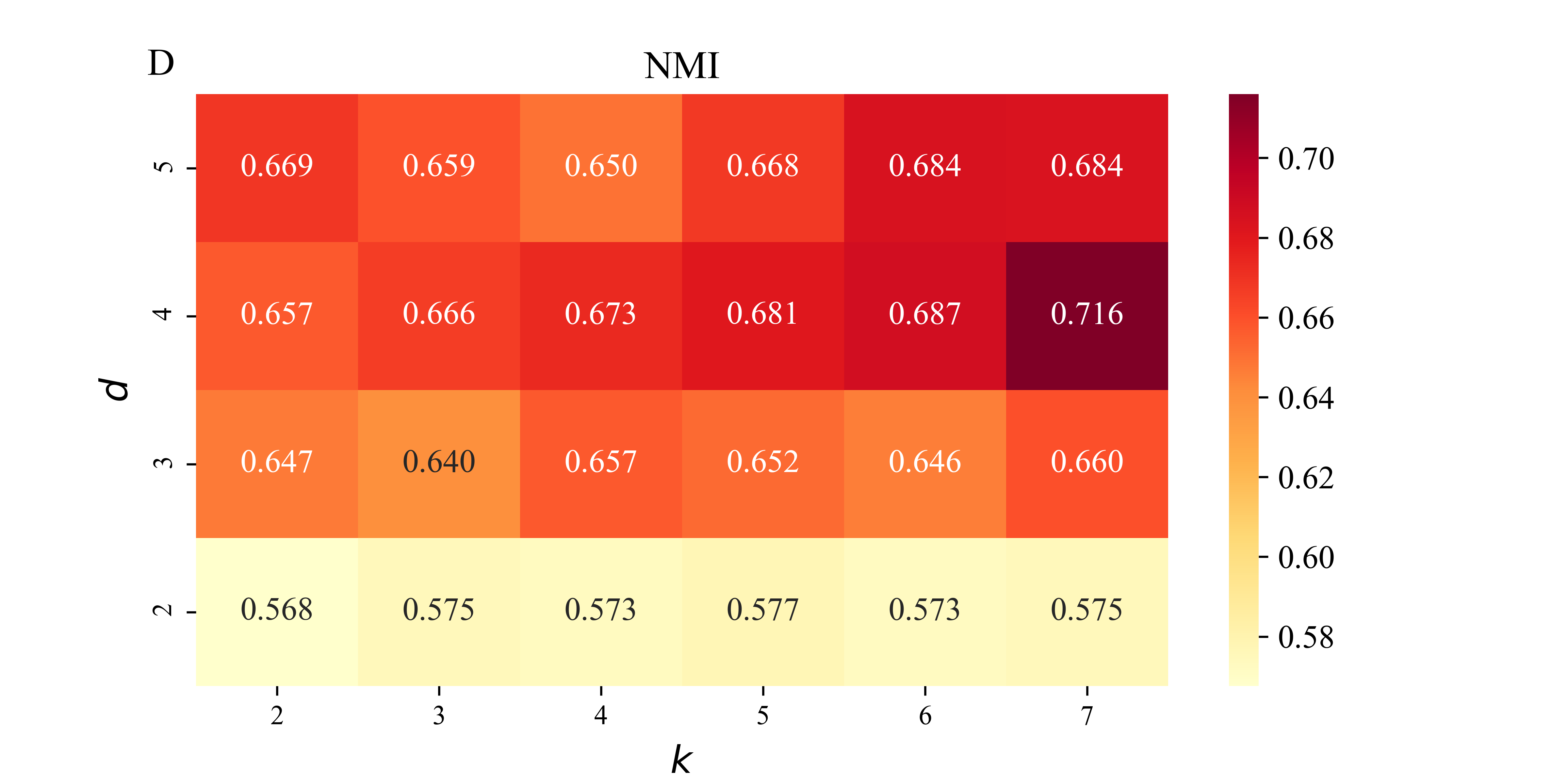}}
\subfloat{\includegraphics[width = 0.35\textwidth]{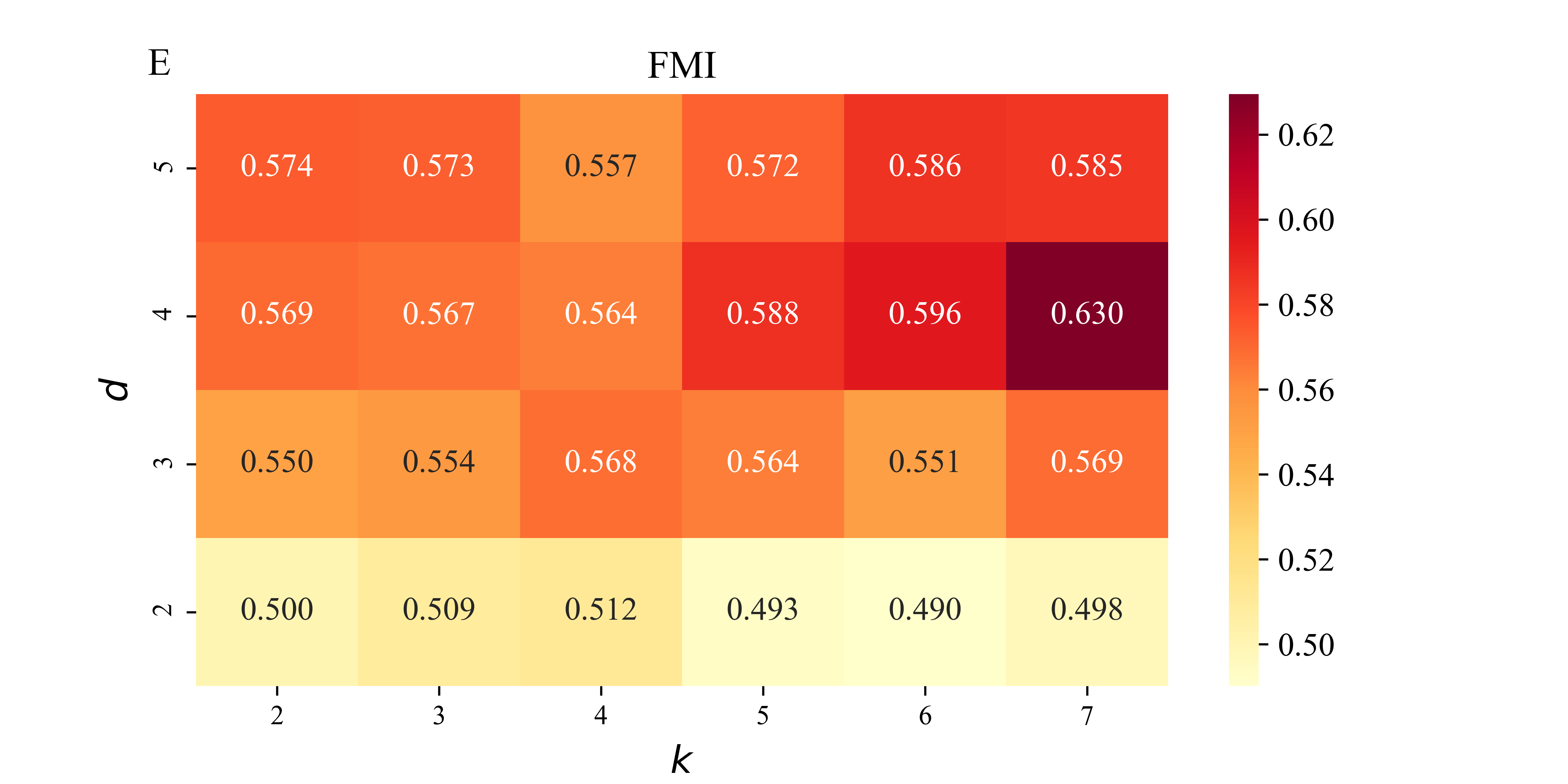}\centering}
\subfloat{\includegraphics[width = 0.35\textwidth]{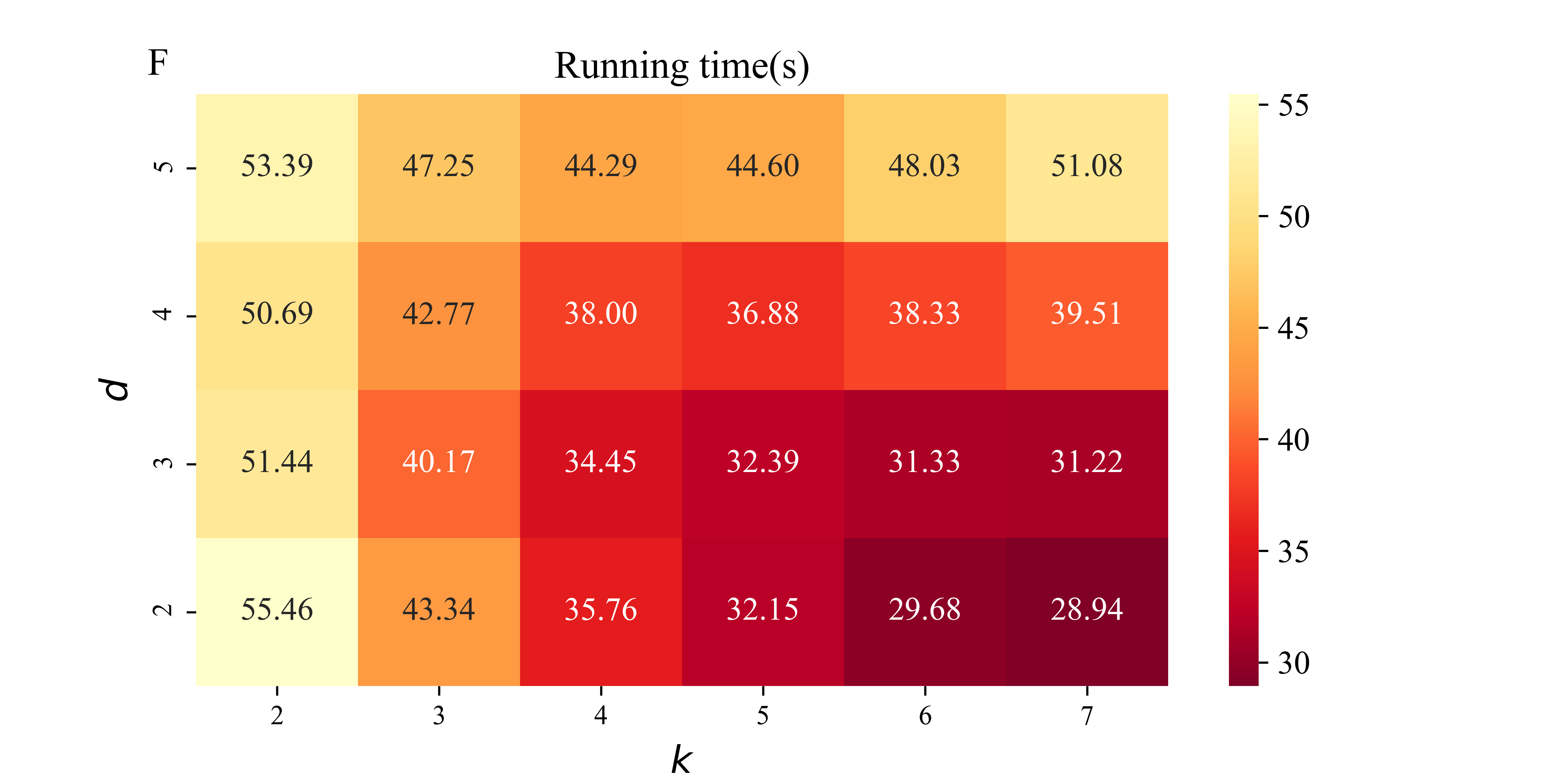}\centering}
\caption{\textbf {Results of six indexes obtained by SKTDPC algorithm for Pendigits dataset under different parameter combinations. }}
\label{fig:labe11}
\end{figure*}

\begin{table*}
\centering
\caption{EXPERIMENTAL RESULTS OF PCA+SKTDPC AND SKTDPC UNDER OPTIMAL PARAMETERS.}
\setlength{\tabcolsep}{3pt}
\begin{tabular}{lccccccc}
\hline\hline
DATA (Pendigits)&	Acc&	AMI&	ARI&	NMI&	FMI&	Running time (s)&	Par\\
\hline
PCA+SKTDPC	&0.693&	0.716&	0.568&	0.716&	0.630&39.51&	7/4\\
SKTDPC&	0.707&	0.744	&0.610	&0.744&	0.661	&182.4&	7\\
\hline\hline
\end{tabular}
\label{tab11}
\end{table*}

We can clearly see that most of the datasets of SKTDPC show stable and ideal clustering effect relatively in a small range of parameters, which shows the excellent robustness of the algorithm. On Spiral, Iris, Wine and Ecoli datasets, although Acc shows relatively large fluctuations, the algorithm can also quickly achieve ideal clustering effect in a small range of parameter adjustment, which is not easy. According to our analysis, the reason for this large fluctuation may be that the algorithm automatically identifies incorrect cluster centers under the parameter $k$ with low Acc value. Therefore, the proposed SKTDPC algorithm can achieve the ideal clustering effect and robustness suitable for most datasets within a small range of parameter adjustment.

\subsection{ADDITIONAL DISCUSSION}
The focus of the present work is on dealing with datasets with arbitrary shape and large size. As for higher dimensional datasets, the improvement for clustering efficiency of SKTDPC is not obvious. However, it is not a vexing problem. In this section, the applicability of SKTDPC on higher dimensional dataset is briefly discussed and some reasonable suggestions are given. Many effective dimension reduction methods can be used to handle this case, such as principal component analysis (PCA), locally linear embedding (LLE), laplacian eigenmaps (LE), etc.

The following is a simple example of processing 16-dimensional dataset Pendigits using PCA as a dimension reduction method. Fig. \ref{fig:labe11} shows the heat map of Acc, AMI, ARI, NMI, FMI and running time obtained by PCA+SKTDPC algorithm by different combinations of $k$ and $d$, where $d$ is the dimension after dimension reduction. The darker the color of the rectangular block is, the better the indicator value is. We can see that Acc, AMI, ARI, NMI and FMI are all optimal with the set of parameters $k=7$ and $d=4$. However, the corresponding running time of 39.51s is not the minimum in the range considered. Meanwhile, there are non-unique parameter combinations with values of running time less than 39.51s. In this case, for determining parameter combination the principle should be satisfied, that is the highest efficiency is achieved on the premise of guaranteeing the ideal clustering effect, instead of pursuing speed blindly. In accord with this principle, the final parameter combination, $k=7$ and $d=4$, is adopted. Under this parameters, the running time results of PCA+SKTDPC and SKTDPC are shown in Table \ref{tab11}.

\section{CONCLUSIONS}
An extended DPC algorithm, called SKTDPC, is successfully proposed by K-d tree, sparse search and second-order difference methods. Applying to eight synthetic datasets with two dimensions and six real datasets, comparisons have been carried out between SKTDPC and the six typical clustering algorithms, including FSDPC, the original DPC, DGDPC, DPC-KNN, DBSCAN and K-means algorithms. The main conclusions can be summarized as follows.

Firstly, the algorithmic complexity is obviously reduced by dual accelerations. One is the acceleration for calculation of the local density with a sparse distance matrix, which is attributed to fast search of $k$ nearest neighbors by K-d tree. Another is the acceleration for calculation of the relative-separation by a sparse search strategy with the intersection between the set of $k$ nearest neighbors and the set consisting of the data points with larger local density for any data point. Experimental validation demonstrates that compared with the DPC series algorithms, SKTDPC algorithm can achieve higher clustering efficiency on all datasets. The larger the dataset, the greater the advantage of SKTDPC.

Secondly, experiments indicate that SKTDPC algorithm can realize the best clustering effect in general, compared with the other algorithms. Furthermore, it is indicated that compared with K-means and DBSCAN algorithm, SKTDPC algorithm has a relatively stronger general applicability for datasets with arbitrary distribution characteristics, even though they have better clustering efficiency in some cases.

Finally, the second-order difference method for decision values is adopted to determine the location of the mutation-point adaptively, which avoids the trouble of selecting the cluster centers manually. It is also verified by experiments that the present method can produce the correct number of cluster centers automatically.

For future work, datasets with insufficient target data, high complexity or high sparsity will be further explored and studied to enhance the application ability of clustering algorithm.

\begin{IEEEbiography}[{\includegraphics[width=1in,height=1.25in,clip,keepaspectratio]{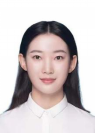}}]{YUNXIAO SHAN}  was born in 1995. She is currently pursuing the Ph.D. degree in the School of Science, Harbin University of Science and Technology, Harbin, China. Her current research interests include machine learning, data mining, fuzzy logic.
\end{IEEEbiography}

\begin{IEEEbiography}[{\includegraphics[width=1in,height=1.25in,clip,keepaspectratio]{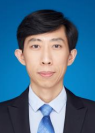}}]{SHU LI}  was born in 1980. He received the Ph.D. degree in Tianjin University, Tianjin, China. He is currently a professor in the School of Electrical and Electronic Engineering, Harbin University of Science and Technology, China. As the first author or corresponding author, he has published over 20 SCI indexed papers in the internationally renowned journals. He also has obtained 6 software copyrights. His research interests include machine learning, data mining, modelling and calculation for phase transition theory.
\end{IEEEbiography}

\begin{IEEEbiography}[{\includegraphics[width=1in,height=1.25in,clip,keepaspectratio]{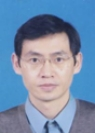}}]{FUXIANG LI}  was born in 1972. He received the Ph.D. degree in Harbin Institute of Technology, Harbin, China. He is currently a professor in the School of Science, Harbin University of Science and Technology, China. He has published more than 20 papers. His research interests include nonlinear numerical analysis, computational mathematics, machine learning.
\end{IEEEbiography}

\begin{IEEEbiography}[{\includegraphics[width=1in,height=1.25in,clip,keepaspectratio]{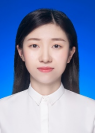}}]{YUXIN CUI}   was born in 1995. She is currently pursuing the Ph.D. degree in the School of Science, Harbin University of Science and Technology, Harbin, China. Her current research interests include machine learning, stochastic logic system analysis. 
\end{IEEEbiography}

\begin{IEEEbiography}[{\includegraphics[width=1in,height=1.25in,clip,keepaspectratio]{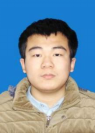}}]{SHUAI LI} was born in 1998. He is currently pursuing the Ph.D. degree in the School of Materials Science and Chemical Engineering, Harbin University of Science and Technology, Harbin, China. His current research interests include machine learning, material research, high entropy alloy.
\end{IEEEbiography}

\begin{IEEEbiography}[{\includegraphics[width=1in,height=1.25in,clip,keepaspectratio]{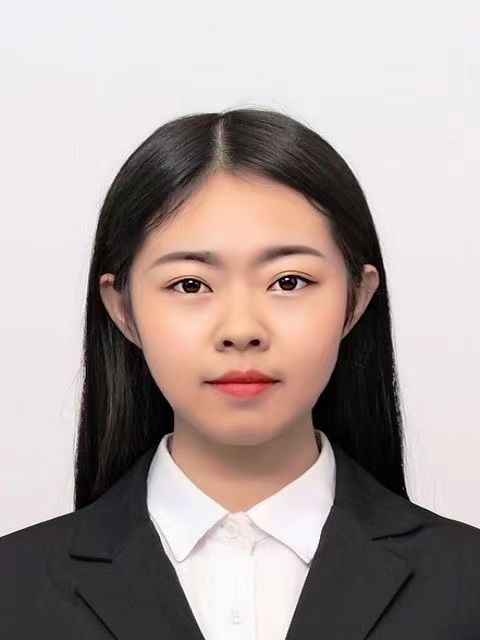}}]{MING ZHOU}  was born in 1997. She is currently pursuing the M.S. degree in the School of Science, Harbin University of Science and Technology, Harbin, China.Her current research interests include machine learning, mathematics of computation, nonlinear numerical analysis.
\end{IEEEbiography}

\begin{IEEEbiography}[{\includegraphics[width=1in,height=1.25in,clip,keepaspectratio]{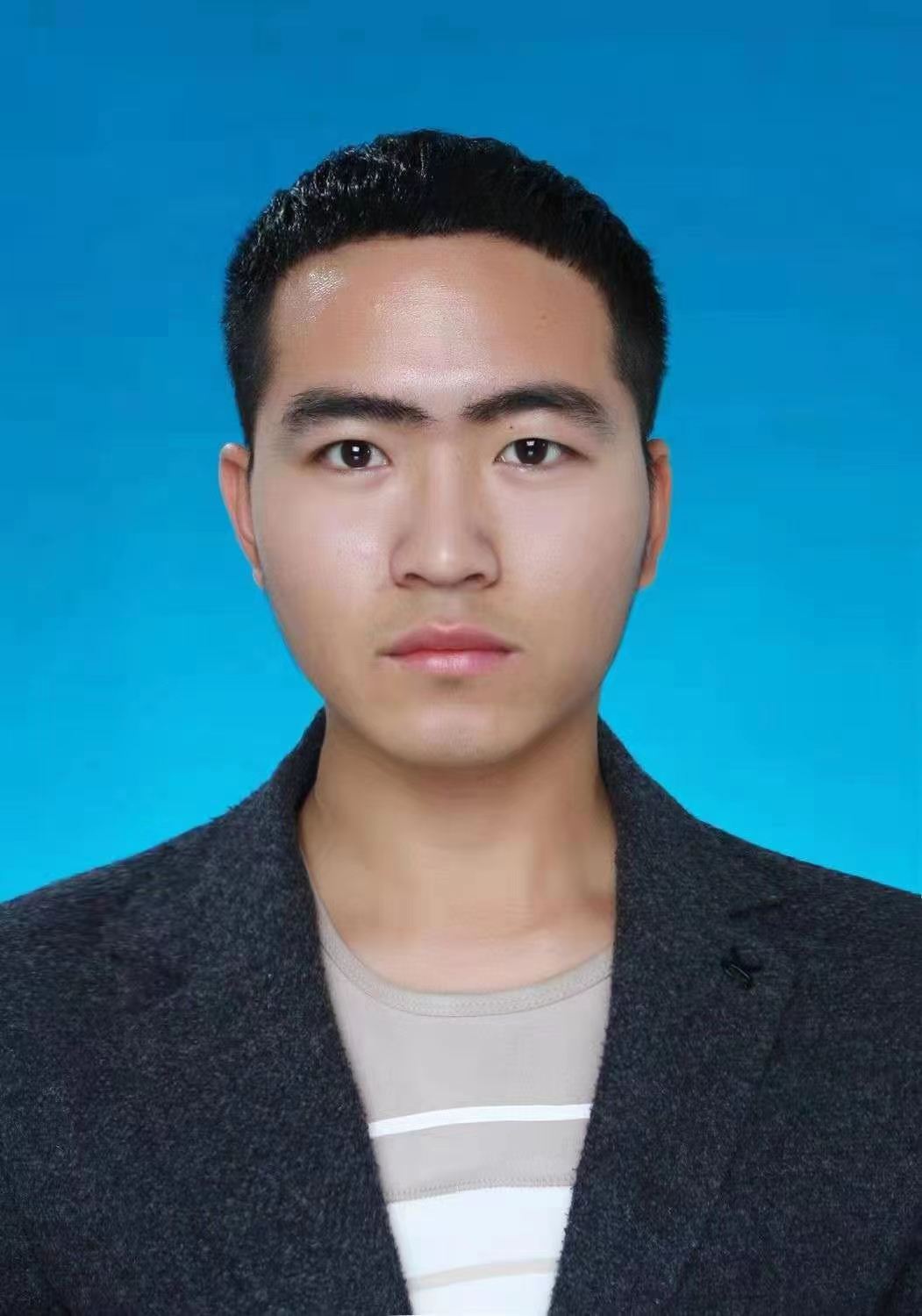}}]{XIANG LI}  was born in 1997. He is currently pursuing the M.S.degree in the School of Science, Harbin University of Science and Technology, Harbin,China. His current research interests include machine learning, data mining.
\end{IEEEbiography}
\EOD

\end{document}